\documentclass[runningheads]{llncs}
\usepackage{graphicx}
\usepackage{amsmath,amssymb} 
\usepackage{color}
\usepackage[width=122mm,left=12mm,paperwidth=146mm,height=193mm,top=12mm,paperheight=217mm]{geometry}

\usepackage{comment}
\usepackage{subfig}
\usepackage{multirow}
\usepackage{booktabs}
\usepackage{algorithm,algcompatible,amsmath}
\usepackage{multirow}
\usepackage{eufrak}
\usepackage{upgreek}

\begin{document}
\pagestyle{headings}
\mainmatter

\title{Joint Blind Motion Deblurring and Depth Estimation of Light Field} 

\author{Dongwoo Lee$^\dag$~\qquad Haesol Park$^\dag$~\qquad In Kyu Park$^\ddag$~\qquad Kyoung Mu Lee$^\dag$\\
$^\dag$Department of ECE, ASRI, Seoul National University\\
$^\ddag$Department of Information and Communication Engineering, Inha University\\
}


\institute{
}

\maketitle

\begin{abstract}
Removing camera motion blur from a single light field is a challenging task since it is highly ill-posed inverse problem.
The problem becomes even worse when blur kernel varies spatially due to scene depth variation and high-order camera motion.
In this paper, we propose a novel algorithm to estimate all blur model variables jointly, including latent sub-aperture image, camera motion, and scene depth from the blurred 4D light field.
Exploiting multi-view nature of a light field relieves the inverse property of the optimization by utilizing strong depth cues and multi-view blur observation.
The proposed joint estimation achieves high quality light field deblurring and depth estimation simultaneously under arbitrary 6-DOF camera motion and unconstrained scene depth.
Intensive experiment on real and synthetic blurred light field confirms that the proposed algorithm outperforms the state-of-the-art light field deblurring and depth estimation methods.

\end{abstract}

\section{Introduction}
For the last decade, motion deblurring has been an active research topic in computer vision.
Motion blur is produced by relative motion between camera and scene during the exposure where blur kernel, {\it i.e.} point spread function (PSF), is spatially non-uniform.
In blind non-uniform deblurring problem, pixel-wise blur kernels and corresponding sharp image are estimated simultaneously.

Early works on motion deblurring~\cite{chan1998total,cho2009fast,fergus2006removing,shan2008high,xu2010two} focus on removing spatially uniform blur in the image.
However, the assumption of uniform motion blur is often broken in real world due to nonhomogeneous scene depth and rolling motion of camera.
Recently, a number of methods~\cite{cho2007removing,gupta2010single,hu2014joint,hu2012fast,ji2012two,kim2014segmentation,sun2015learning,whyte2012non,zheng2013forward} have been proposed for non-uniform deblurring.
However, they still can not completely handle non-uniform blur caused by scene depth variation.
The main challenge lies in the difficulty of estimating the scene depth with only single observation, which is highly ill-posed.

\begin{figure}[t]
	\begin{center}
		\vspace*{0.15in}
		\subfloat[]{\includegraphics[width=0.25\textwidth]{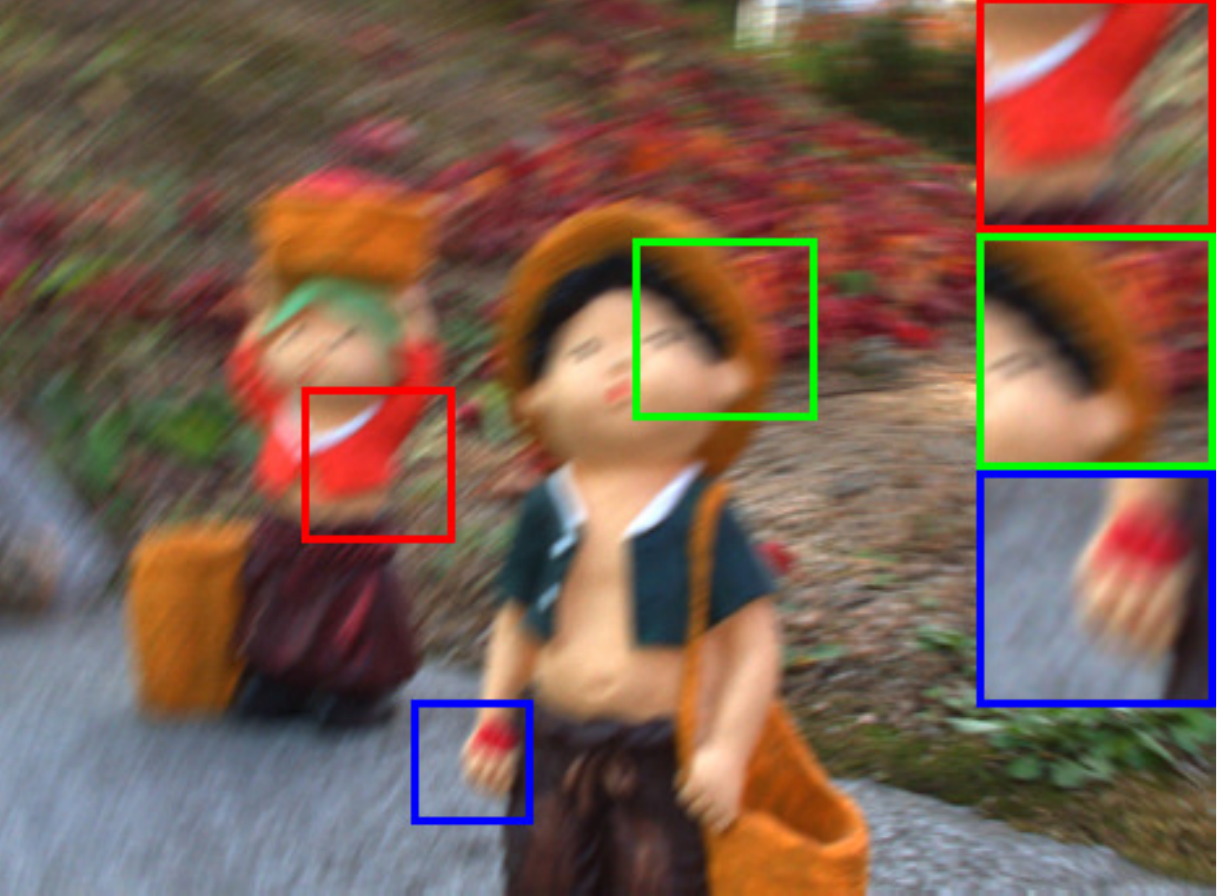}}\hspace*{-0.01in}
		\subfloat[]{\includegraphics[width=0.25\textwidth]{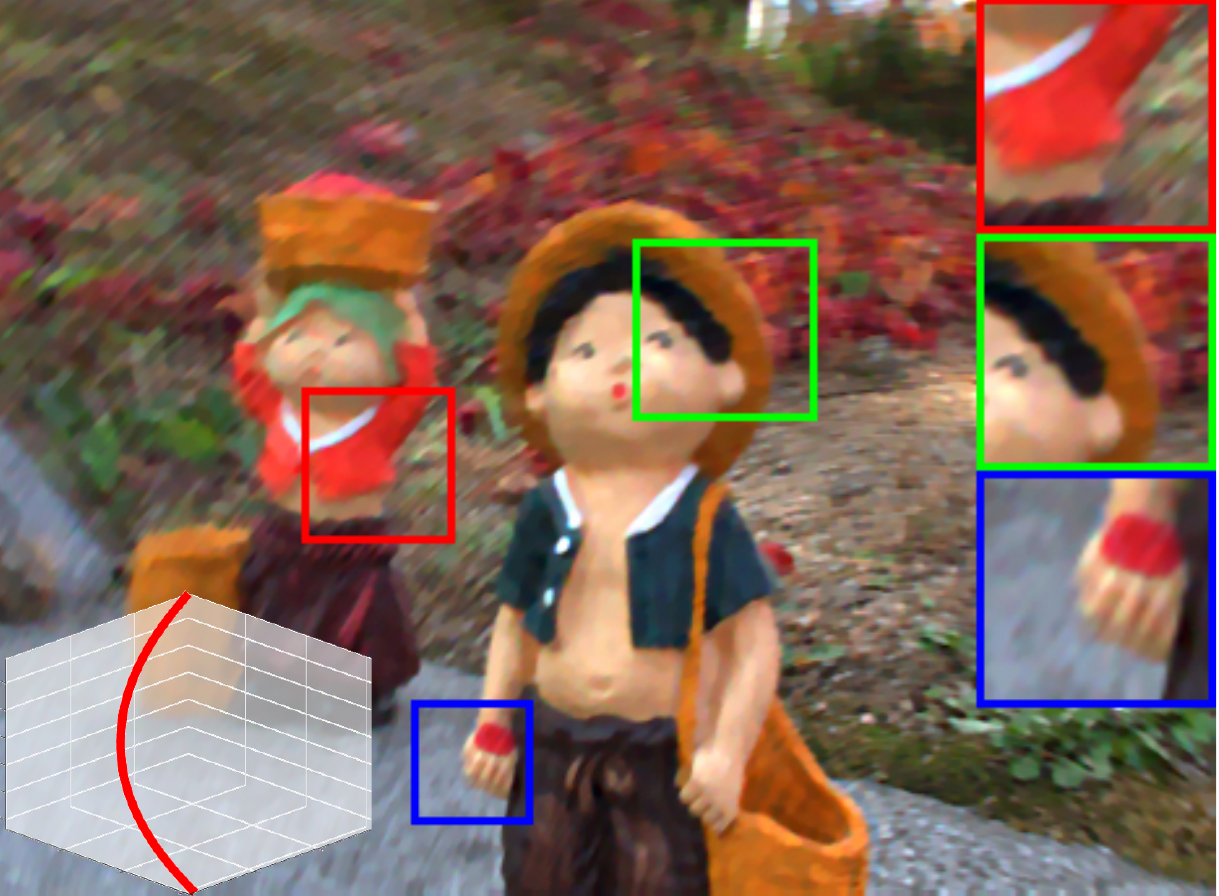}}\hspace*{-0.01in}
		\subfloat[]{\includegraphics[width=0.25\textwidth]{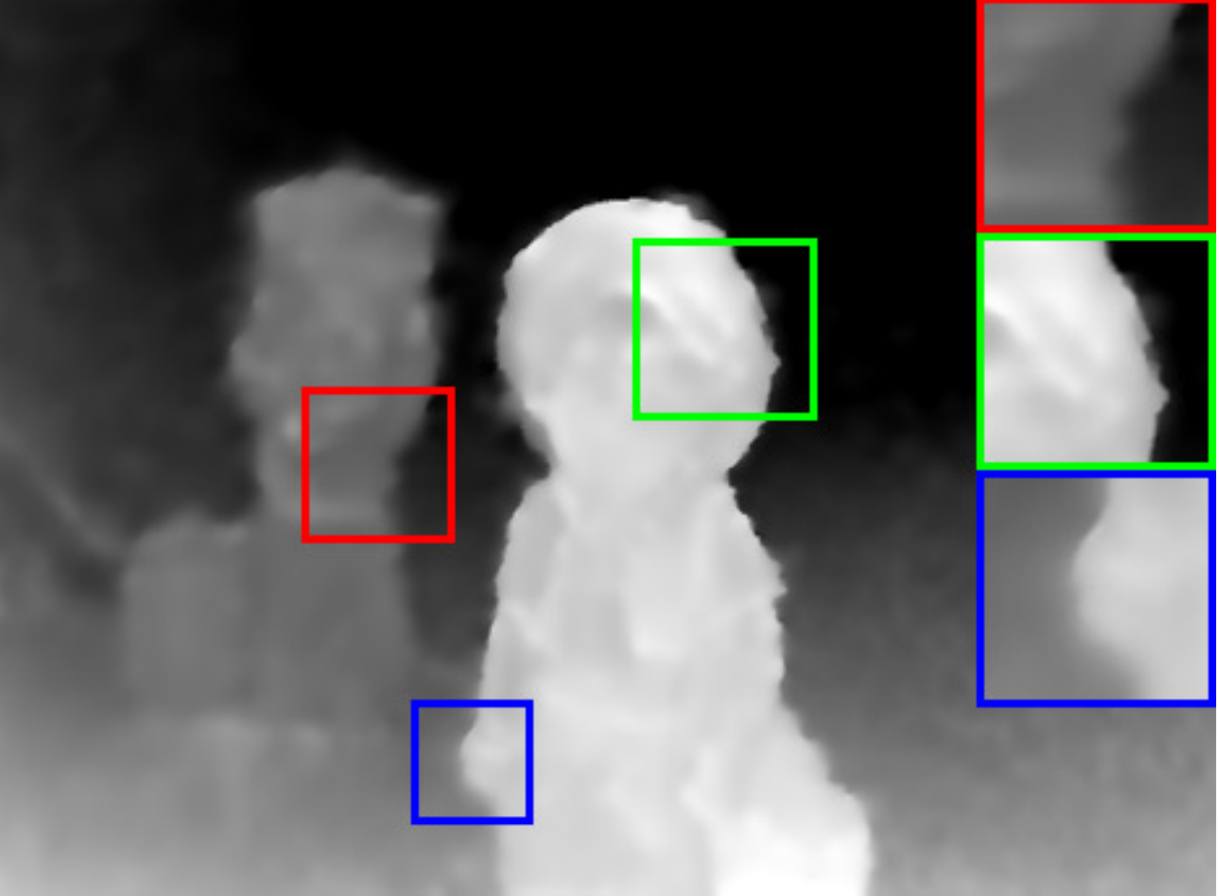}}\hspace*{-0.01in}
		\subfloat[]{\includegraphics[width=0.25\textwidth]{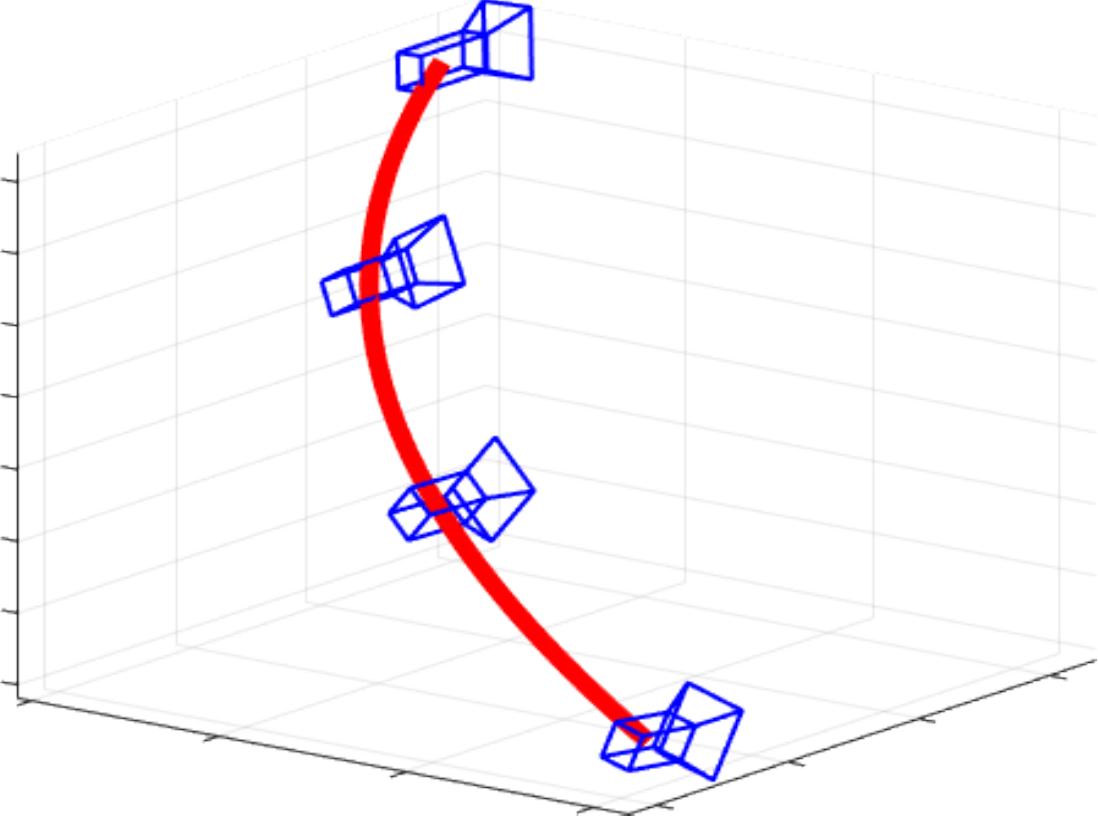}}	
	\end{center}
	\vspace*{-0.05in}
	\caption{The proposed algorithm jointly estimates latent image, depth map, and camera motion from a single light field. (a) Center-view of blurred light field sub-aperture image. (b) Deblurred image of (a). (c) Estimated depth map. (d) Camera motion path and orientation (6-DOF).}
	\label{fig:comp0}
\end{figure}
A light field camera ameliorates the ill-posedness of single-shot deblurring problem of the conventional camera.
4D light field is equivalent to multi-view images with narrow baseline, {\it i.e.} sub-aperture images, taken with an identical exposure~\cite{ng2005light}.
Consequently, motion deblurring using light field can be leveraged by its multi-dimensional nature of captured information.
First, strong depth cue is obtained by employing multi-view stereo matching between sub-aperture images.
In addition, different blurs in the sub-aperture images can help the optimization converge more fast and precise.

\begin{figure*}[t]
	\vspace*{0.15in}
	\centering
	\scalebox{1.1}{
		\includegraphics[width=0.90\textwidth]{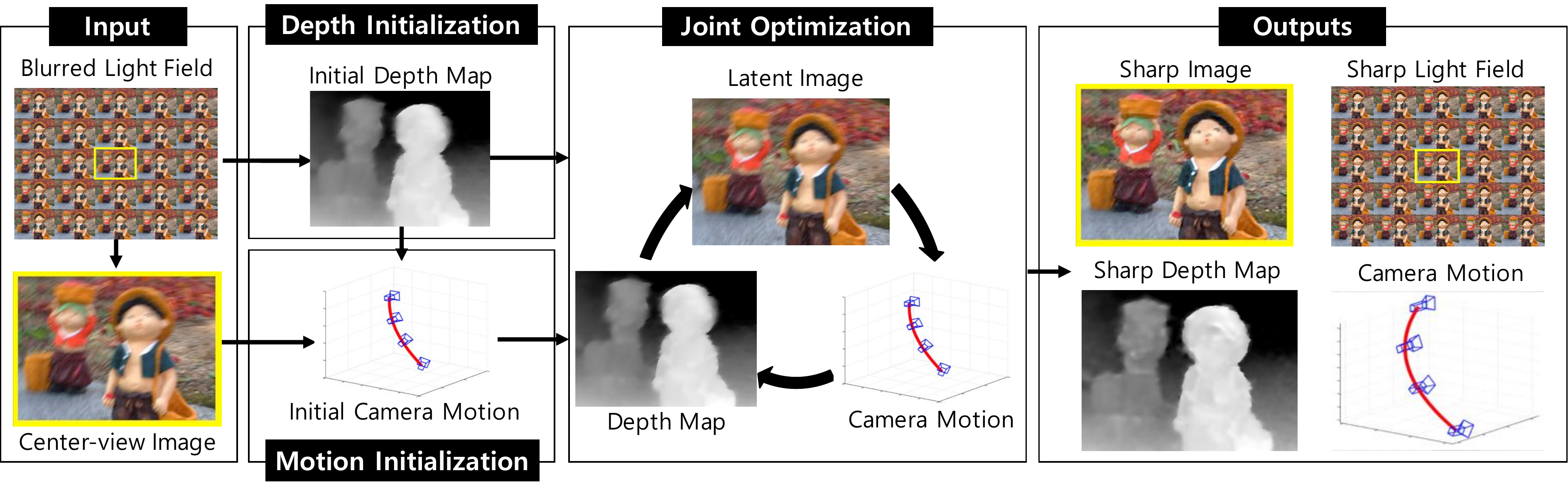}
	}
	\vspace*{-0.05in}
	\caption{Overview of the proposed algorithm. The proposed algorithm jointly estimates the latent image, depth map, and camera motion from a single light field.}
	\label{fig:comp1}
\end{figure*}

In this paper, we propose an efficient algorithm to jointly estimate latent image, sharp depth map, and 6-DOF camera motion from a blurred single 4D light field as shown in Figure~\ref{fig:comp0}.
In the proposed light field blur model, latent sub-aperture images are formulated by 3D warping of the center-view sharp image using the depth map and the 6-DOF camera motion.
Then, motion blur is modeled as the integral of latent sub-aperture images during the shutter open.
Note that the proposed center-view parameterization reduces light field deblurring problem in lower dimension comparable to a single image deblurring.
The joint optimization is performed in an alternating manner, in which the deblurred image, depth map, and camera motion are refined during iteration.
The overview of the proposed algorithm is shown in Figure~\ref{fig:comp1}.
In overall, the contribution of this paper is summarized as follows.
\begin{itemize}
  \item[$\bullet$] We propose a joint method which simultaneously solves deblurring, depth estimation, and camera motion estimation problems from a single light field.
  \item[$\bullet$] Unlike the previous state-of-the-art algorithm, the proposed method handles blind light field motion deblurring under 6-DOF camera motion.
  \item[$\bullet$] Practical and extensible blur formulation that can be extended to any multi-view camera system.
\end{itemize}
\section{Related Works}

\textbf{Conventional Single Image Deblurring.}~One way to effectively remove the spatially-variant motion in a conventional single image is to first find the motion density function (MDF) and then generate the pixel-wise kernel from this function~\cite{gupta2010single,hu2014joint,hu2012fast}.
Gupta~et al.~\cite{gupta2010single} modeled the camera motion in discrete 3D motion space comprising $x$, $y$ translation and in-plane rotation.
They performed deblurring by iteratively optimizing the MDF and the latent image that best describe the blurred image.
Similar model was used by Hu and Yang~\cite{hu2012fast} in which MDF was modeled with 3D rotations.
These methods of using MDF well parameterize the spatially-variant blur kernel into low dimensions.
However, modeling the motion blur using MDF only in depth varying images is difficult, because the motion blur is determined by both camera motion and scene depth.
In~\cite{hu2014joint}, the image was segmented by the matting algorithm, and the MDF and representative depth values ​​of each region were found through the expectation-maximization algorithm.

A few methods~\cite{kim2014segmentation,sun2015learning} estimated linear blur kernels locally, and they showed acceptable results for the arbitrary scene depth.
Kim and Lee~\cite{kim2014segmentation} jointly estimated the spatially varying motion flow and the latent image.
Sun~et al.~\cite{sun2015learning} adopted a learning method based on convolutional neural network~(CNN) and assumed that the motion was locally linear.
However, the locally linear blur assumption does not hold in large motion.
\vspace*{0.3cm}

\noindent\textbf{Video and Multi-View Deblurring.}~Xu and Jia~\cite{xu2012depth} decomposed the image region according to the depth map obtained from a stereo camera and recombined them after independent deblurring.
Recently, several methods~\cite{cho2012video,hyun2015generalized,park2017joint,sellent2016stereo,wulff2014modeling} have addressed the motion blur problem in video sequences.
Video deblurring shows good performance, because it exploits optical flow as a strong guide for motion estimation.
\vspace*{0.3cm}

\noindent\textbf{Light Field Deblurring.}~Light field with two plane parameterization is equivalent to  multi-view images with narrow baseline.
It contains rich geometric information of rays in a single-shot image.
These multi-view images are called sub-aperture images and individual sub-aperture images show slightly different blur pattern due to the viewpoint variation.
In last a few years, several approaches~\cite{chandramouli2014motion,dansereau2016motion,jin2015bilayer,snoswell2014lf,srinivasan2017light} have been proposed to perform motion deblurring on the light field.
Chandramouli~et~al.~\cite{chandramouli2014motion} addressed the motion blur problem in the light field for the first time.
They assumed constant depth and uniform motion to alleviate the complexity of the imaging model.
Constant depth means that the light field has little information about 3D scene structure, which depletes the advantages of light field.
Jin~et al.~\cite{jin2015bilayer} quantized the depth map into two layers and removed the motion blur in each layer.
Their method assumed that the camera motion is in-plane translation and utilized depth value as a scale factor of translational motion.
Although their model handles non-uniform blur kernel related to the depth map, a more general depth variation and camera motion should be considered for application to real-world scenes.
Dansereau~et al.~\cite{dansereau2016motion} applied the Richardson-Lucy deblurring algorithm to the light field with non-blind 6-DOF motion blur.
Although their method dealt with 6-DOF motion blur, it was assumed that the ground truth camera motion was known.
Unlike~\cite{dansereau2016motion}, in this paper, we address the problem of blind deblurring which is a more highly ill-posed problem.
Srinivasan~et al.~\cite{srinivasan2017light} solved the light field deblurring under 3D camera motion path and showed visually pleasing result.
However, their methods do not consider 3D orientation change of the camera.

In contrast to the previous works of light field deblurring, the proposed method completely handles 6-DOF motion blur and unconstrained scene depth variation.

\section{Motion Blur Formulation in Light Field}
\label{sec3}

A pixel in a 4D light field has four coordinates, {\em i.e.} $(x,y)$ for spatial and $(u,v)$ for angular coordinates.
A light field can be interpreted as a set of $u\times v$ multi-view images with narrow baseline, which are often called sub-aperture images~\cite{ng2006digital}.
Throughout this paper, a sub-aperture image is represented as $I(\mathbf{x},{\mathbf{u}})$ where $\mathbf{x}=(x,y)$ and $\mathbf{u}=(u,v)$.
For each sub-aperture image, the blurred image $B(\mathbf{x},{\mathbf{u}})$ is the average of the sharp images ${I}_{t}(\mathbf{x},{\mathbf{u}})$ during the shutter open over $[t_0 , t_1]$ as follows:
\begin{align}
B(\mathbf{x},{\mathbf{u}}) = \int_{t_0}^{t_1} {I}_{t}(\mathbf{x},{\mathbf{u}}) dt.
\label{eq:blurEq}
\end{align}

Following the blur model of~\cite{park2017joint,sellent2016stereo}, we approximate all the blurred sub-aperture images by projecting a single latent image with 3D rigid motion.
We choose the center-view ($\mathbf{c}$) of sub-aperture images and the middle of the shutter time ($t_r$) as the reference angular position and the time stamp of the latent image.
With above notations, the pixel correspondence from each sub-aperture image to the latent image $I_{t_r}(\mathbf{x},\mathbf{c})$ is expressed as follows:
\begin{align}
{I}_{t}(\mathbf{x},{\mathbf{u}})={I}_{t_r}(w_{t}(\mathbf{x},\mathbf{u}),{\mathbf{c}}),
\end{align}
where
\begin{align}
w_{t}(\mathbf{x},\mathbf{u})=\Pi_{\mathbf{c}}(\mathrm{P}^{\mathbf{c}}_{t_r}(\mathrm{P}^{\mathbf{u}}_{t})^{\scriptscriptstyle -1}\Pi^{\scriptscriptstyle -1}_{\mathbf{u}}(\mathbf{x},D_{t}(\mathbf{x},\mathbf{u}))).
\label{eq:warping}
\end{align}
$w_{t}(\mathbf{x},\mathbf{u})$ computes the warped pixel position from $\mathbf{u}$ to $\mathbf{c}$, and from $t$ to $t_r$.
$\Pi_{\mathbf{c}}$, $\Pi^{\scriptscriptstyle -1}_{\mathbf{u}}$ are the projection and back-projection function between the image coordinate and the 3D homogeneous coordinate using the camera intrinsic parameters.
Matrices $\mathrm{P}^{\mathbf{c}}_{t_r}$ and $\mathrm{P}^{\mathbf{u}}_{t}\in SE(3)$ denote the 6-DOF camera pose at the corresponding angular position and the time stamp. $D_{t}(\mathbf{x},\mathbf{u})$ is the depth map at the time stamp $t$.

In the proposed model, the blur operator $\Psi(\cdot)$ is defined by approximating the integral in (\ref{eq:blurEq}) as a finite sum as follows:
\begin{align}
B(\mathbf{x},\mathbf{u})\approx (\Psi\circ I)(\mathbf{x},\mathbf{u}),
\end{align}
where
\begin{align}
(\Psi\circ I)(\mathbf{x},\mathbf{u}) = \frac{1}{M}\sum^{M-1}_{m=0}I_{t_r}(w_{t_m}(\mathbf{x},\mathbf{u}), \mathbf{c}).
\label{eq:blurOp}
\end{align}
In (\ref{eq:blurOp}), $t_m$ is $m_{th}$ uniformly sampled time stamp during the interval $[t_0,t_1]$.

Our goal is to formulate $(\Psi\circ I)(\mathbf{x},\mathbf{u})$ with only center-view variables, {\em i.e.} $I_{t_r}(\mathbf{x},\mathbf{c})$, $D_{t_r}(\mathbf{x},\mathbf{c})$, and $\mathrm{P}^{\mathbf{c}}_{t_0}$.
$\mathrm{P}^{\mathbf{u}}_{t_m}$ and $D_{t_m}(\mathbf{x},\mathbf{u})$ are variables related to $\mathbf{u}$ in the warping function (\ref{eq:blurOp}).
Therefore, we parameterize $\mathrm{P}^{\mathbf{u}}_{t_m}$ and $D_{t_m}(\mathbf{x},\mathbf{u})$ by employing center-view variables.
Because the relative camera pose  $\mathrm{P}^{\scriptscriptstyle \mathbf{c\rightarrow u}}$ is fixed over time, $\mathrm{P}^{\mathbf{u}}_{t_m}$ is expressed by $\mathrm{P}^{\mathbf{c}}_{t_0}$ and $\mathrm{P}^{\mathbf{c}}_{t_1}$ as follows:
\begin{align}
\mathrm{P}^{\mathbf{u}}_{t_m}=\mathrm{P}^{\scriptscriptstyle \mathbf{c\rightarrow u}}\mathrm{P}^{\mathbf{c}}_{t_m},
\end{align}
\begin{align}
\mathrm{P}^{\mathbf{c}}_{t_m}=\exp(\frac{m}{M}\log(\mathrm{P}^{\mathbf{c}}_{t_1}{(\mathrm{P}^{\mathbf{c}}_{t_0})}^{\scriptscriptstyle -1}))\mathrm{P}^{\mathbf{c}}_{t_0},
\label{eq:motionPath}
\end{align}
where $\exp$ and $\log$ denote the exponential and logarithmic maps between Lie group $SE(3)$ and Lie algebra $\mathfrak{se}(3)$ space~\cite{blanco2010tutorial}.
To minimize the viewpoint shift of the latent image, we assume $\mathrm{P}^{\mathbf{c}}_{t_1}=(\mathrm{P}^{\mathbf{c}}_{t_0})^{\scriptscriptstyle -1}$ which makes $\mathrm{P}^{\mathbf{c}}_{t_m}$ an identity matrix when $t_m=t_r$.
Note that we use the camera path model used in~\cite{park2017joint,sellent2016stereo}.
However, the B\'{e}zier camera path model used in~\cite{srinivasan2017light} can be directly applied to (\ref{eq:motionPath}) as well.
$D_{t_m}(\mathbf{x},\mathbf{u})$ is also represented by $D_{t_r}(\mathbf{x},\mathbf{c})$ by forward warping and interpolation.

In order to estimate all blur variables in the proposed light field blur model, we need to recover the latent variables, {\it i.e.} $I_{t_r}(\mathbf{x},\mathbf{c})$, $D_{t_r}(\mathbf{x},\mathbf{c})$, and $\mathrm{P}^{\mathbf{c}}_{t_0}$.
We model an energy function as follows:
\begin{align}
\begin{split}
E & =  \sum_{\mathbf{u}}\sum_{\mathbf{x}}\lambda_u\|(\Psi\circ I)(\mathbf{x},\mathbf{u})-B(\mathbf{x},{\mathbf{u}})\|_1\\
& + \lambda_L\sum_{\mathbf{x}}\|\nabla I_{t_r}(\mathbf{x},\mathbf{c})\|_2+\lambda_D\sum_{\mathbf{x}}\|\nabla D_{t_r}(\mathbf{x},\mathbf{c})\|_2.
\label{eq:energy}
\end{split}
\end{align}
The data term imposes the brightness consistency between the input blurred light field and the restored light field.
Notice that the L1-norm is employed in our approach as in~\cite{kim2014segmentation}, where it effectively removes the ringing artifact around object boundary and provides more robust deblurring results on large depth change.
The last two terms are the total variation~(TV) regularizers~\cite{beck2009fast} for the latent image and the depth map, respectively.

In our energy model, $D_{t_r}(\mathbf{x},\mathbf{c})$ and $\mathrm{P}^{\mathbf{c}}_{t_0}$ are implicitly included in the warping function (\ref{eq:blurOp}).
The pixel-wise depth $D_{t_r}(\mathbf{x},\mathbf{c})$ determines the scale of the motion at each pixel.
At the boundary of an object where depth changes abruptly, there is a large difference of the blur kernel size between the near and farther objects.
If the optimization is performed without considering this, the blur will not be removed well at the boundary of the object.

Simultaneously optimizing the three variables is complicated because the warping function (\ref{eq:blurOp}) has severe nonlinearity.
Therefore, our strategy is to optimize three latent variables in an alternating manner.
We minimize one variable while the others are fixed. The optimization (\ref{eq:energy}) is carried out in turn for the three variables.
The L1 optimization is approximated using iterative reweighted least square (IRLS)~\cite{scales1988robust}.
The optimization procedure converges in small number of iterations~$(<10)$.

\begin{figure*}[t]
	\vspace*{0.15in}
	\centering
	\subfloat[]{\includegraphics[width=0.265\textwidth]{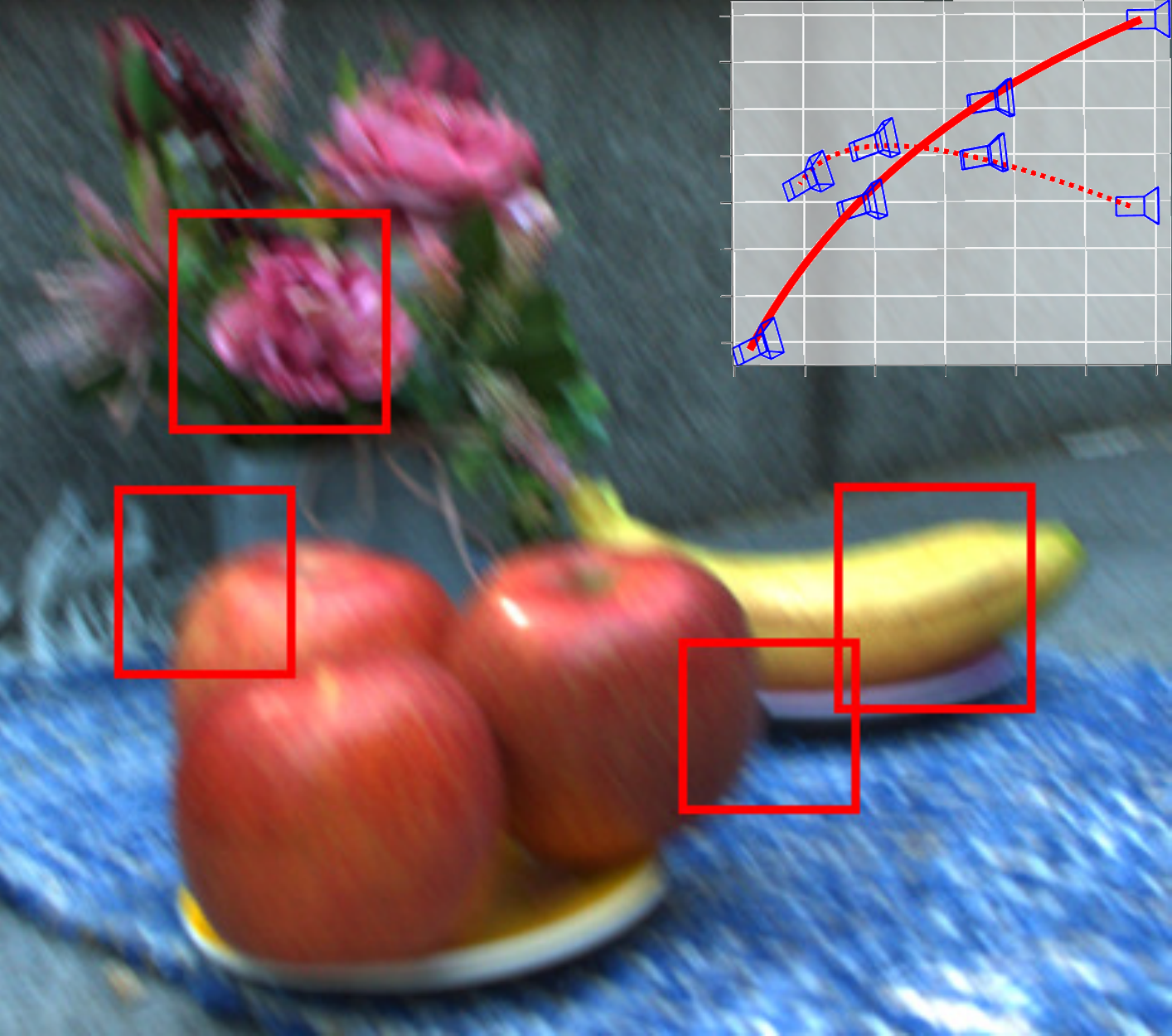}}\hspace*{-0.05in}
	\subfloat[]{\includegraphics[width=0.25\textwidth]{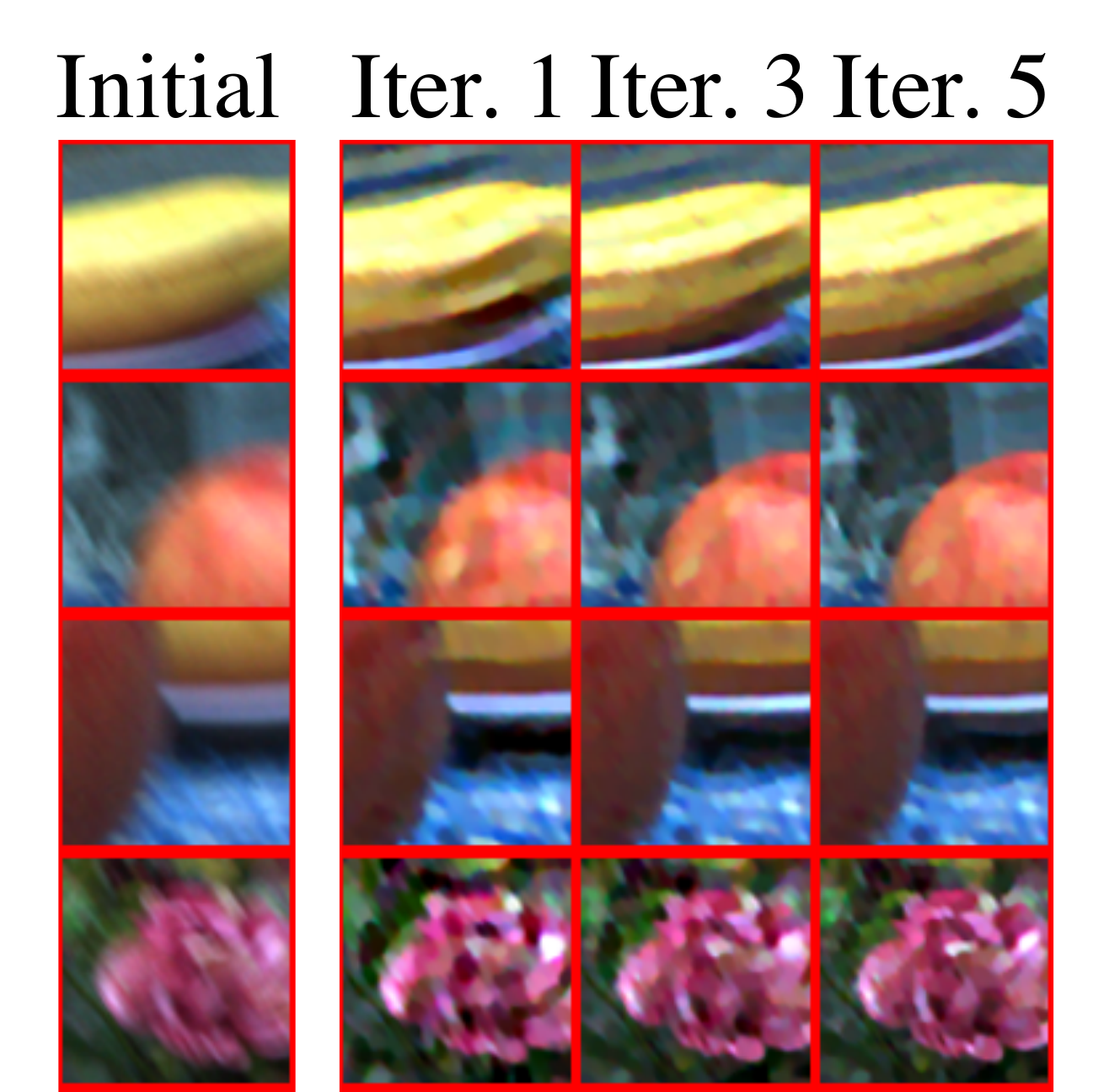}}\hspace*{-0.05in}
	\subfloat[]{\includegraphics[width=0.265\textwidth]{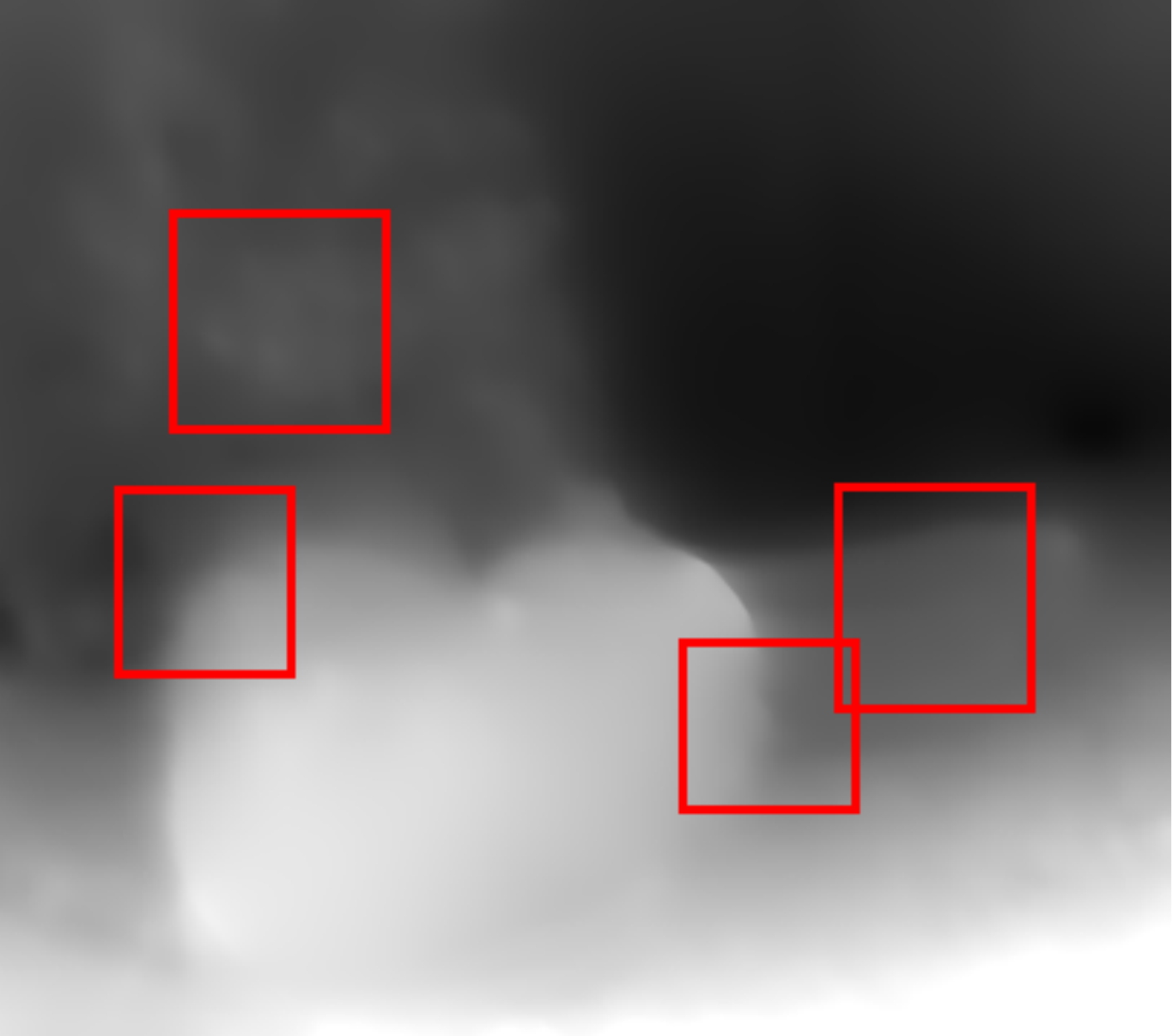}}\hspace*{-0.05in}
	\subfloat[]{\includegraphics[width=0.25\textwidth]{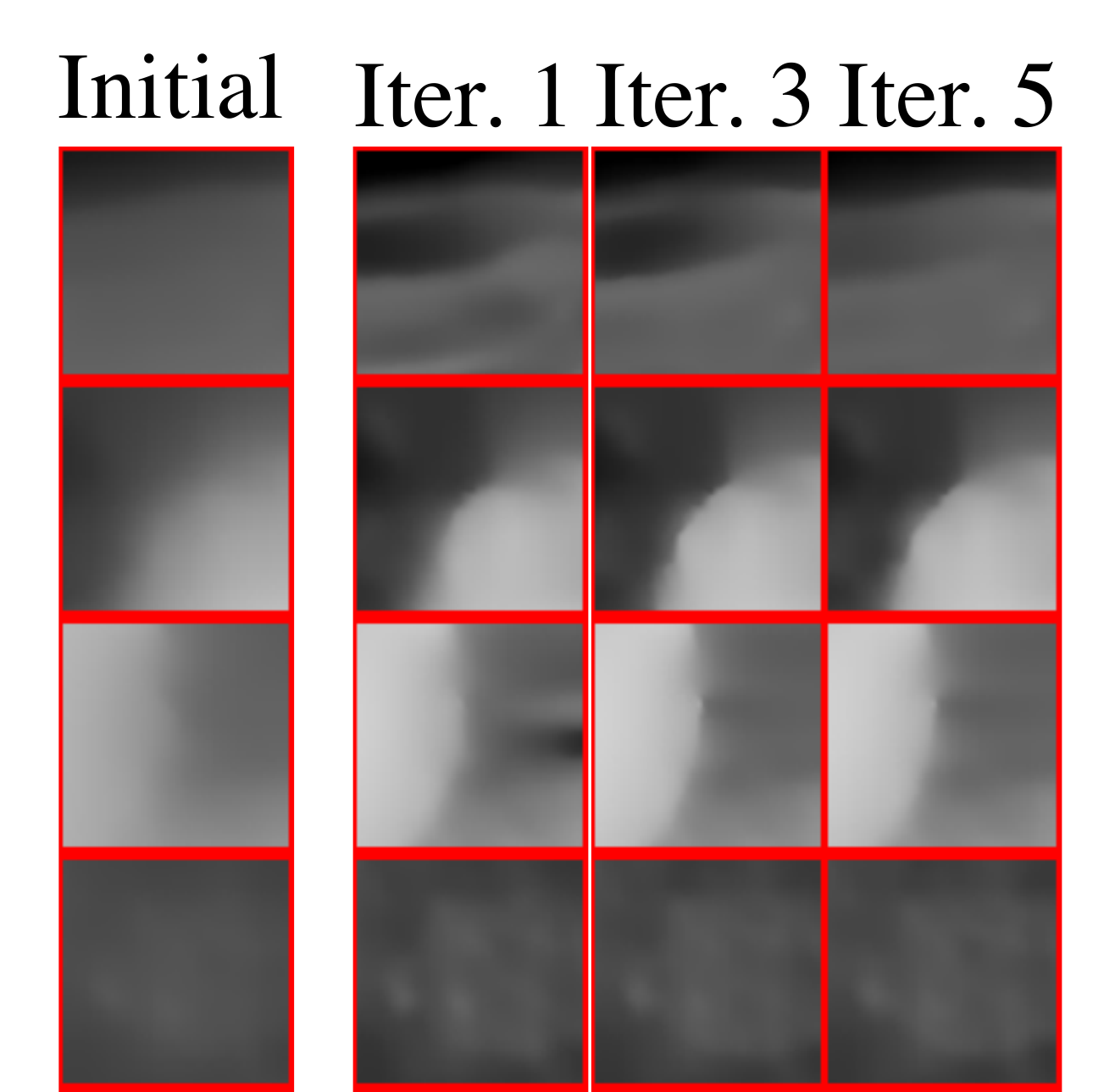}}
	\vspace*{-0.05in}
\caption{Example of the iterative joint estimation. The proposed method converges in small number of iteration.
(a)$\sim$(b) Input blurred image and deblurring results by iteration.
(c)$\sim$(d) Initial blurred depth map and depth estimation results by iteration.
}
	\label{fig:comp2}
	\label{fig3}	
\end{figure*}

An example of the iterative optimization is illustrated in Figure~\ref{fig3} which shows the benefit of the iterative joint estimation of sharp depth map and latent image.
The initial depth map from the blurred light field is blurry as shown in Figure~\ref{fig3}(c).
However, both depth maps and latent images get sharper as the iteration continues as shown in Figure~\ref{fig3}(d).

\section{Joint Estimation of Latent Image, Camera Motion, and Depth Map}
\label{sec4}

\subsection{Update of the Latent Image}
\label{sec:4.1}

The proposed algorithm first updates the latent image $I_{t_r}(\mathbf{x},\mathbf{c})$.
In our data term, the blur operator (\ref{eq:blurOp}) is simplified as the linear matrix multiplication, if $D_{t_r}(\mathbf{x},\mathbf{c})$ and $\mathrm{P}^{\mathbf{c}}_{t_0}$ remain fixed.
Updating the latent image is equivalent to minimizing (\ref{eq:energy}) as follows:
\begin{align}
\min_{I^{\mathbf{c}}_{t}}\sum_{\mathbf{u}}\|K^{\mathbf{u}}I^{\mathbf{c}}_{t_r}-B^{\mathbf{u}} \|_1 + \lambda_L\|\nabla I^{\mathbf{c}}_{t_r}\|_2.
\label{eq:imageUpdate}
\end{align}
$I^{\mathbf{c}}_{t_r}$, $B^{\mathbf{u}}\in\mathbb{R}^n$ are vectorized images and ${K}^{\mathbf{u}}\in\mathbb{R}^{n\times n}$ is the blur operator in square matrix form, where $n$ is the number of pixels in the center-view sub-aperture image.
TV regularization serves as a prior to the latent image with clear boundary while eliminating the ringing artifacts.

\subsection{Update of the Camera Pose and Depth Map}
\label{sec:4.2}

Since (\ref{eq:blurOp}) is a non-linear function of $D_{t_r}(\mathbf{x},\mathbf{c})$ and $\mathrm{P}^{\mathbf{c}}_{t_0}$, it is necessary to approximate it in a linear form for efficient computation.
In our approach, the blur operation (\ref{eq:blurOp}) is approximated as a first-order expansion.
Let $D_{0}(\mathbf{x},\mathbf{c})$ and $\mathrm{P}^{\mathbf{c}}_{0}$ denote the initial variables, then (\ref{eq:blurOp}) is approximated as follow:
\begin{align}
\begin{split}
&(\Psi\circ I)(\mathbf{x},\mathbf{u})\\
&= B_0(\mathbf{x},\mathbf{u})+ \textstyle \frac{\partial B_0}{\partial \mathbf{f}}(\frac{\partial\mathbf{f}}{\partial D_{t_r}(\mathbf{x},\mathbf{c})}\Delta D_{t_r}(\mathbf{x},\mathbf{c})+\frac{\partial\mathbf{f}}{\partial \varepsilon_{t_0}}\varepsilon_{t_0}),
\end{split}
\end{align}
where
\begin{align}
B_0(\mathbf{x},\mathbf{u})=(\Psi\circ I)(\mathbf{x},\mathbf{u})\vert_{D_{t_r}(\mathbf{x},\mathbf{c})=D_{0}(\mathbf{x},\mathbf{c}),\mathrm{P}^{\mathbf{c}}_{t_0}=\mathrm{P}^{\mathbf{c}}_{0}},
\end{align}
Note that $\mathbf{f}$ is motion flow generated by warping function, and $\varepsilon_{t_0}$ denotes six-dimensional vector on $\mathfrak{se}(3)$.
The partial derivatives related to $D_{t_r}(\mathbf{x},\mathbf{c})$ and $\varepsilon_{t_0}$ are given in~\cite{blanco2010tutorial}.

Once it is approximated using $\Delta D_{t_r}(\mathbf{x},\mathbf{c})$ and $\varepsilon_{t_0}$, (\ref{eq:energy}) can be optimized using IRLS.
The resulting $\Delta D_{t_r}(\mathbf{x},\mathbf{c})$ and $\varepsilon_{t_0}$ are incremental values for the current $D_{t_r}(\mathbf{x},\mathbf{c})$ and $\mathrm{P}^{\mathbf{c}}_{t_0}$, respectively.
They are updated as follows:
\begin{align}
\begin{split}
&D_{t_r}(\mathbf{x},\mathbf{c})=D_{t_r}(\mathbf{x},\mathbf{c})+\Delta D_{t_r}(\mathbf{x},\mathbf{c}),\\
&\mathrm{P}^{\mathbf{c}}_{t_0} = \exp (\varepsilon_{t_0})\mathrm{P}^{\mathbf{c}}_{t_0},
\end{split}
\end{align}
where $\mathrm{P}^{\mathbf{c}}_{t_0}$ is updated through the exponential mapping of the motion vector $\varepsilon_{t_0}$.

Figure~\ref{fig3} shows the initial latent variables and final outputs.
After joint estimation, both the latent image and the depth map become clean and sharp.

The proposed blur formulation and joint estimation approach are not limited to the light field but can also be applied to images obtained from a stereo camera or general multi-view camera system. The only property of the light field we use is that sub-aperture images are equivalent to the images obtained from multi-view camera array.
Note that the proposed method is not limited to a simple motion path model (moving smoothly in $\mathfrak{se}(3)$ space).
More complex parametric curves, such as the B\'{e}zier curve used in the prior work~\cite{srinivasan2017light}, can be directly applied only if they are differentiable.\\

\begin{figure}[t]
	\vspace*{0.15in}
	\centering
	\subfloat[]{\includegraphics[width=0.248\columnwidth]{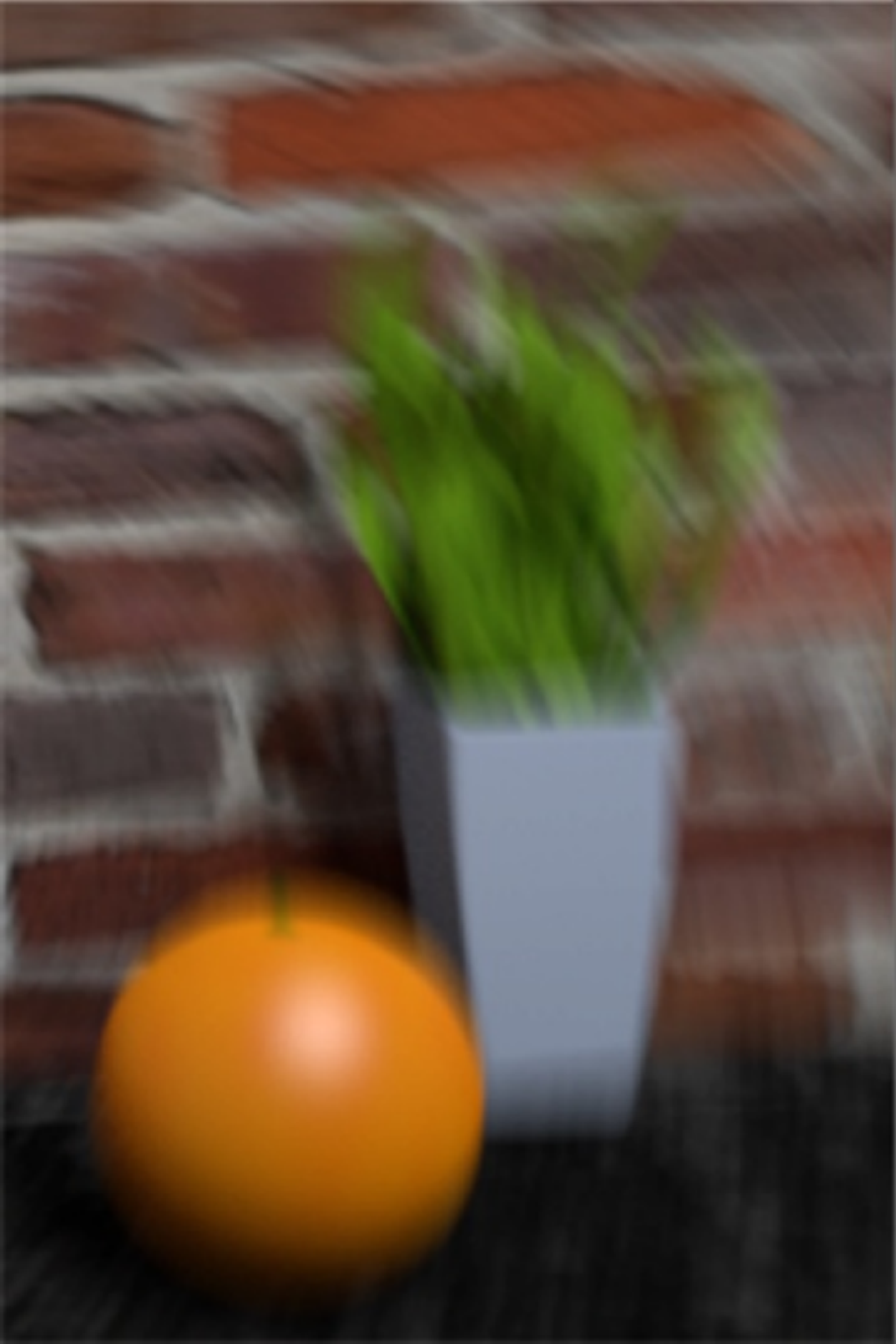}}\hspace*{-0.01in}
	\subfloat[]{\includegraphics[width=0.248\columnwidth]{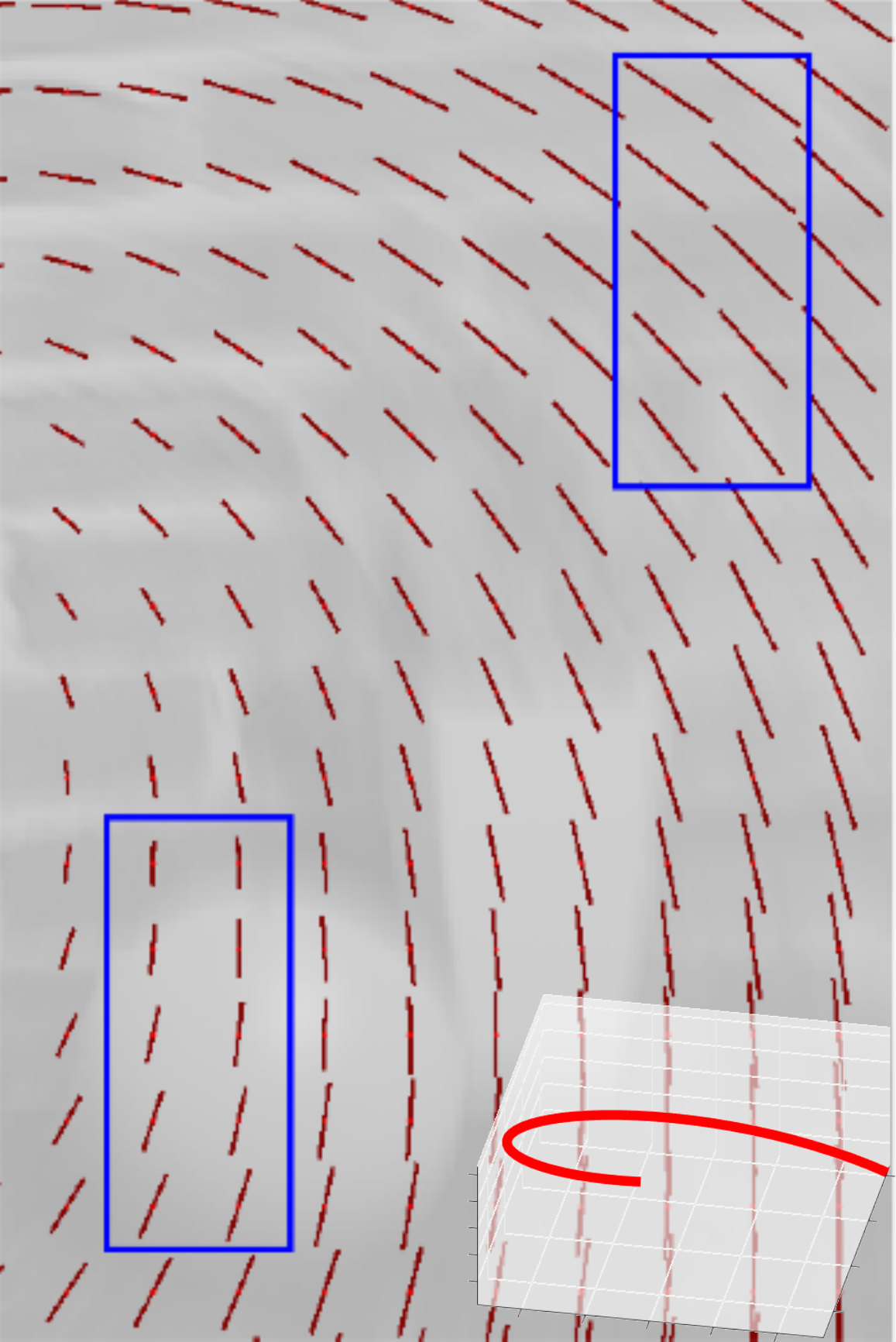}}\hspace*{-0.01in}
	\subfloat[]{\includegraphics[width=0.248\columnwidth]{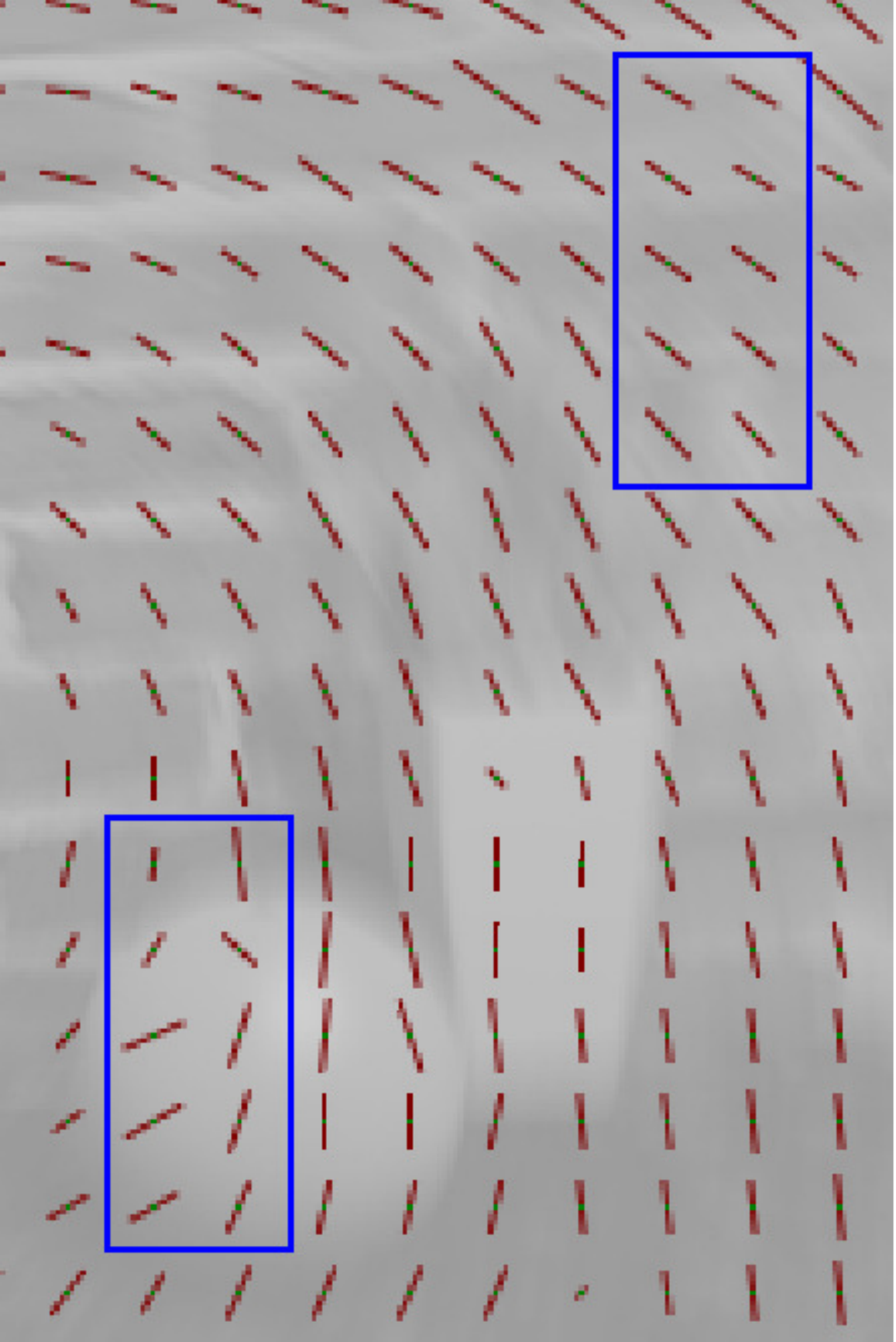}}\hspace*{-0.01in}
	\subfloat[]{\includegraphics[width=0.248\columnwidth]{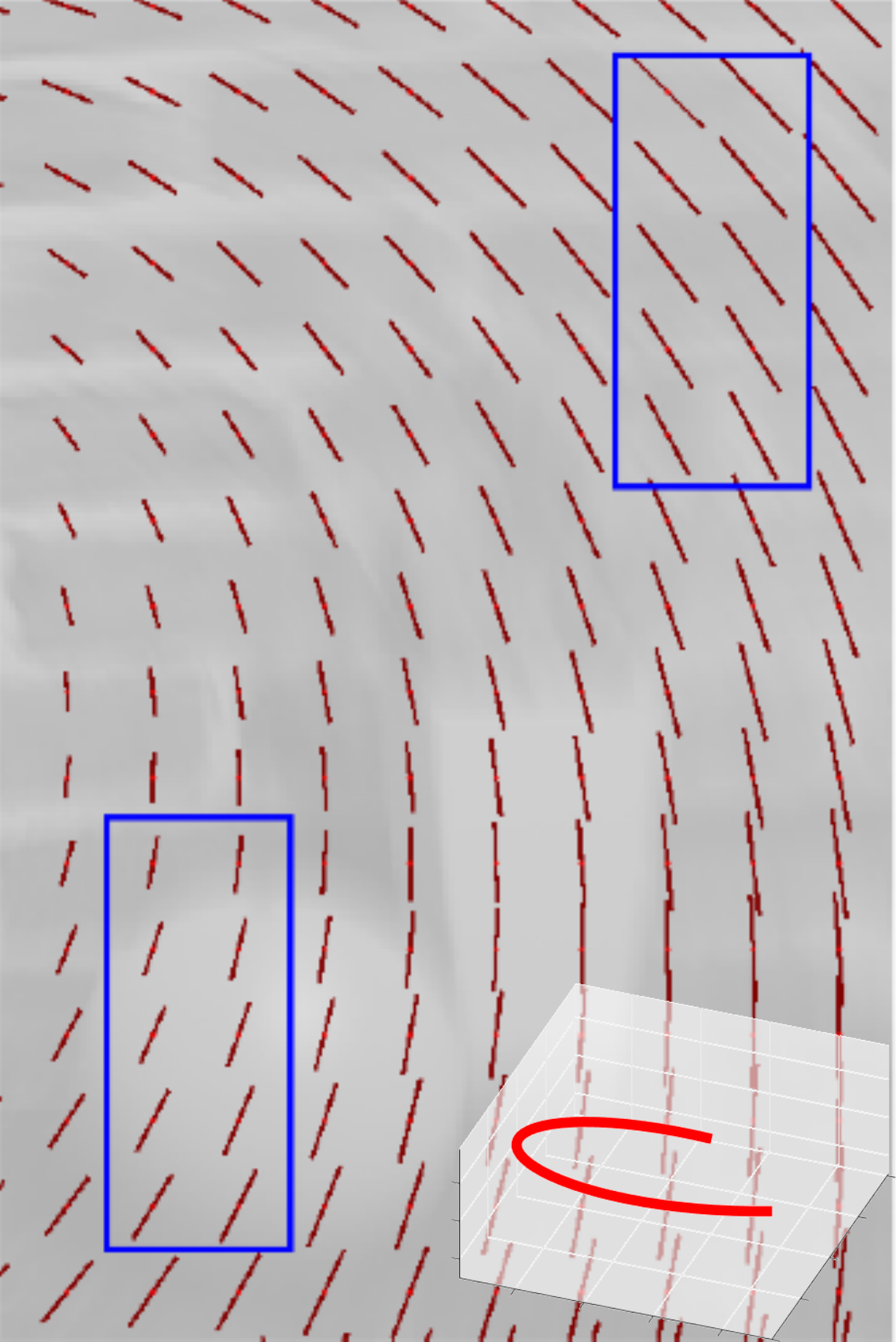}}\hspace*{-0.01in}
	\vspace*{-0.05in}
\caption{
Example of camera motion initialization on a synthetic light field.
(a) Blurred input light field.
(b) Ground truth motion flow.
(c) Sun~et al.~\cite{sun2015learning} (EPE = 3.05),
(d) Proposed initial motion (EPE = 0.95).
In (b) and (d), the linear blur kernels are approximated only using the end points of camera motion for the visualization.}
	\label{fig4}	
\end{figure}
\subsection{Initialization}
\label{sec:initialization}

Since deblurring is a highly ill-posed problem and the optimization is done in a greedy and iterative fashion, it is important to start with good initial values.
First, we initialize the depth map using the input sub-aperture images of the light field.
It is assumed that the camera is not moving and (\ref{eq:energy}) is minimized to obtain the initial $D_{t_r}(\mathbf{x},\mathbf{c})$.
Minimizing (\ref{eq:energy}) becomes a simple multi-view stereo matching problem.
Figure~\ref{fig:comp2}(c) shows the initial depth map which exhibits  fattened object boundary.

Camera motion $\mathrm{P}^{\mathbf{c}}_{t_0}$ is initialized from the local linear blur kernels and initial scene depth.
We first estimate the local linear blur kernel of $B(\mathbf{x},{\mathbf{c}})$ using~\cite{sun2015learning}.
Then, we fit the pixel coordinates moved by the linear kernel and the re-projected coordinates by the warping function as follows:
\begin{align}
\min_{\mathrm{P}^{\mathbf{c}}_{t_0}}\sum^N_{i=1}\|w_{t_0}(\mathbf{x}_i,\mathbf{c})-l(\mathbf{x}_i)\|^2_2,
\label{eq:fitting}
\end{align}
where $\mathbf{x}_i$ is the sampled pixel position and $l(\mathbf{x}_i)$ is the point that $\mathbf{x}_i$ is moved by the end point of the linear kernel.
$\mathrm{P}^{\mathbf{c}}_{t_0}$ is obtained by fitting $\mathbf{x}_i$ moved by $w_{t_0}(\cdot,\mathbf{c})$ and $l(\cdot)$.
$\mathrm{P}^{\mathbf{c}}_{t_0}$ is the only variable of $w_{t_0}(\cdot,\mathbf{c})$ since the scene depth is fixed to the initial depth map.
In our implementation, RANSAC is used to find the camera motion that best describes the pixel-wise linear kernels.
$N$ is the number of random samples, which is fixed to 4.

Figure~\ref{fig4} shows an example of camera motion initialization.
It is shown that \cite{sun2015learning} underestimates the size of the motion (upper blue rectangle) and produces noisy motion where the texture is insufficient (lower blue rectangle).

	
\section{Experimental Results}

The proposed algorithm is implemented using Matlab on an Intel i7 7770K @ 4.2GHz with 16GB RAM and is evaluated for both synthetic and real light fields.
Synthetic light field is generated using \textit{Blender}~\cite{blender} for qualitative as well as quantitative evaluation.
It includes 6~types of camera motion for 3~different scenes in which each light field has $7\times7$ angular structure of 480$\times$360 sub-aperture images.
Synthetic blur is simulated by moving the camera array over a sequence of frames $(\geq40)$ and then by averaging the individual frames.
On the other hand, real light field data is captured using Lytro Illum camera which generates $7\times7$ angular structure of 552$\times$383 sub-aperture images.
We generate the sub-aperture images from light field using the toolbox~\cite{bok2014geometric} which provides the relative camera poses between sub-aperture images.
Light fields are blurred by moving camera quickly under arbitrary motion, while the scene remains static.
In our implementation, we fixed most of the parameters except $\lambda_D$ such that $ \lambda_u=15, \lambda_c = 1, \lambda_L=5$.
$\lambda_D$ is set to a larger value for a real light field ($\lambda_D=400$) than for synthetic data ($\lambda_D=20$).

For quantitative evaluation of deblurring, we use both peak signal to noise ratio~(PSNR) and structural similarity~(SSIM).
Note that PSNR and SSIM are measured by the maximum (best) ones among individual PSNR and SSIM values computed between the deblurred image and the ground truth images (along the motion path) as adopted in~\cite{kohler2012recording}.
For comparison with light field depth estimation methods, we use the relative mean absolute error~(L1-rel) defined as
\begin{equation}
\text{L1-rel}(D,\hat{D})=\frac{1}{n}\sum_{i}\frac{|D_{i}-\hat{D}_{i}|}{\hat{D}_{i}},
\end{equation}
which computes the relative error of the estimated depth $\hat{D}$ to the ground truth depth $D$.
The accuracy of camera motion estimation is measured by the average end point error (EPE) to the end point of ground truth blur kernels.
In our evaluation, we compute the EPE by generating an end point of blur kernel using the estimated camera motion and ground truth depth.
We compare the performance of the proposed algorithm to linear blur kernel methods that directly computes the EPE between the ground truth and their pixel-wise blur kernel.

\begin{figure*}[ht]
	\begin{center}		
		\vspace*{0.15in}
		\includegraphics[width=0.24\textwidth]{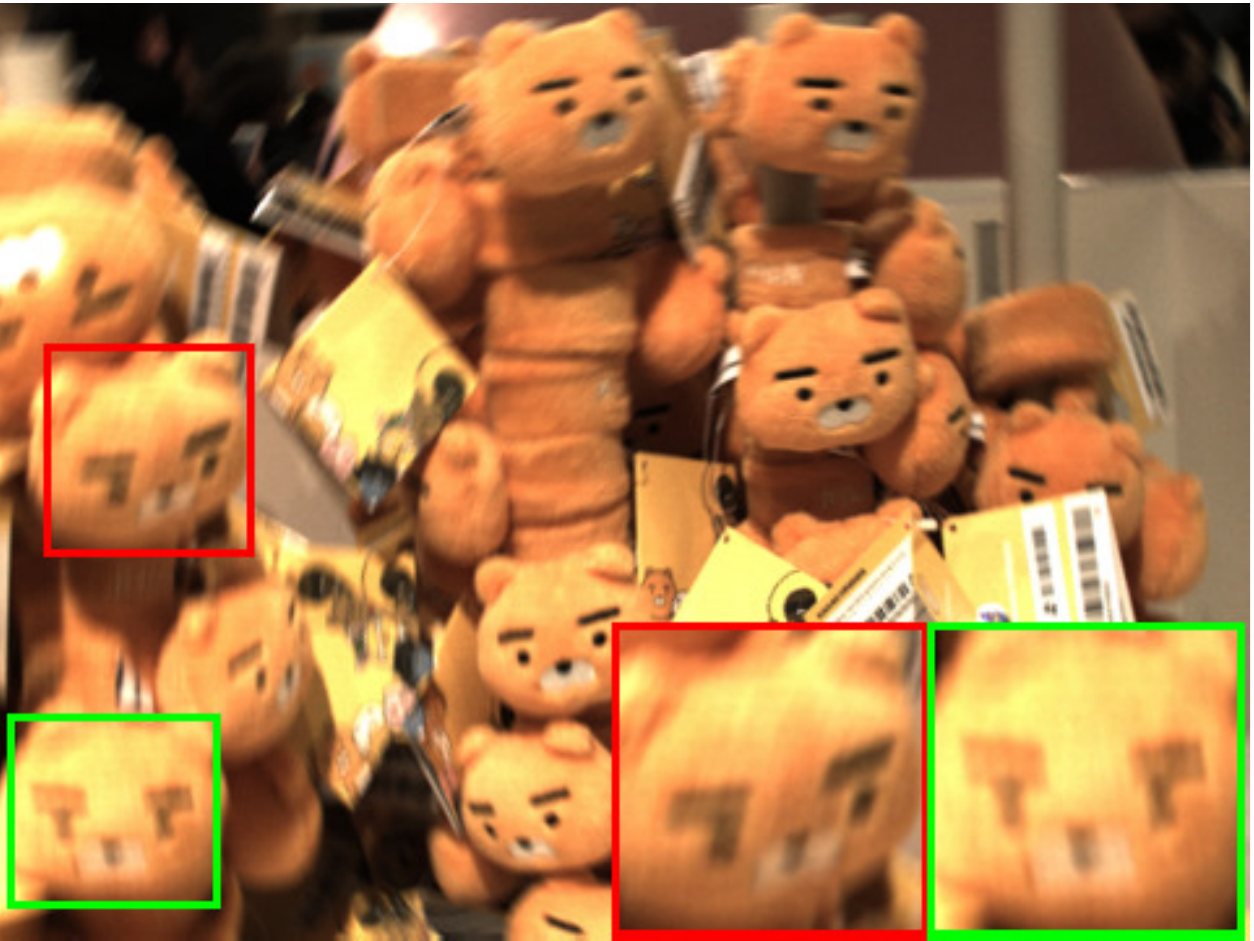}\hspace*{0.01in}
		\includegraphics[width=0.24\textwidth]{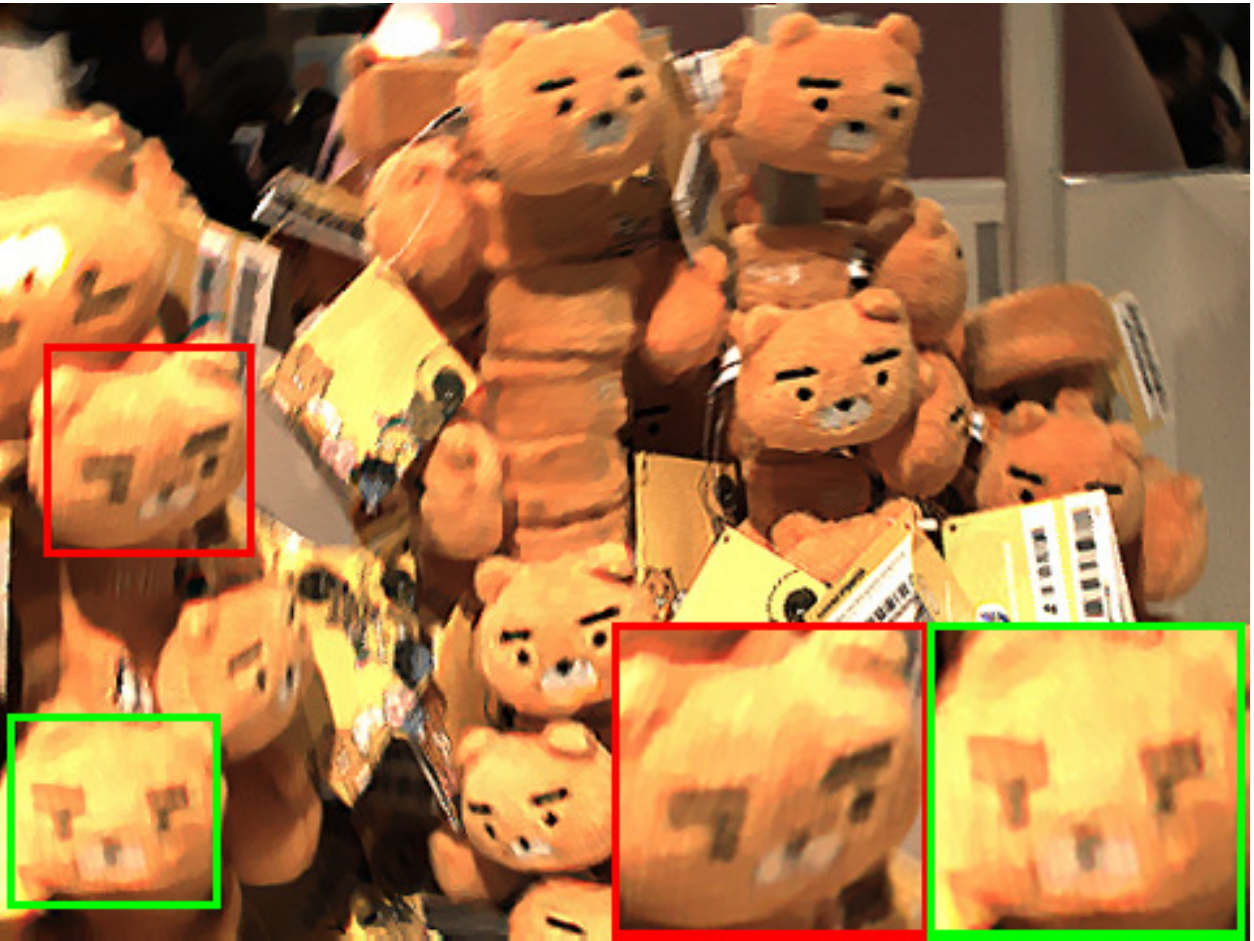}\hspace*{0.01in} 
		\includegraphics[width=0.24\textwidth]{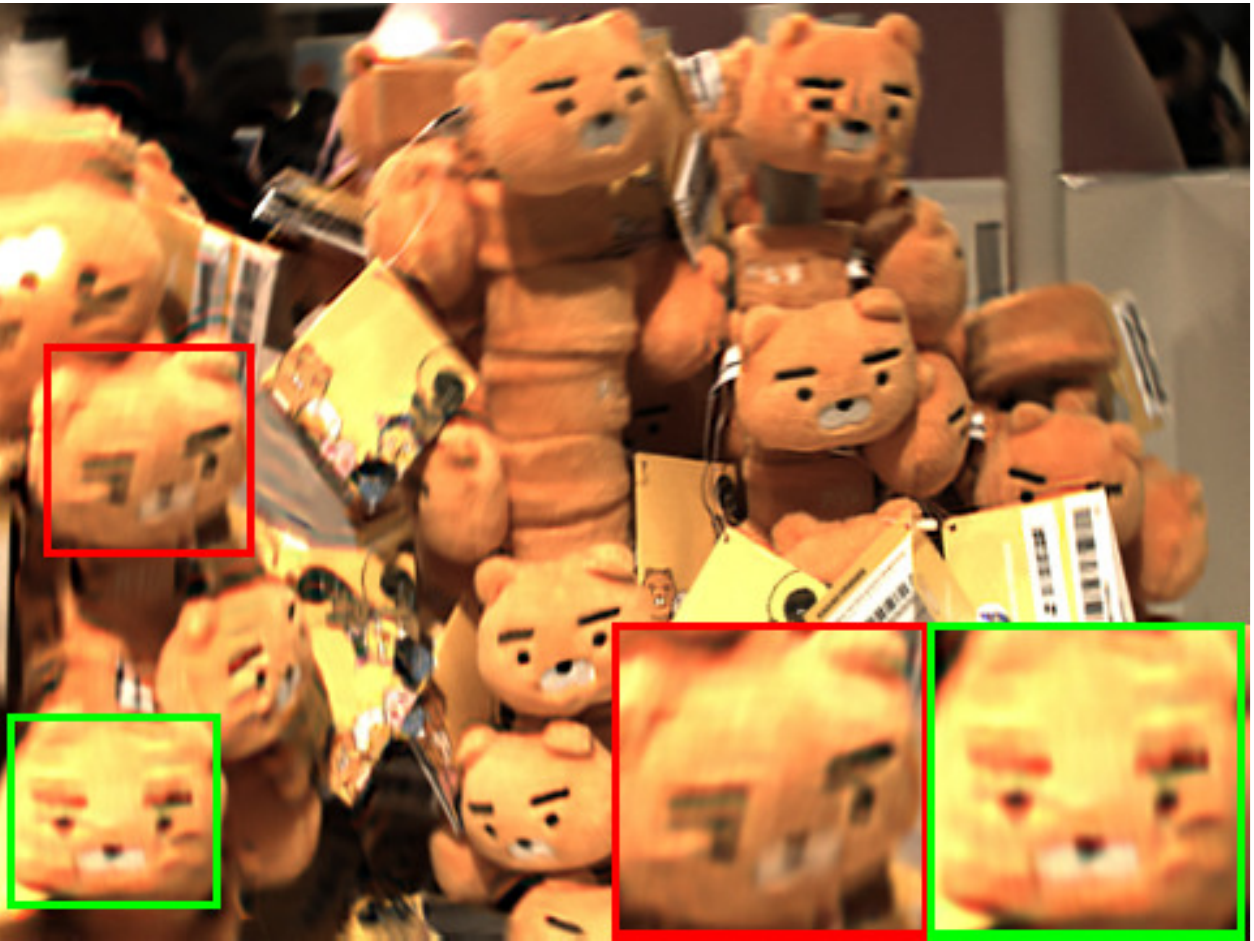}\hspace*{0.01in}
		\includegraphics[width=0.24\textwidth]{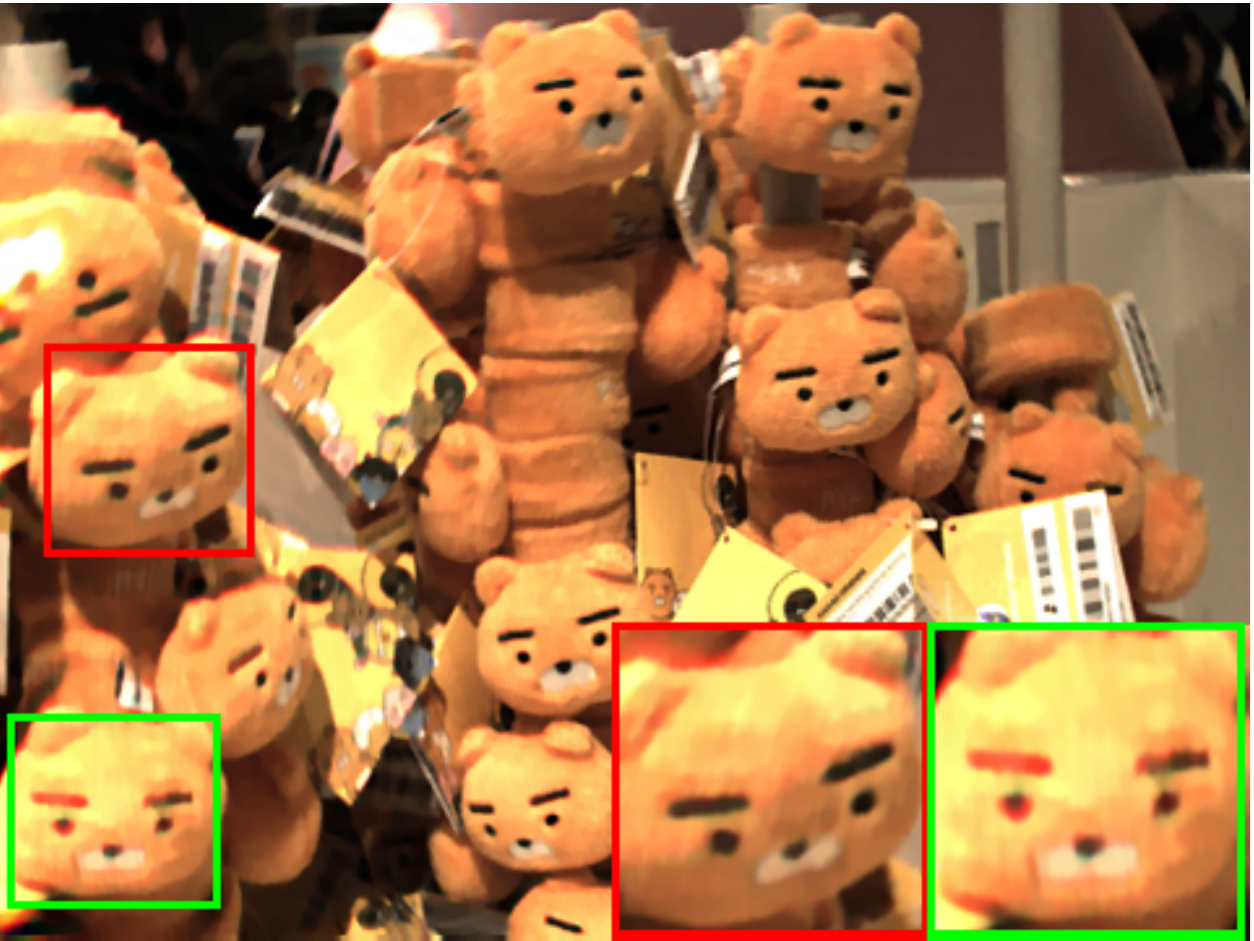}
		\vspace*{0.01in}
		
		\includegraphics[width=0.24\textwidth]{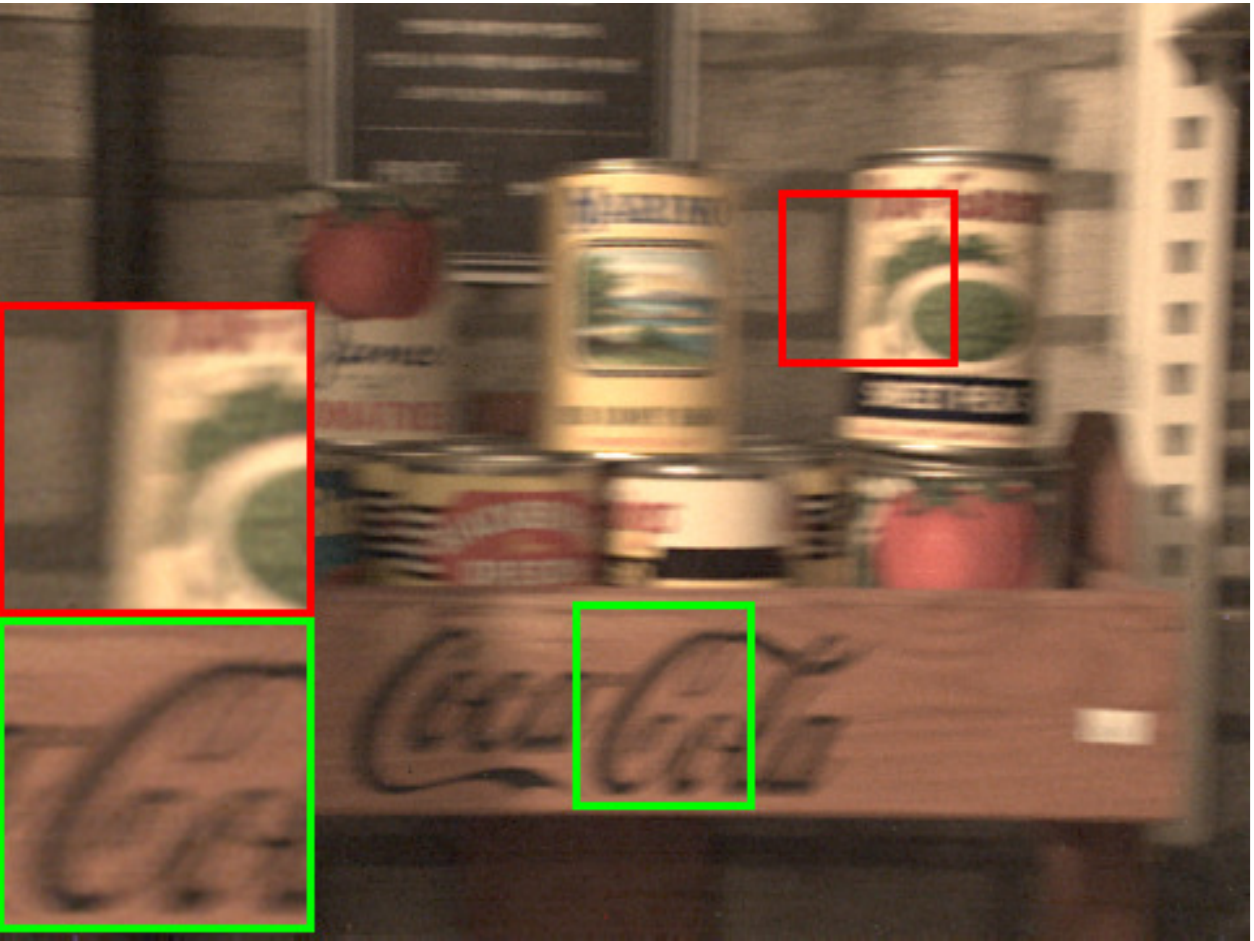}\hspace*{0.01in}
		\includegraphics[width=0.24\textwidth]{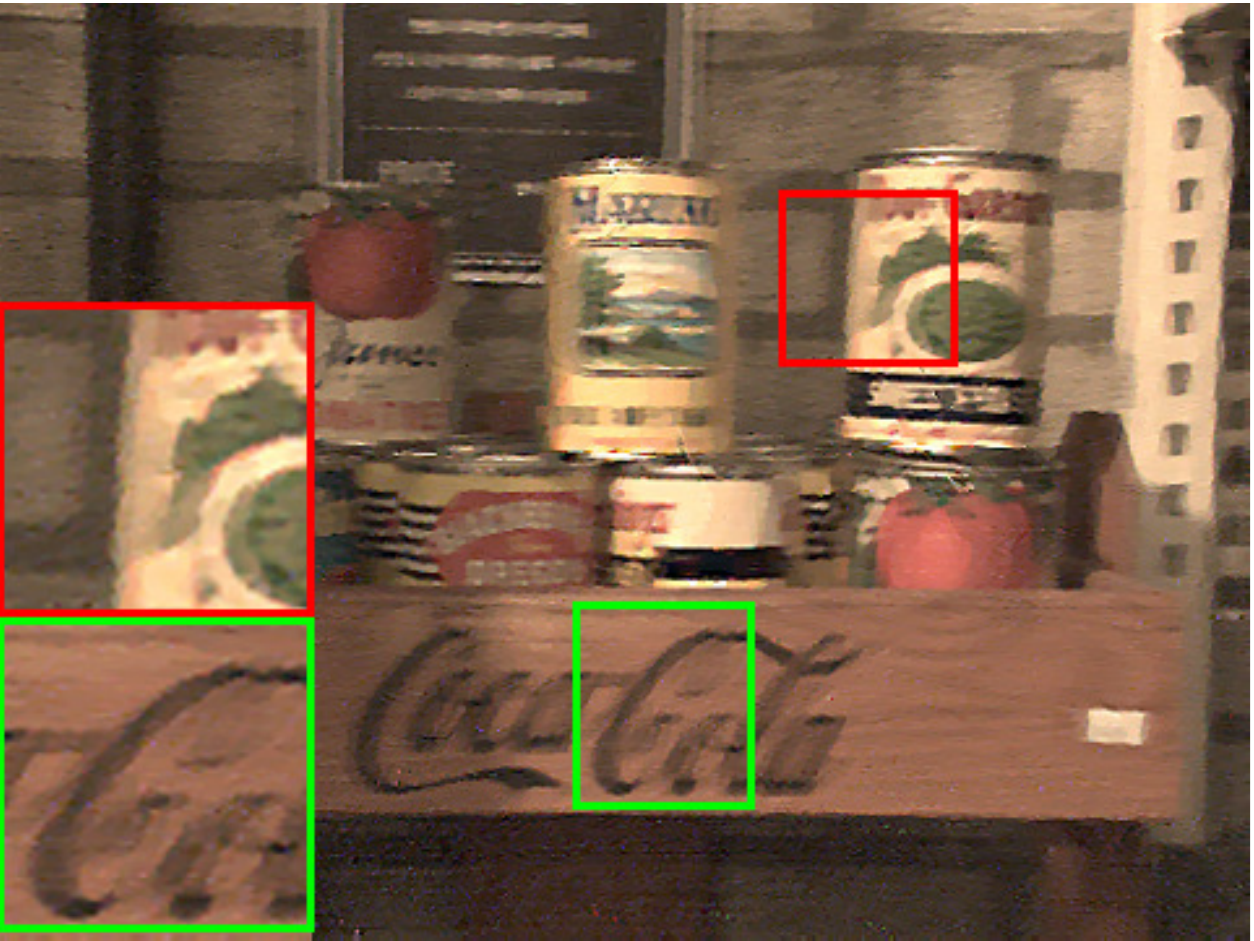}\hspace*{0.01in}
		\includegraphics[width=0.24\textwidth]{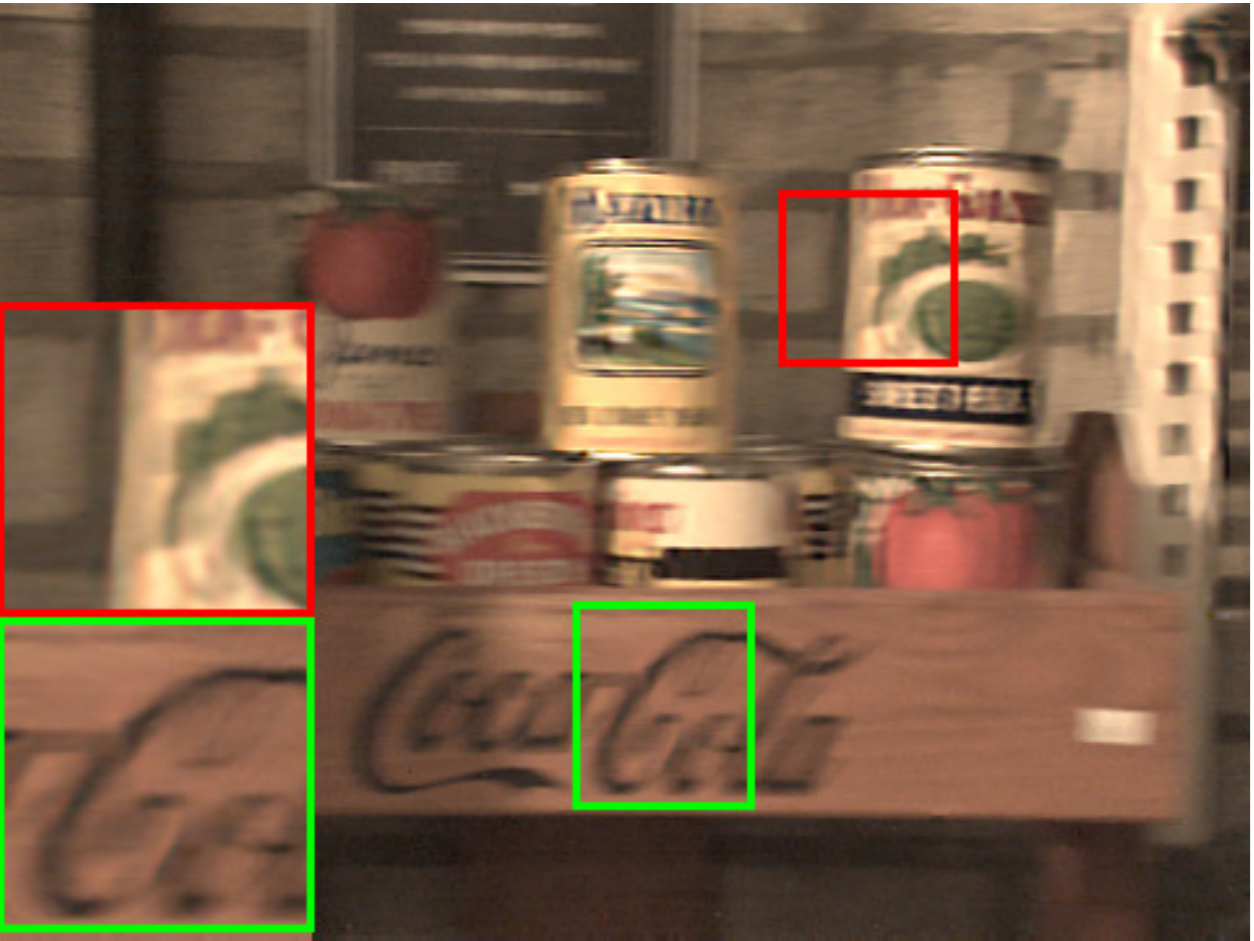}\hspace*{0.01in}
		\includegraphics[width=0.24\textwidth]{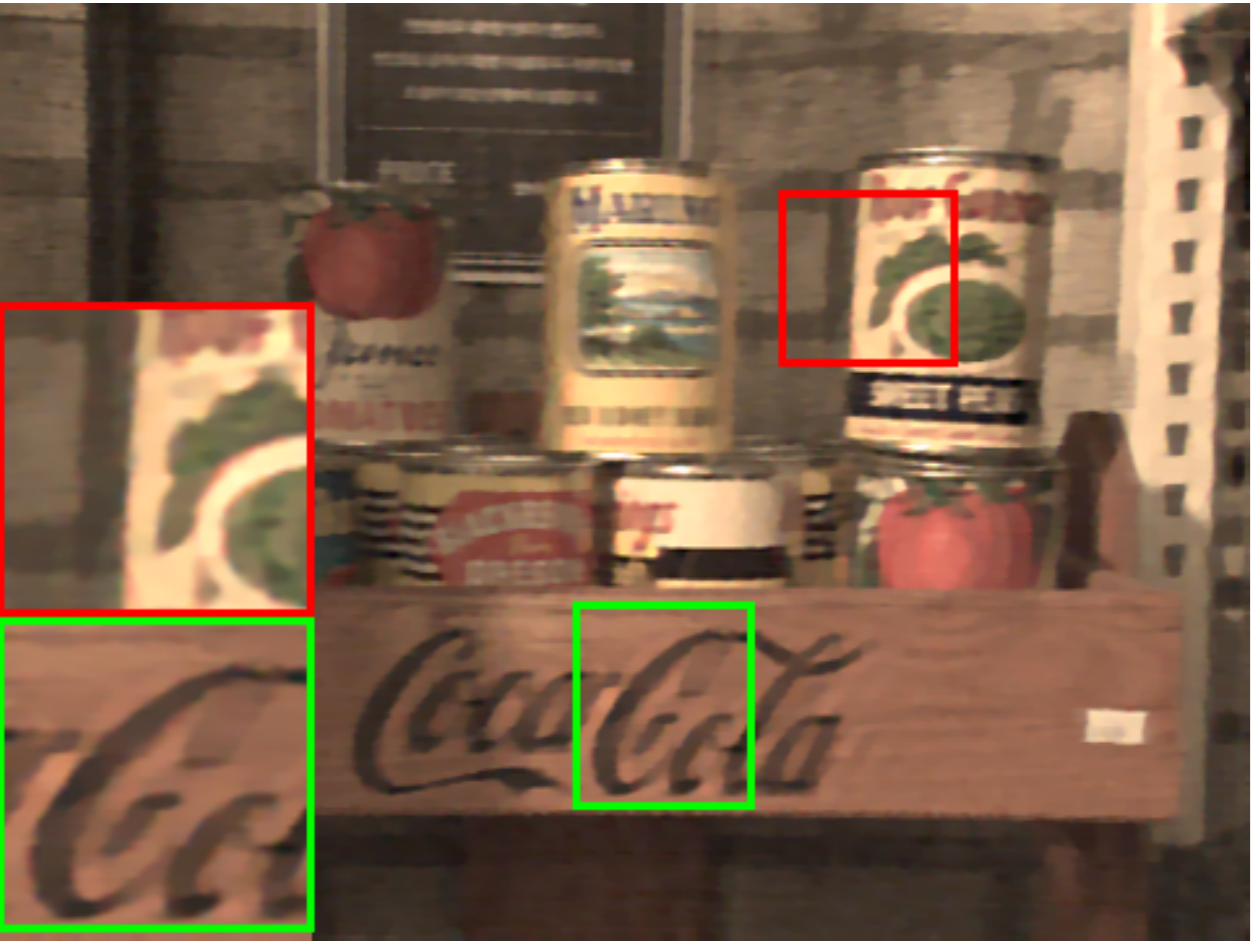}
		\vspace*{-0.12in}
	
		\subfloat[]{\includegraphics[width=0.24\textwidth]{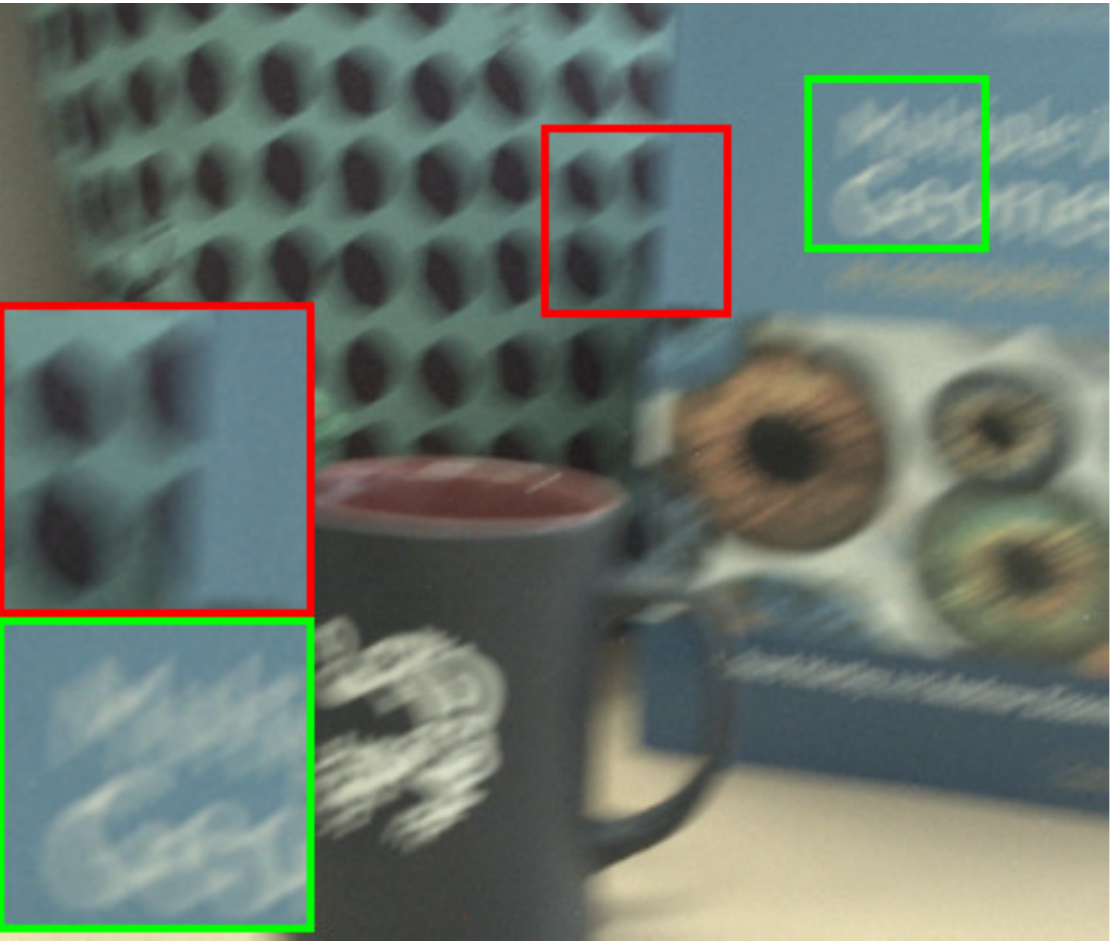}}\hspace*{0.01in}
		\subfloat[]{\includegraphics[width=0.24\textwidth]{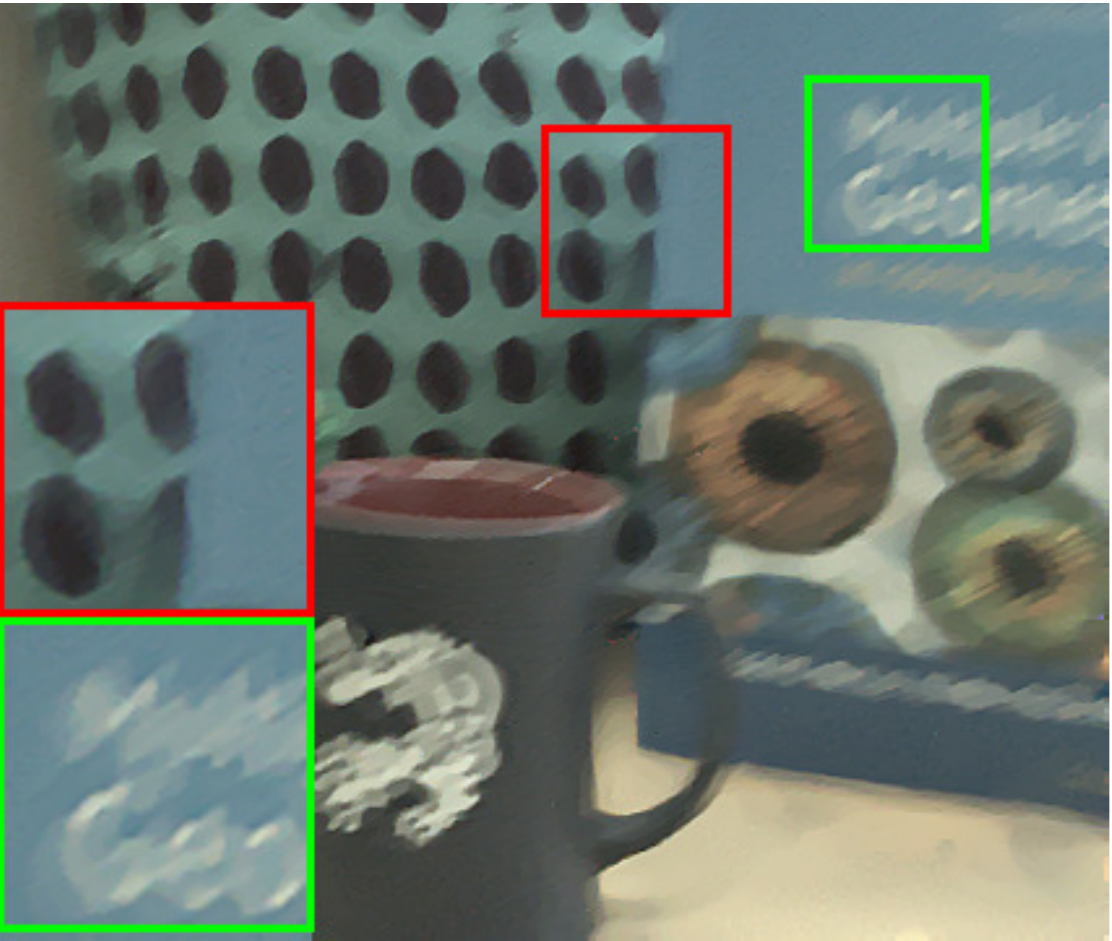}}\hspace*{0.01in}
		\subfloat[]{\includegraphics[width=0.24\textwidth]{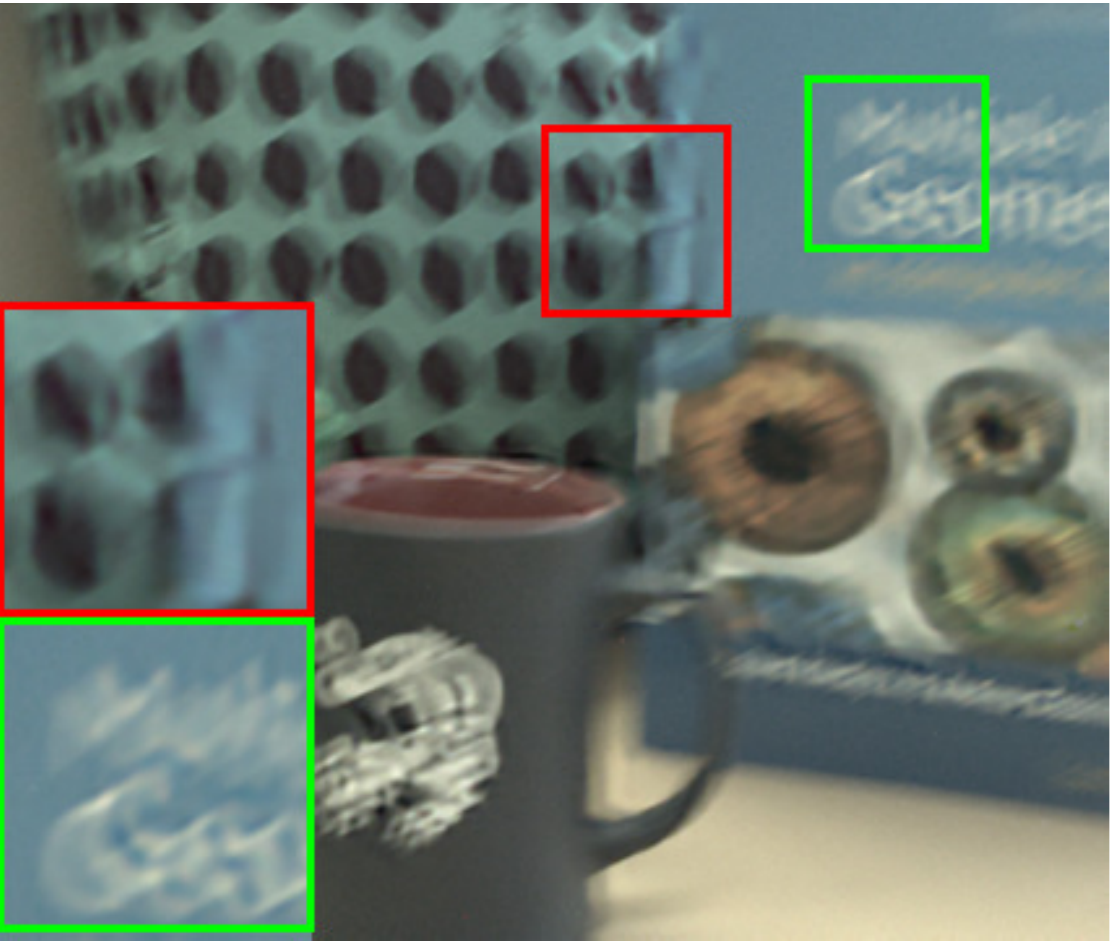}}\hspace*{0.01in}
		\subfloat[]{\includegraphics[width=0.24\textwidth]{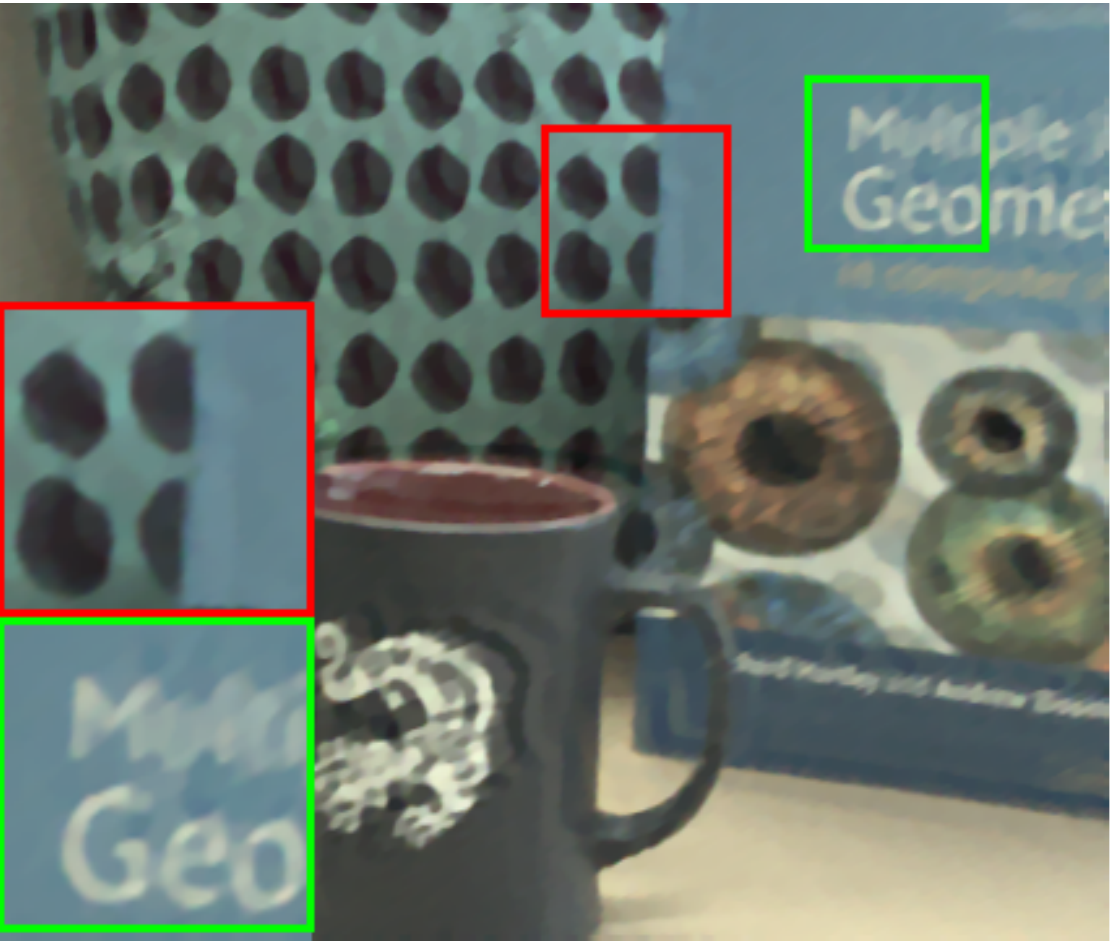}}
		
	\end{center}	
	\vspace*{-0.23in}
	\caption{Deblurring result for real light field dataset with comparison to local linear blur kernel deblurring methods. (a) Blurred input image. (b) Result of Kim and Lee~\cite{kim2014segmentation}. (c) Sun~et al.~\cite{sun2015learning}. (d) Proposed algorithm.}
	\label{qual:deblur_real1}
\end{figure*}
\begin{figure*}[!t]
	\begin{center}	
		\includegraphics[width=0.24\textwidth]{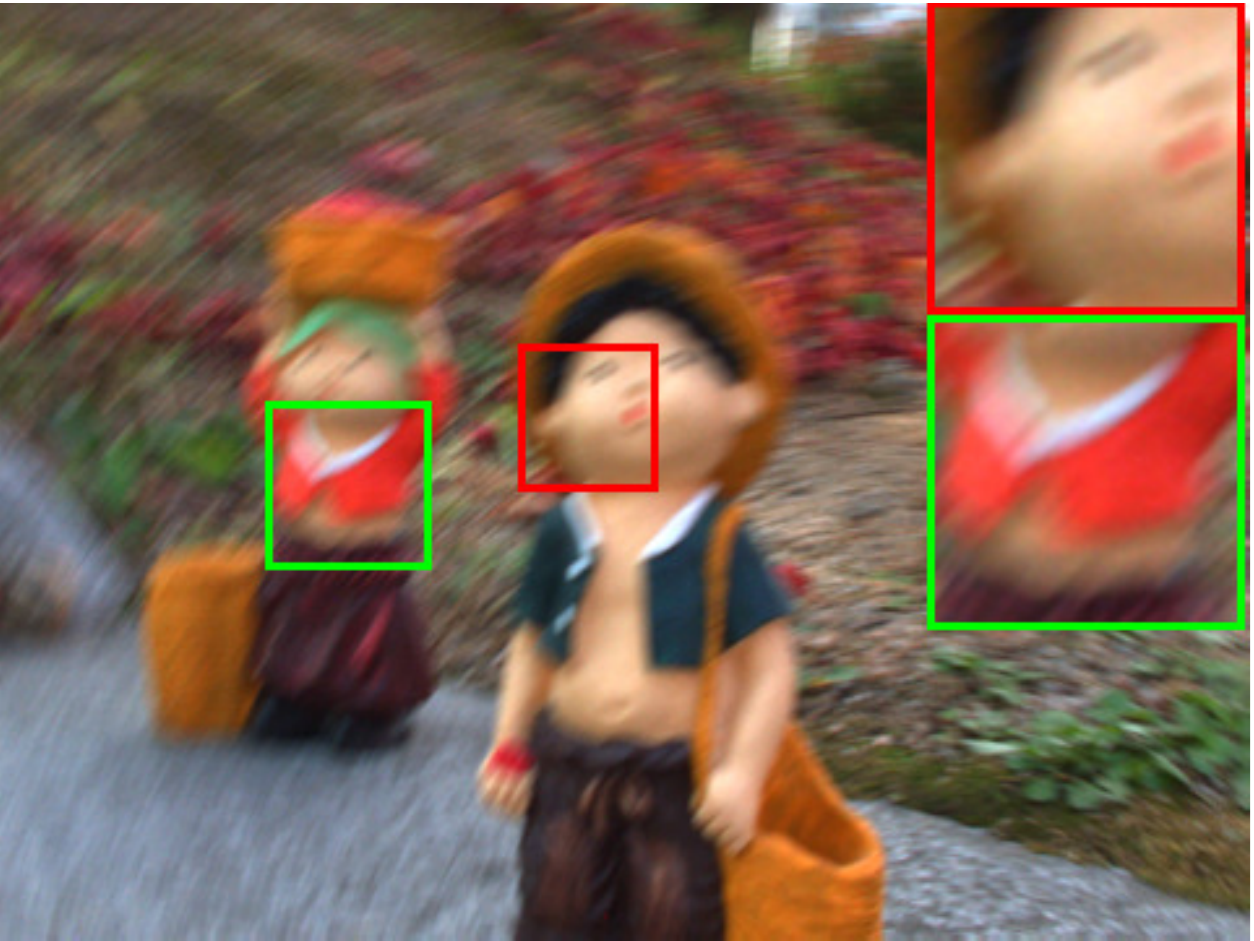}\hspace*{0.01in}
		\includegraphics[width=0.24\textwidth]{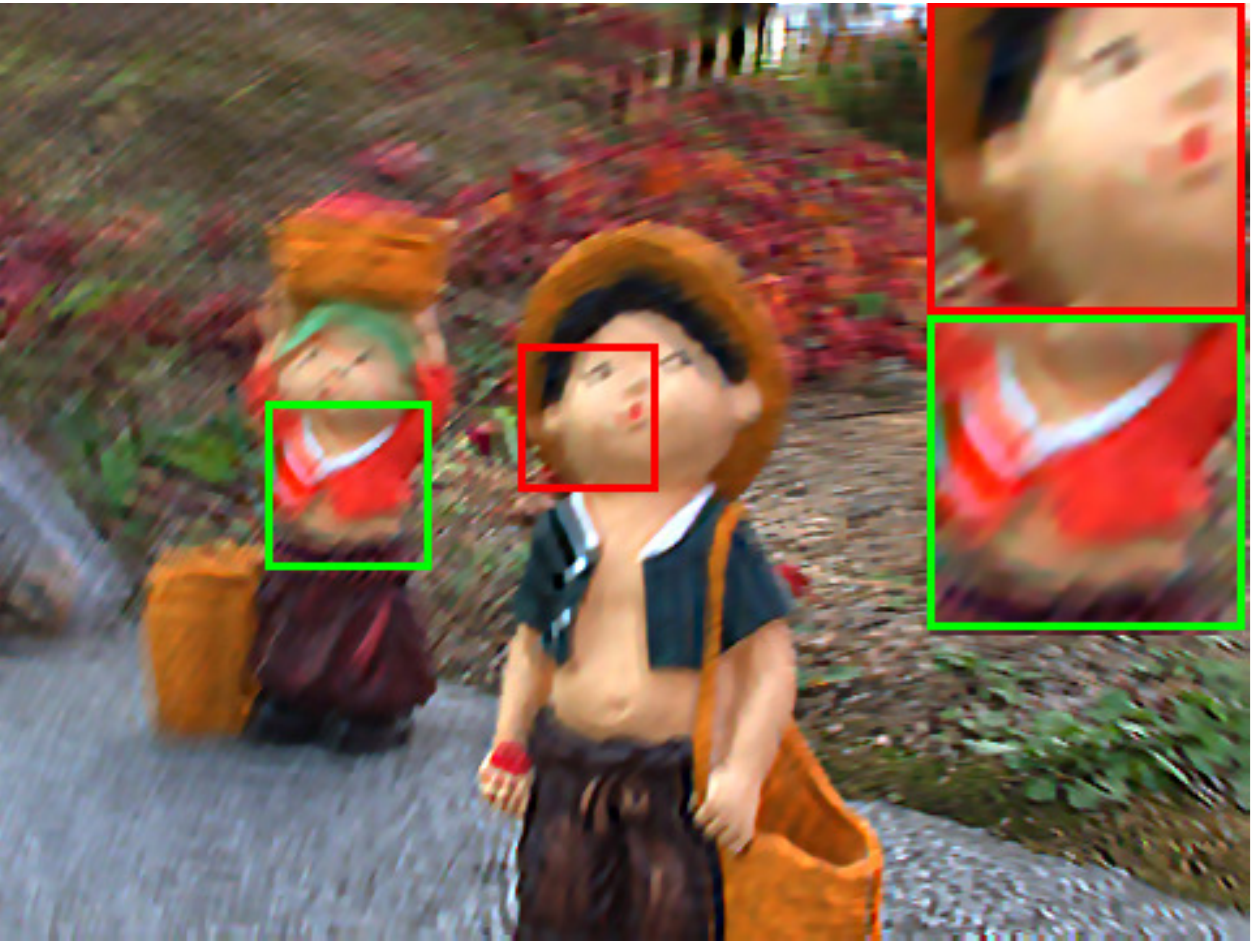}\hspace*{0.01in}
		\includegraphics[width=0.24\textwidth]{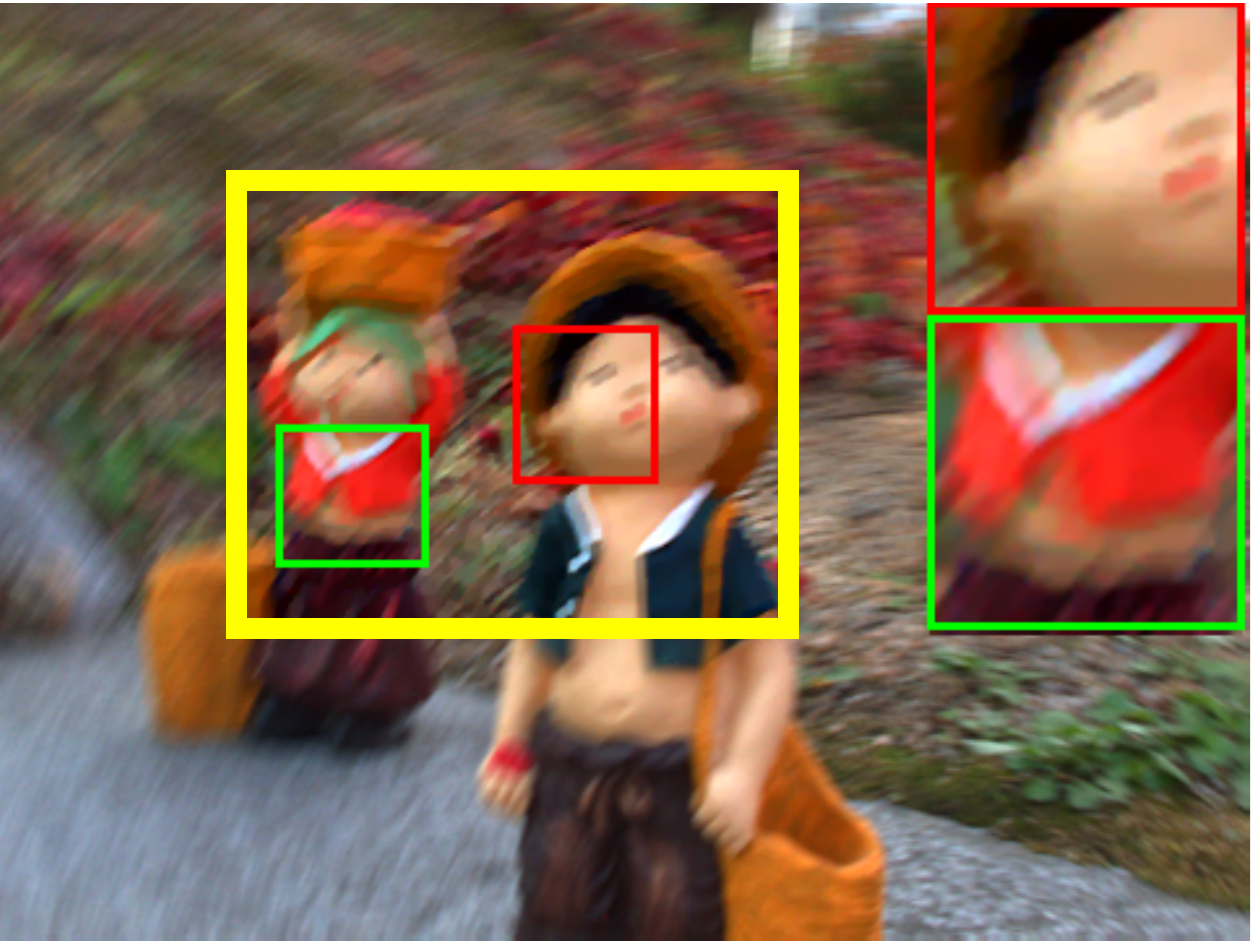}\hspace*{0.01in}
		\includegraphics[width=0.24\textwidth]{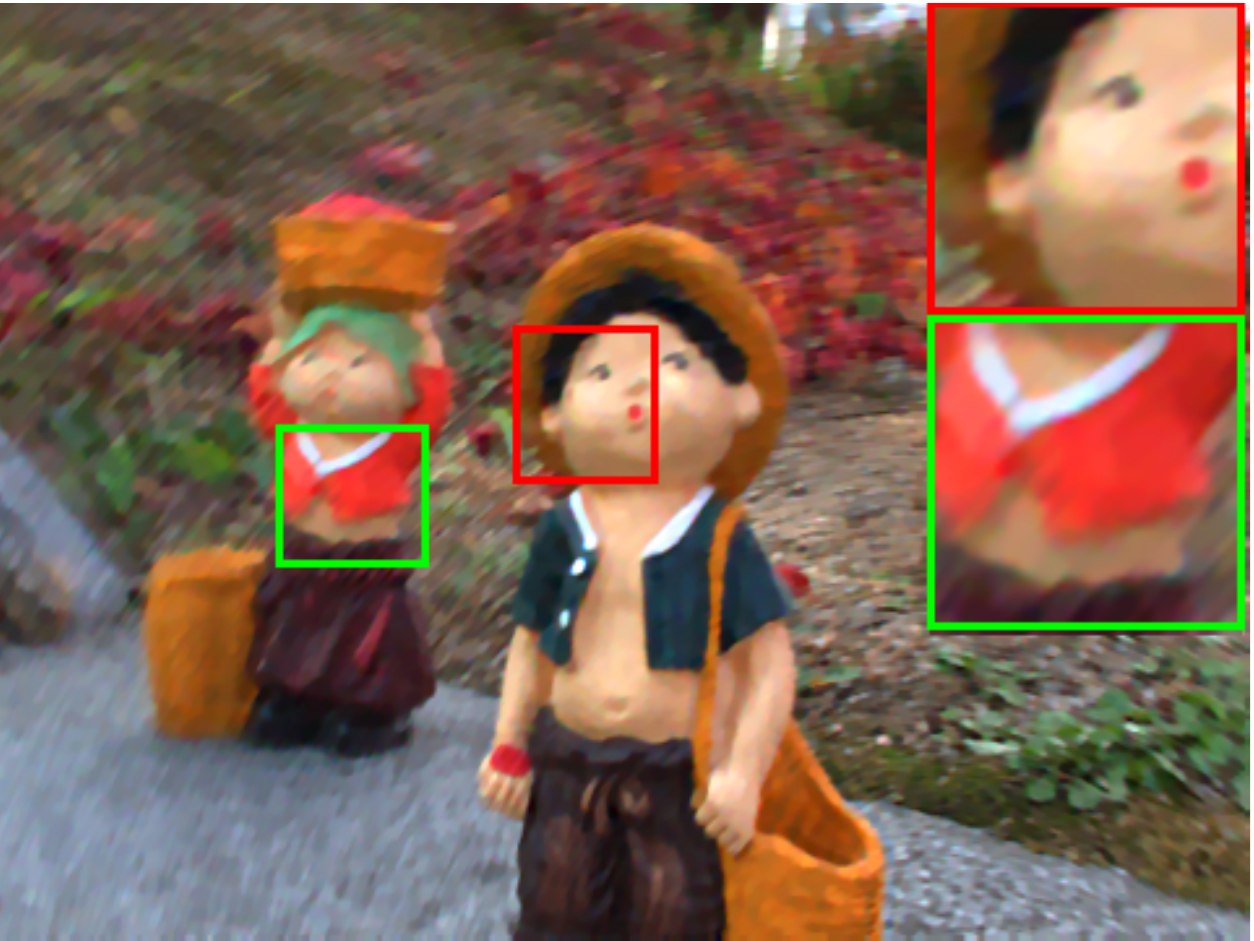}
		\vspace*{0.01in}
		
		\includegraphics[width=0.24\textwidth]{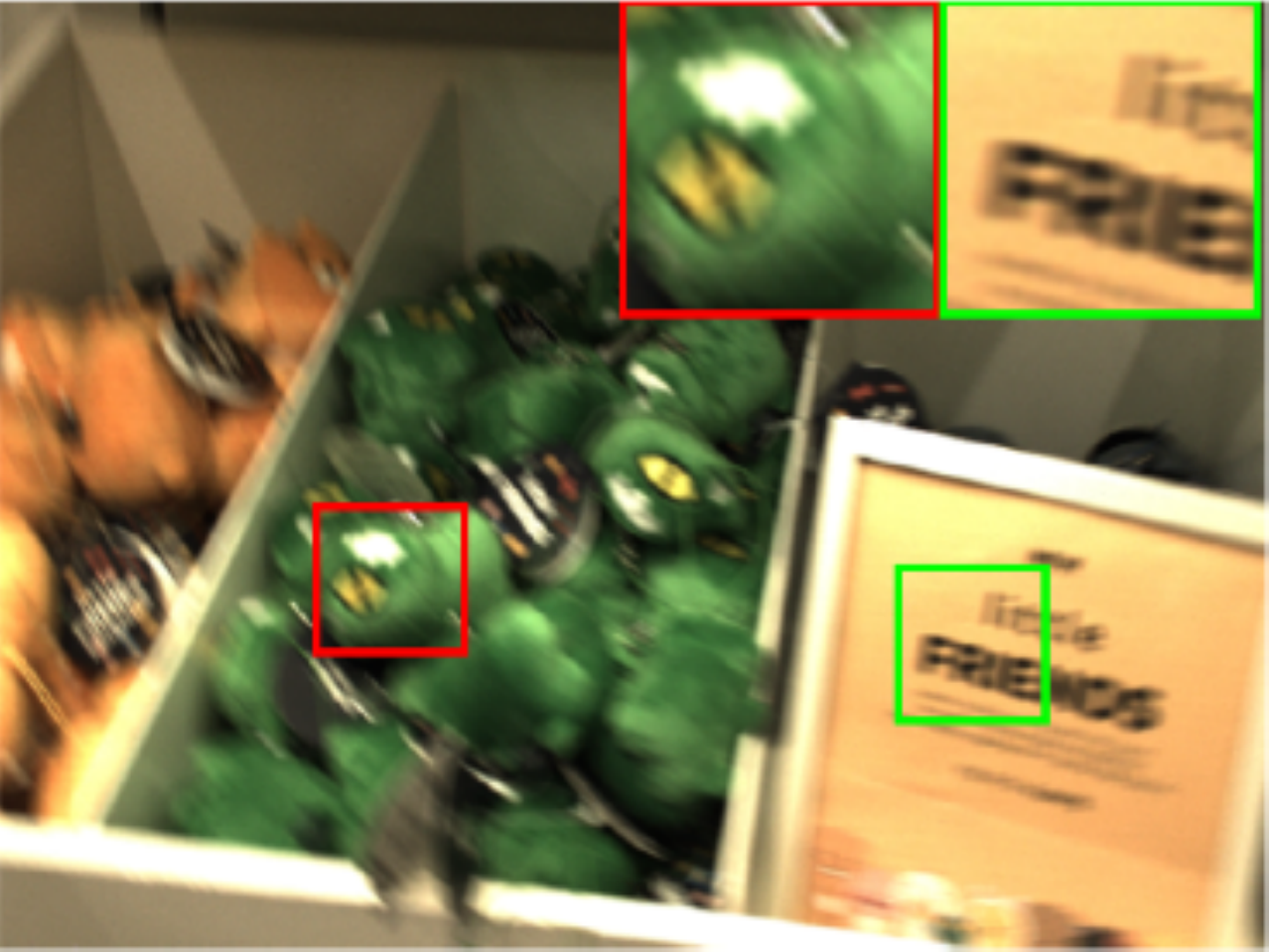}\hspace*{0.01in}
		\includegraphics[width=0.24\textwidth]{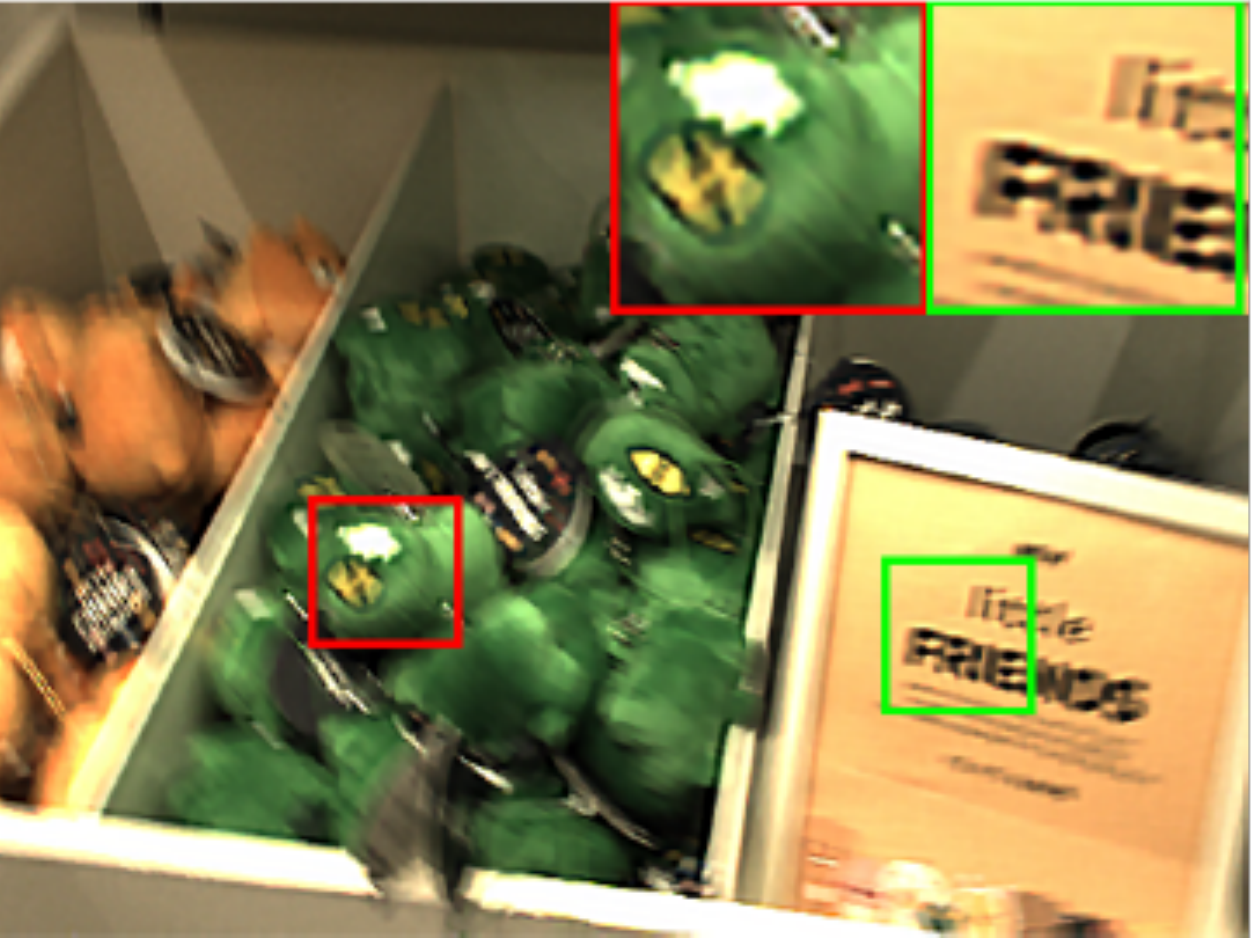}\hspace*{0.01in}
		\includegraphics[width=0.24\textwidth]{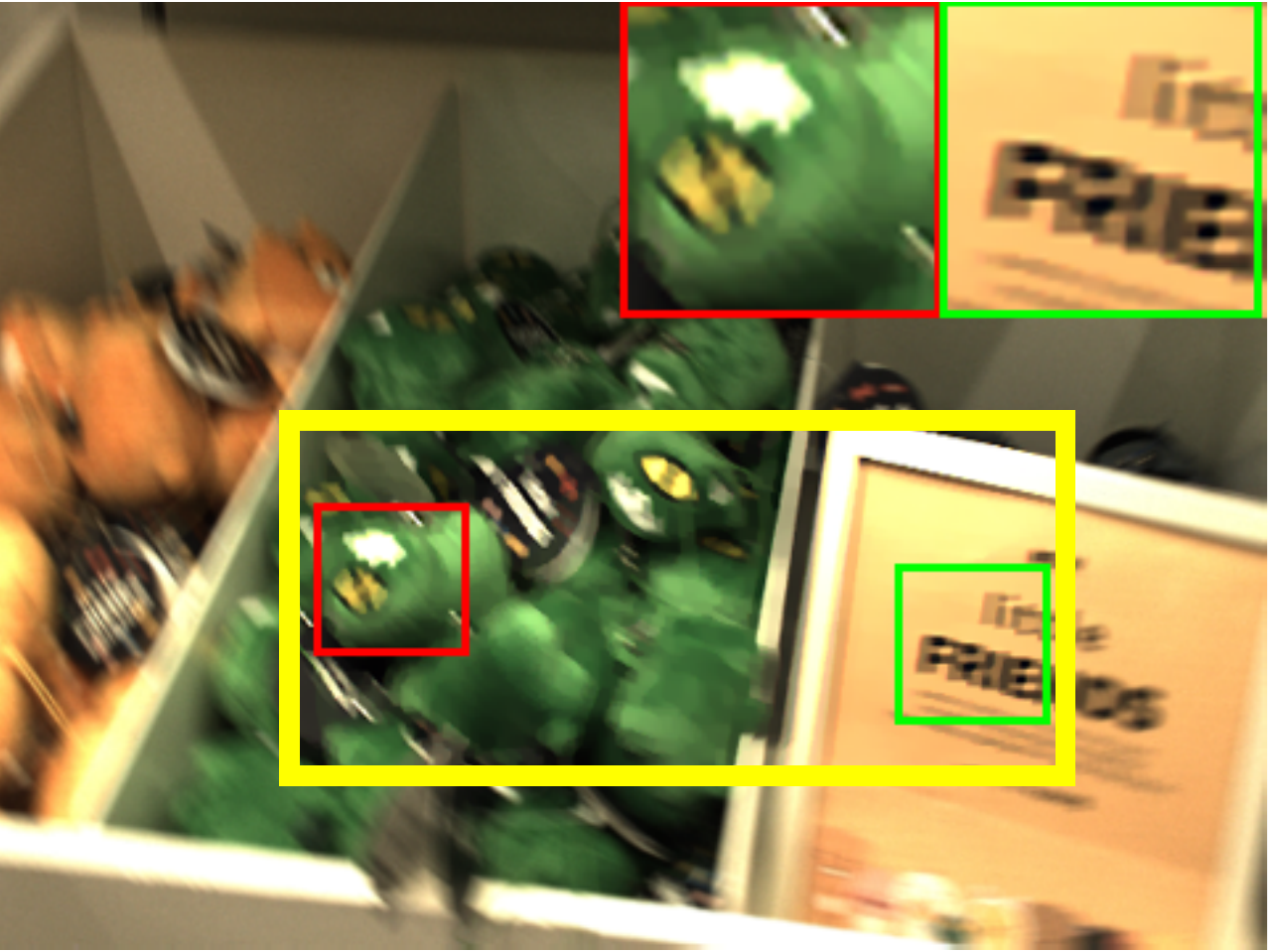}\hspace*{0.01in}
		\includegraphics[width=0.24\textwidth]{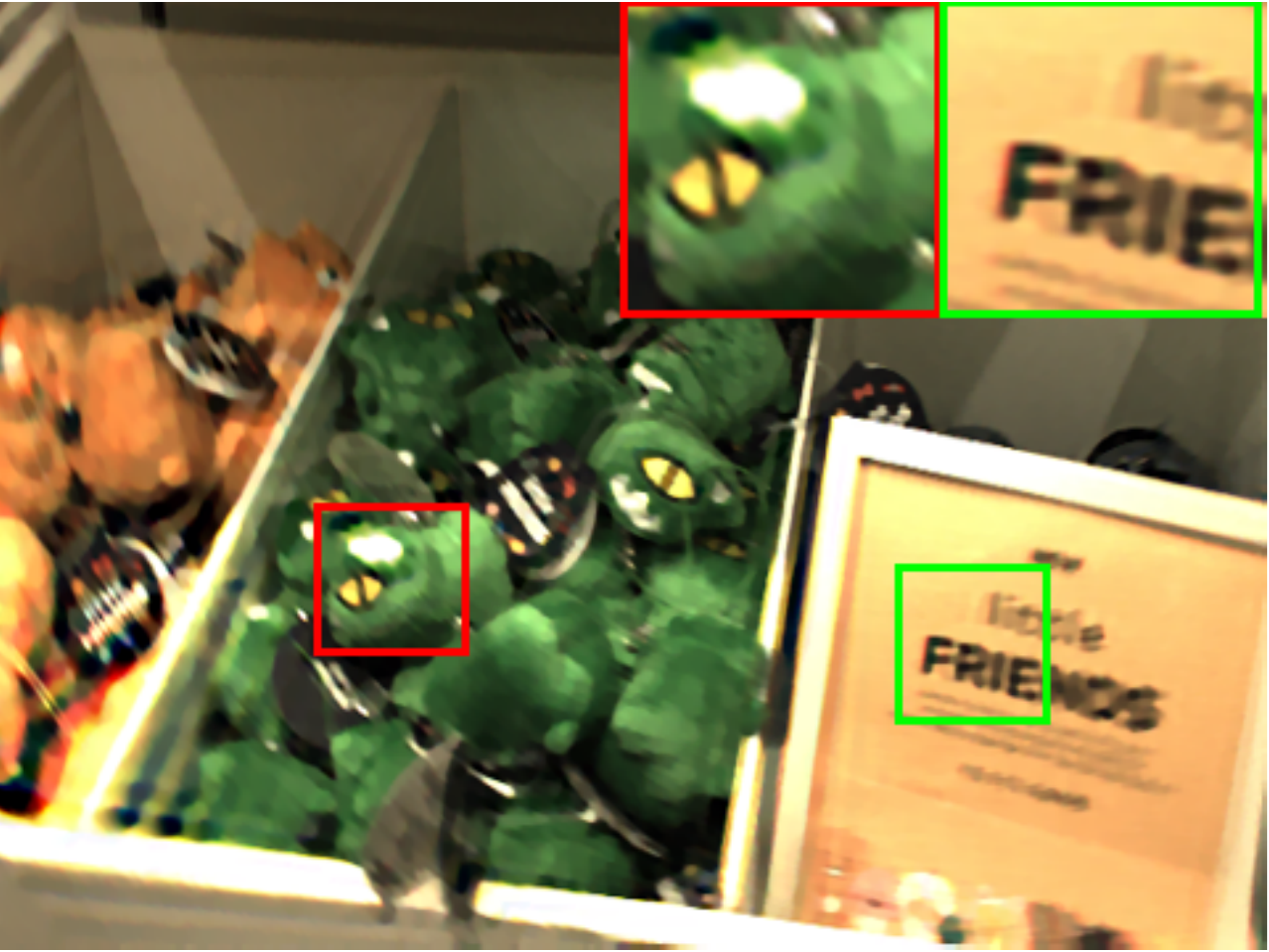}
		\vspace*{-0.12in}
		
		\subfloat[]{\includegraphics[width=0.24\textwidth]{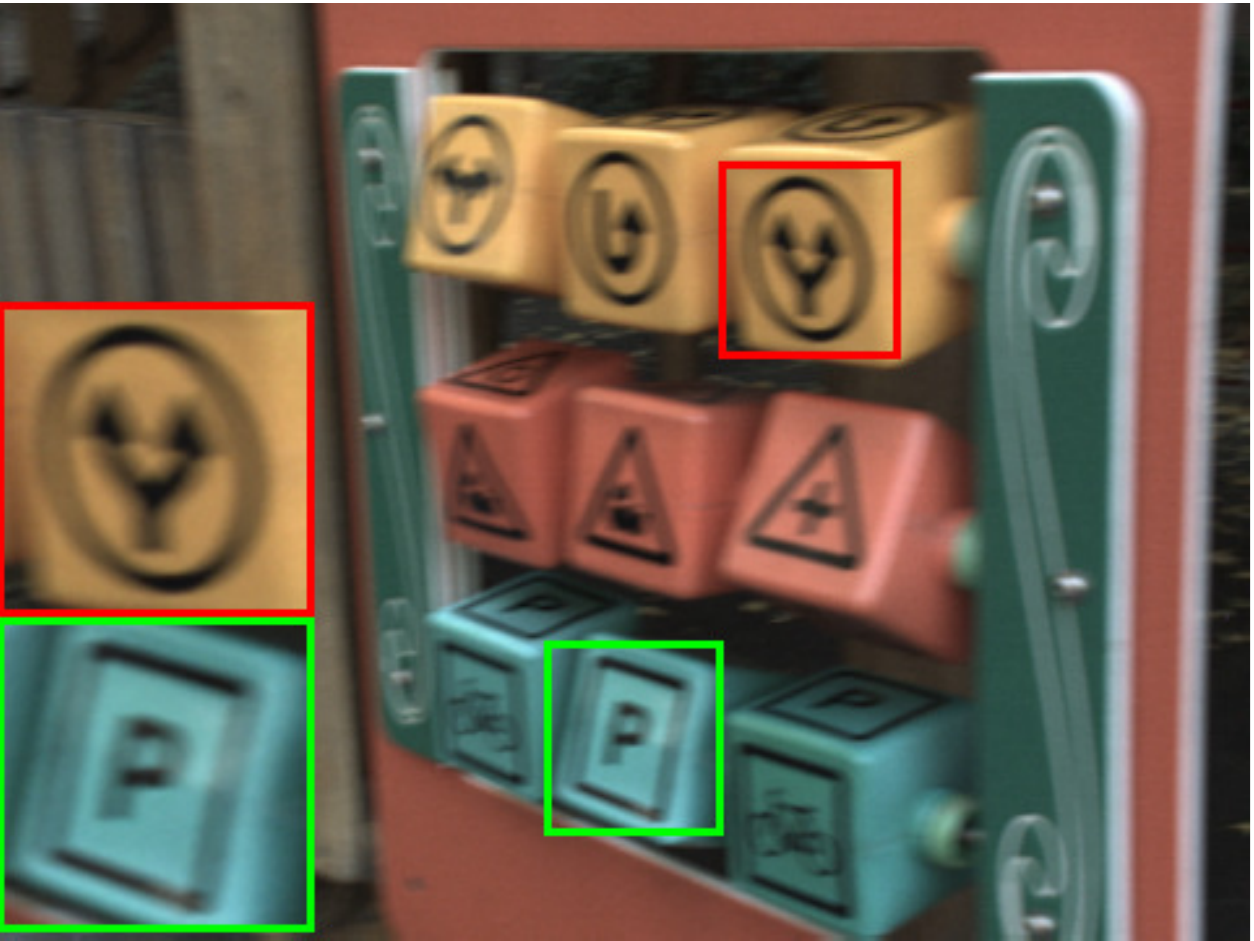}}\hspace*{0.01in}
		\subfloat[]{\includegraphics[width=0.24\textwidth]{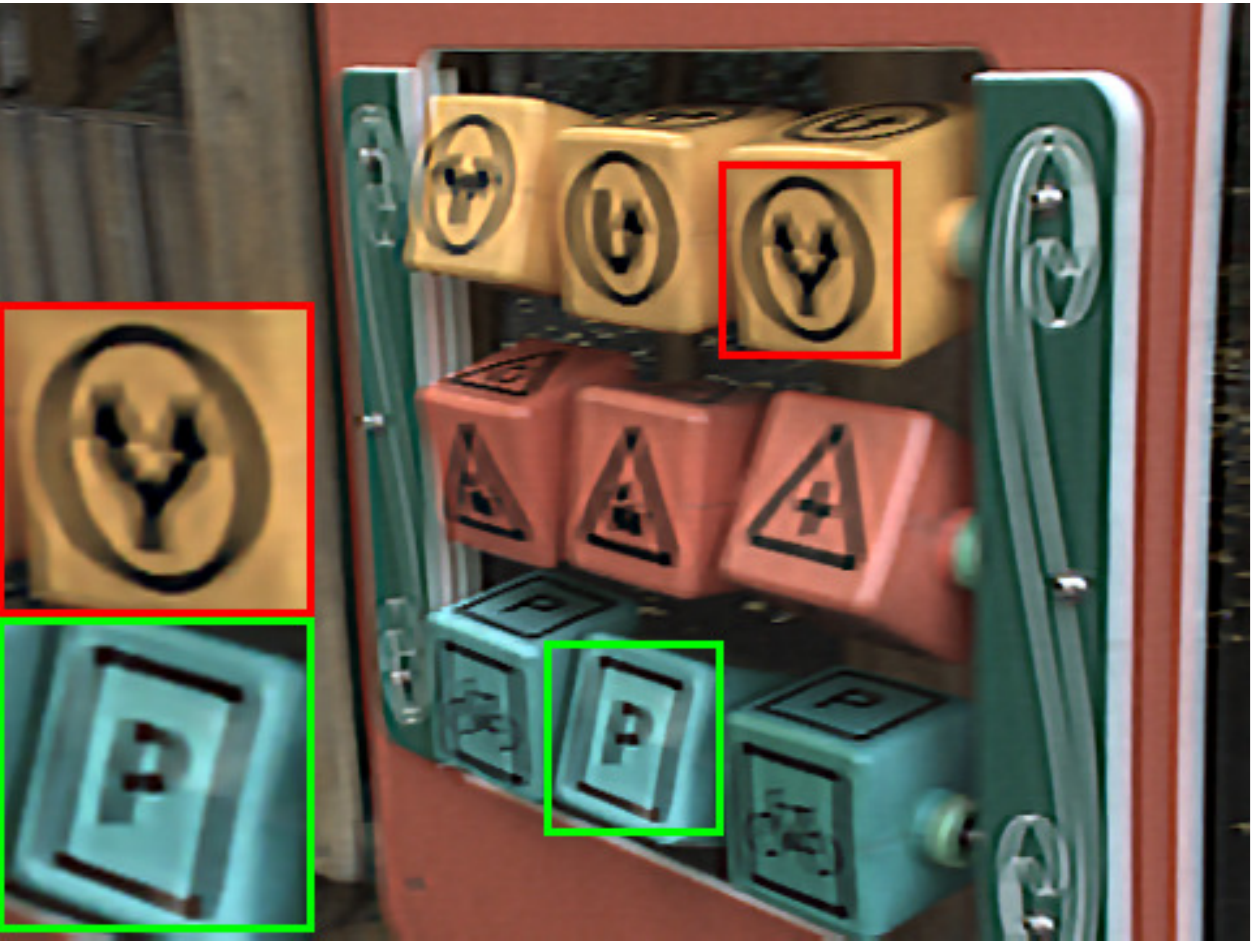}}\hspace*{0.01in}
		\subfloat[]{\includegraphics[width=0.24\textwidth]{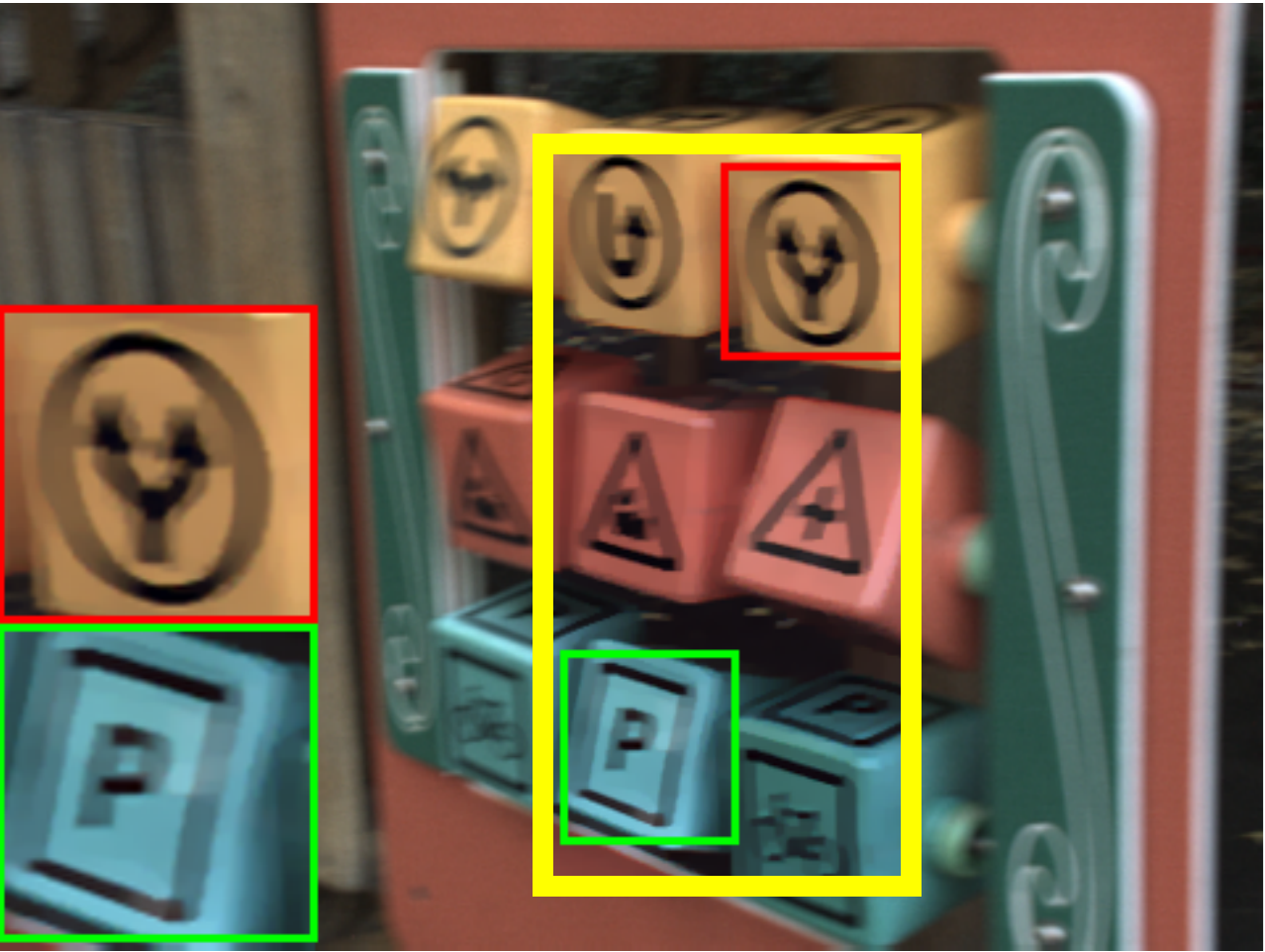}}\hspace*{0.01in}
		\subfloat[]{\includegraphics[width=0.24\textwidth]{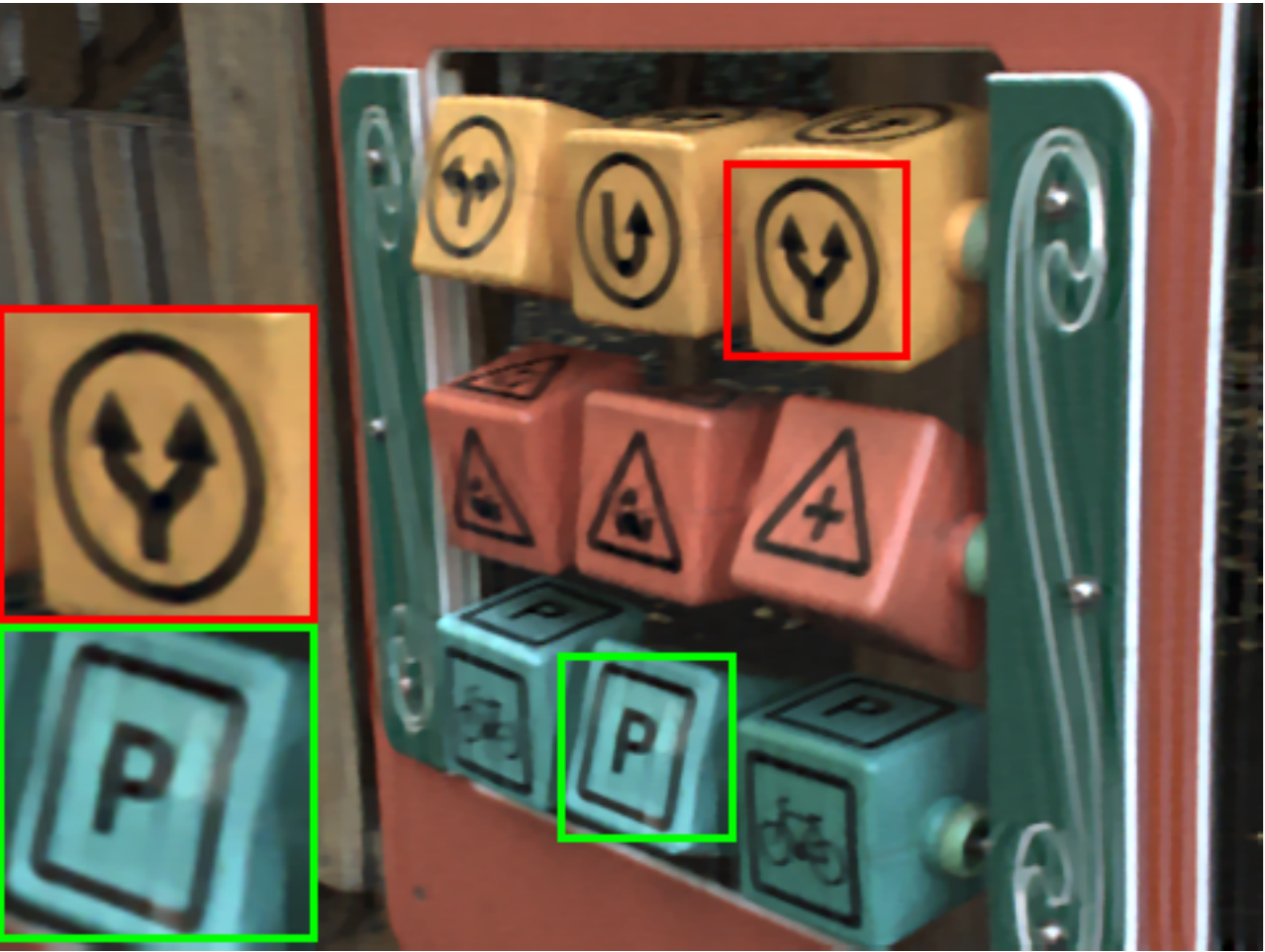}}
	\end{center}
	\vspace*{-0.23in}
	\caption{Deblurring result for real light field dataset with comparison to global camera motion estimation methods. (a) Blurred input image. (b)  Result of Hu~et al.~\cite{hu2014joint}. (c) Srinivasan et al.~\cite{srinivasan2017light}. (d) Proposed algorithm.}
	\label{qual:deblur_real2}
\end{figure*}

\subsection{Light Field Deblurring}

\textbf{Real Data.}~Figure~\ref{qual:deblur_real1} and Figure~ \ref{qual:deblur_real2} show the light field deblurring results for blurred real light field with spatially varying blur kernels.
In Figure~\ref{qual:deblur_real1}, the result is compared with the existing motion deblurring methods~\cite{kim2014segmentation,sun2015learning} which utilize motion flow estimation.
It is shown that the proposed algorithm reconstructs sharper latent image better than others.
Note that \cite{kim2014segmentation,sun2015learning} show satisfactory performance only for small blur kernels.

Figure~\ref{qual:deblur_real2} shows the comparison results with the deblurring method based on the global camera motion model~\cite{hu2014joint,srinivasan2017light}.
In comparison with~\cite{srinivasan2017light}, we deblur only cropped regions shown in the yellow boxes of Figure~\ref{qual:deblur_real2}(c) due to GPU memory overflow ($>$12GB) for larger spatial resolution.

\cite{hu2014joint} assumes the scene depth is piecewisely planar. Therefore, it cannot be generalized to arbitrary scene, yielding unsatisfactory deblurring result.
\cite{srinivasan2017light} estimates the reasonably correct camera motion of the blurred light field while their output is less deblurred.
Note that \cite{srinivasan2017light} can not handle the rotational camera motion which produces completely different blur kernels from translational motion.
On the other hand, the proposed algorithm fully utilizes the 6-DOF camera motion and the scene depth, yielding outperforming results for the arbitrary scene.

The light field deblurring experiments with real data show that the proposed algorithm works robustly even for the hand-shake motion which does not match the proposed motion path model.
The proposed algorithm showed superior deblurring performance for both natural indoor and outdoor scenes, which confirms the robustness of the proposed algorithm to noise and depth level.
\begin{figure*}[t]
	\begin{center}	
		\vspace*{0.15in}
		\includegraphics[width=0.2\textwidth]{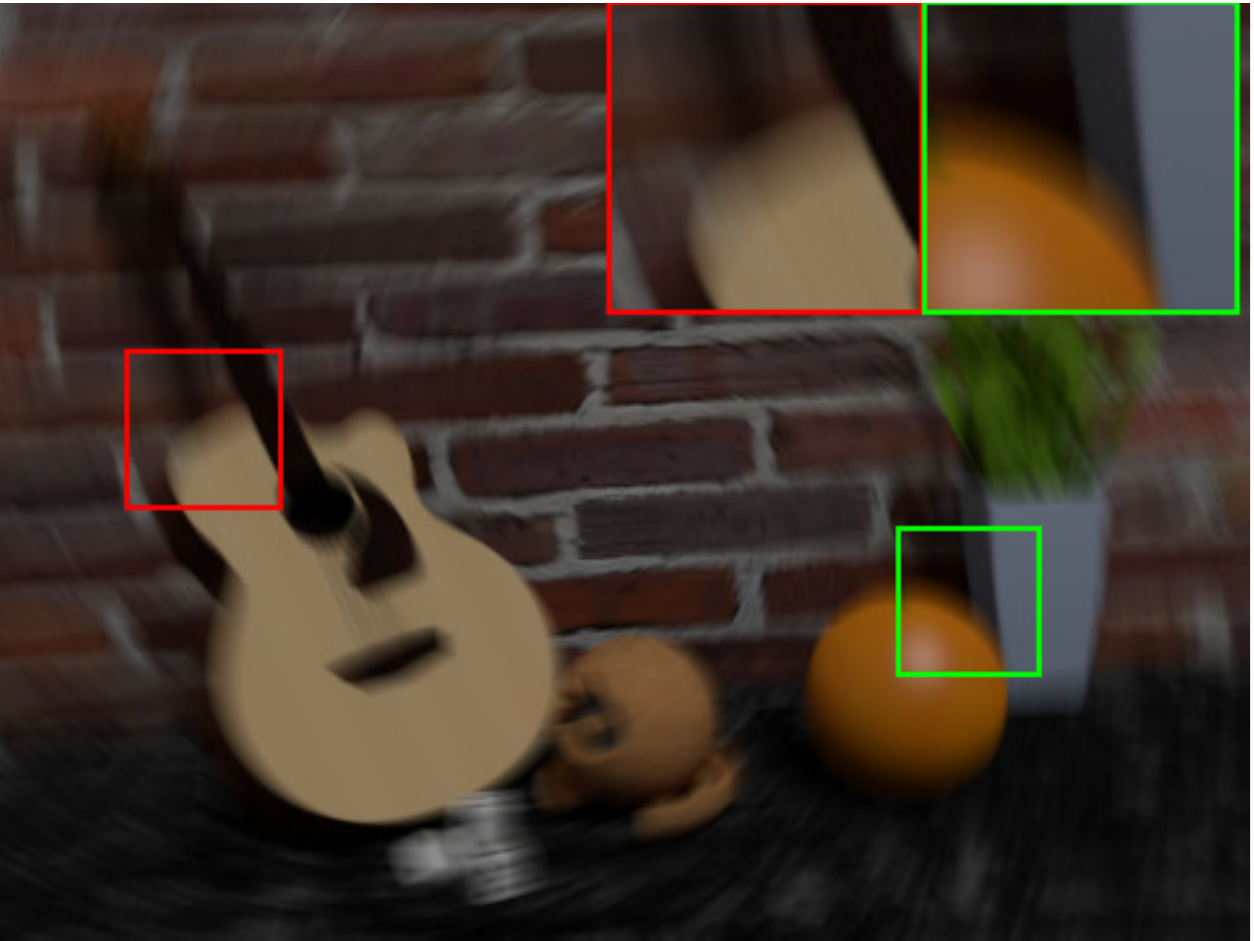}\hspace*{0.01in}
		\includegraphics[width=0.2\textwidth]{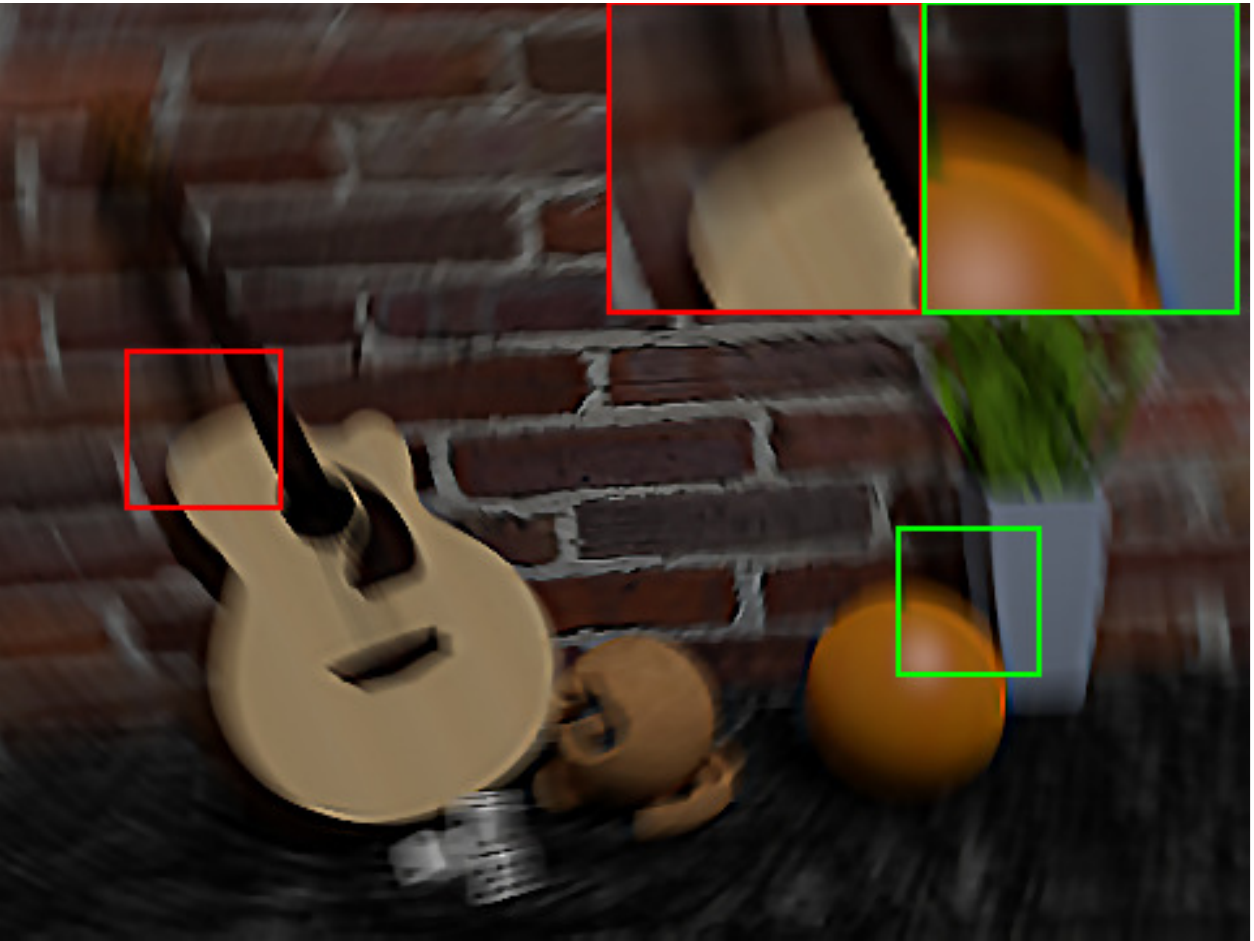}\hspace*{0.01in}
		\includegraphics[width=0.2\textwidth]{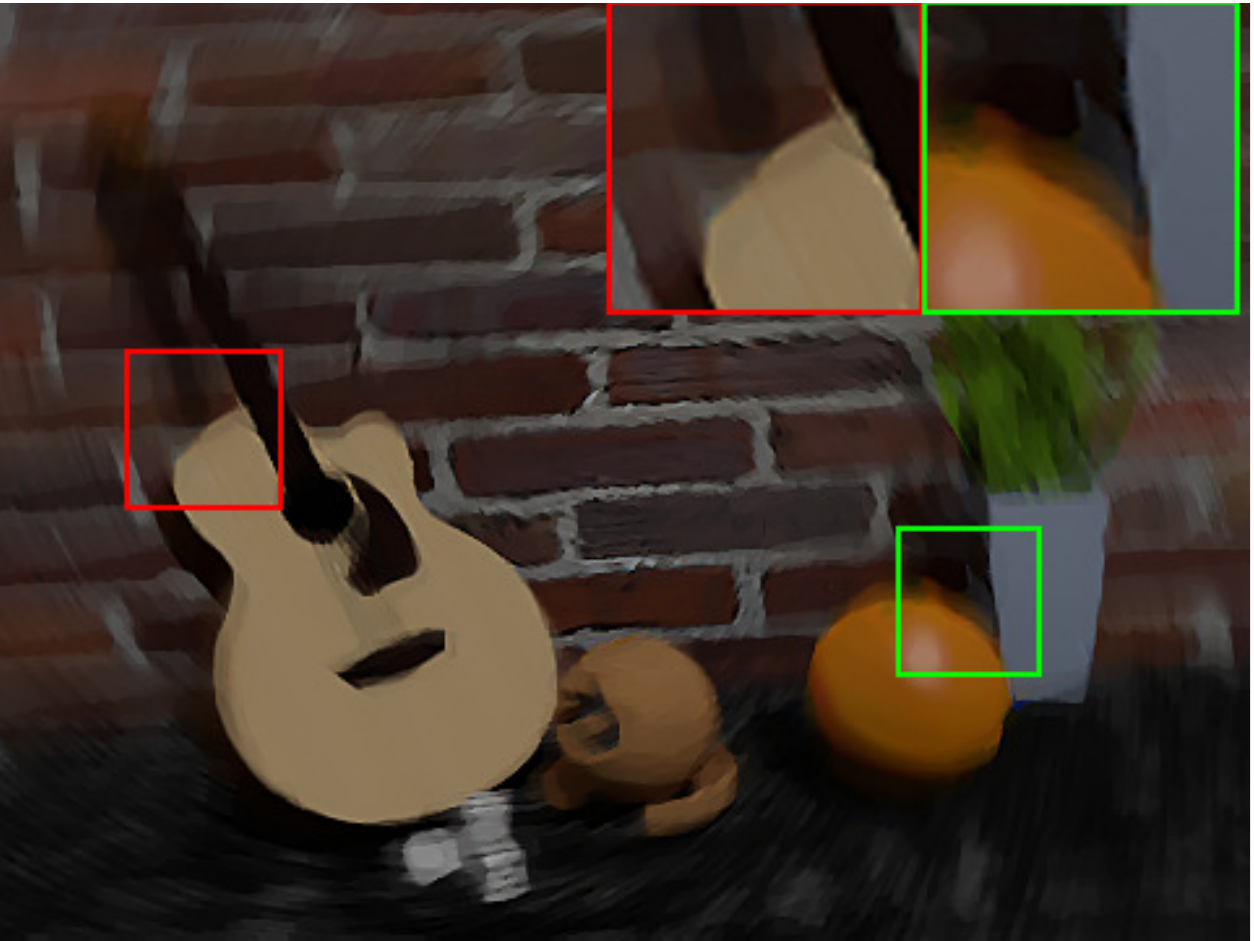}\hspace*{0.01in}
		\includegraphics[width=0.2\textwidth]{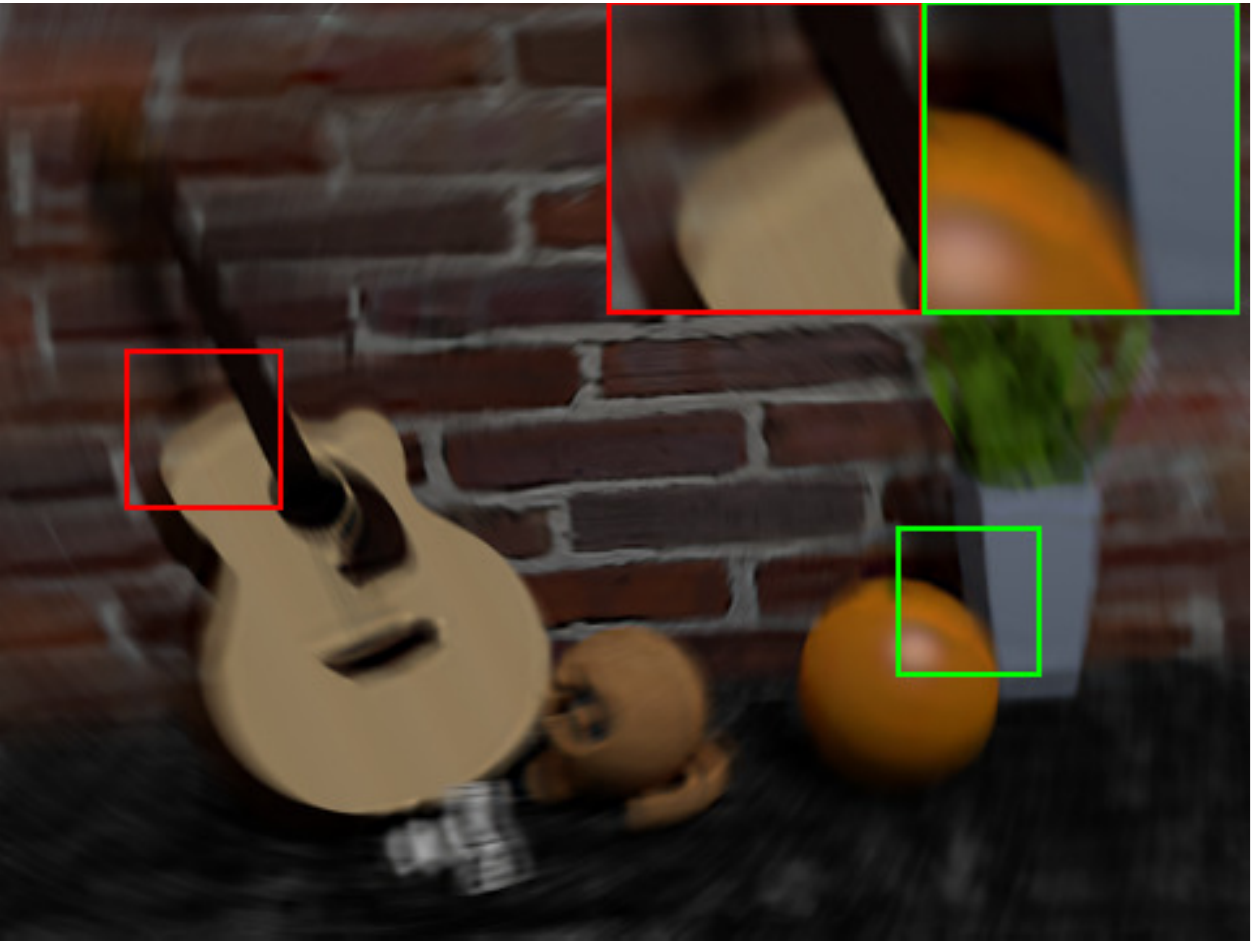}\hspace*{0.01in}
		\includegraphics[width=0.2\textwidth]{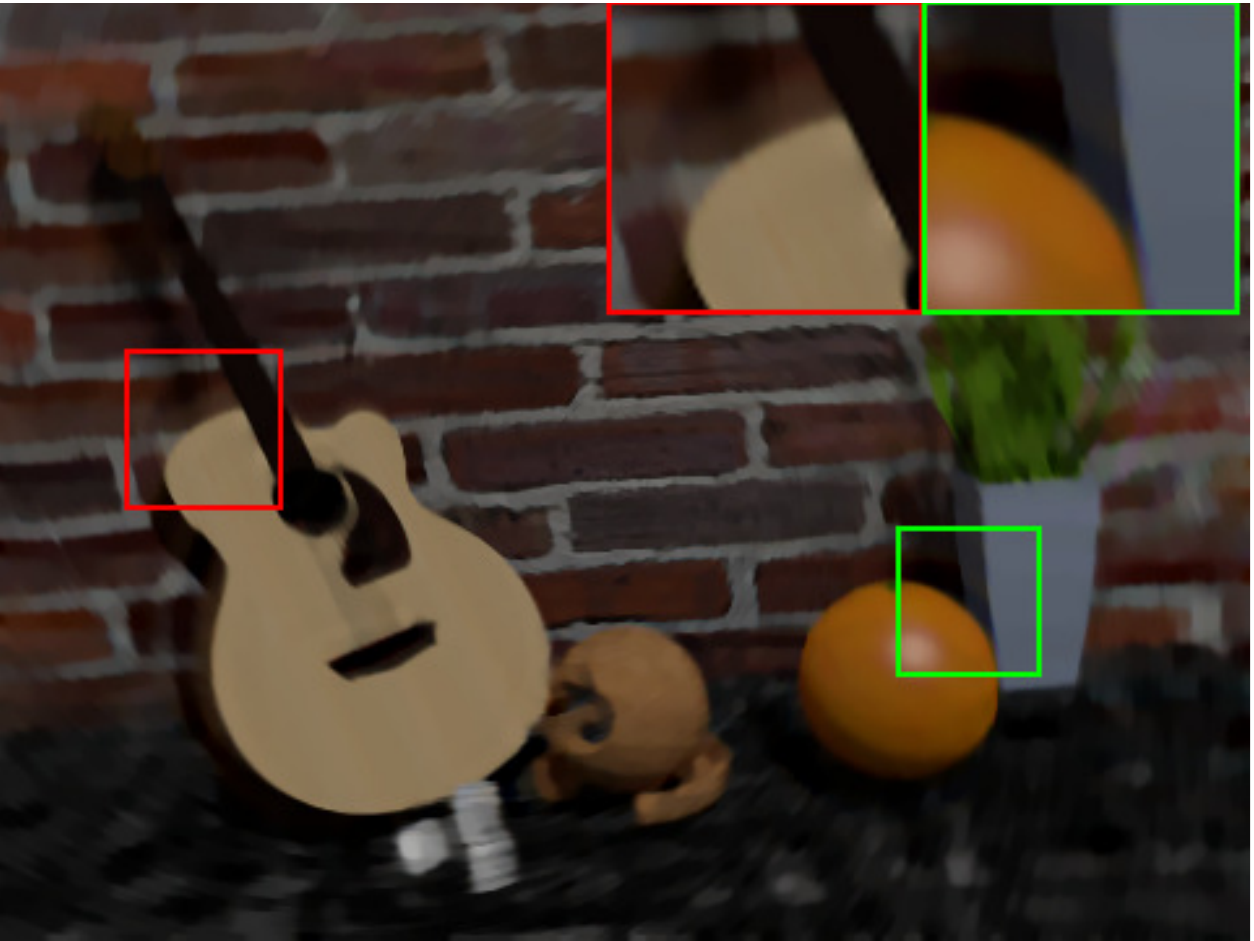}\hspace*{0.01in}
		\vspace*{0.03in}		
\\		\includegraphics[width=0.2\textwidth]{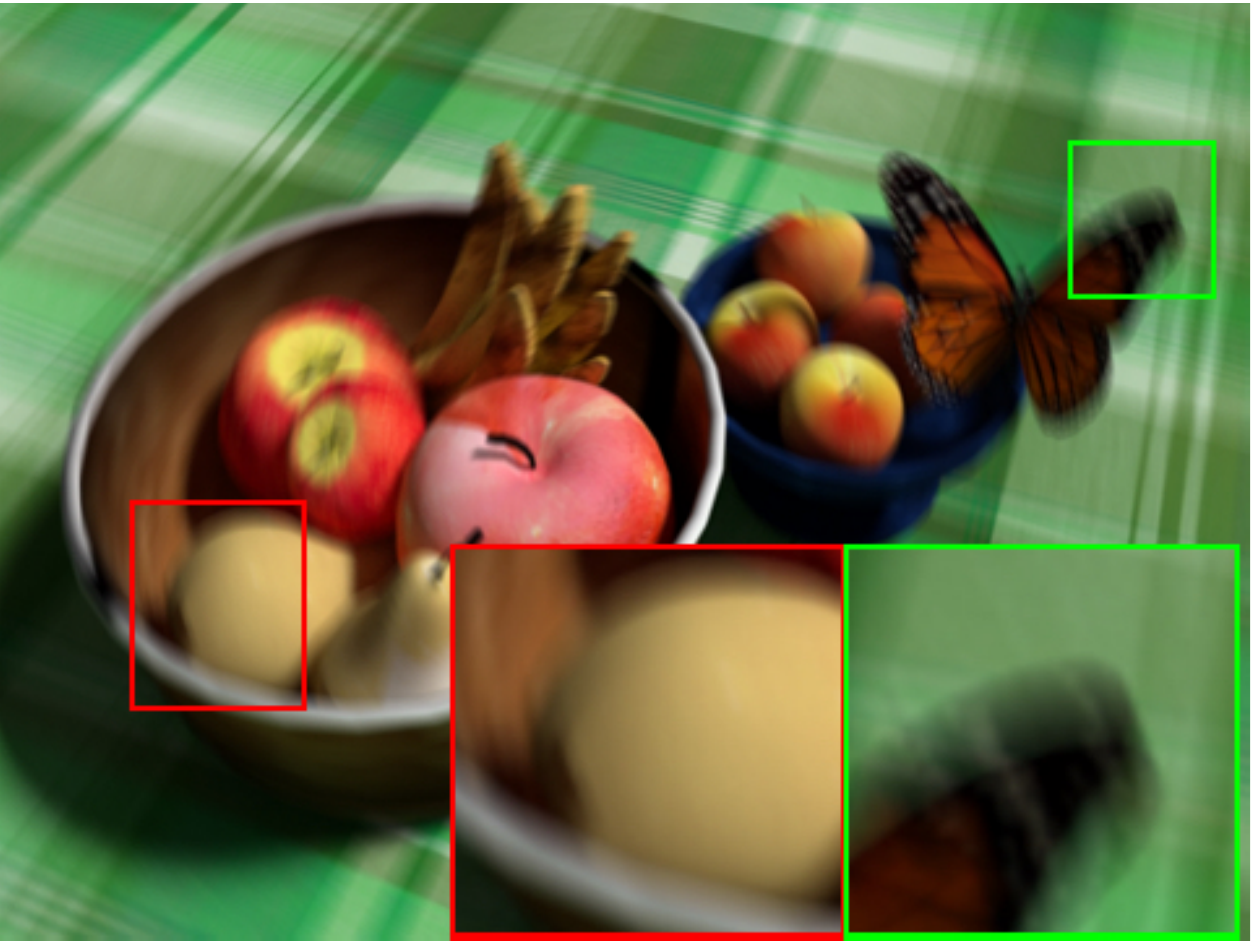}\hspace*{0.01in}
		\includegraphics[width=0.2\textwidth]{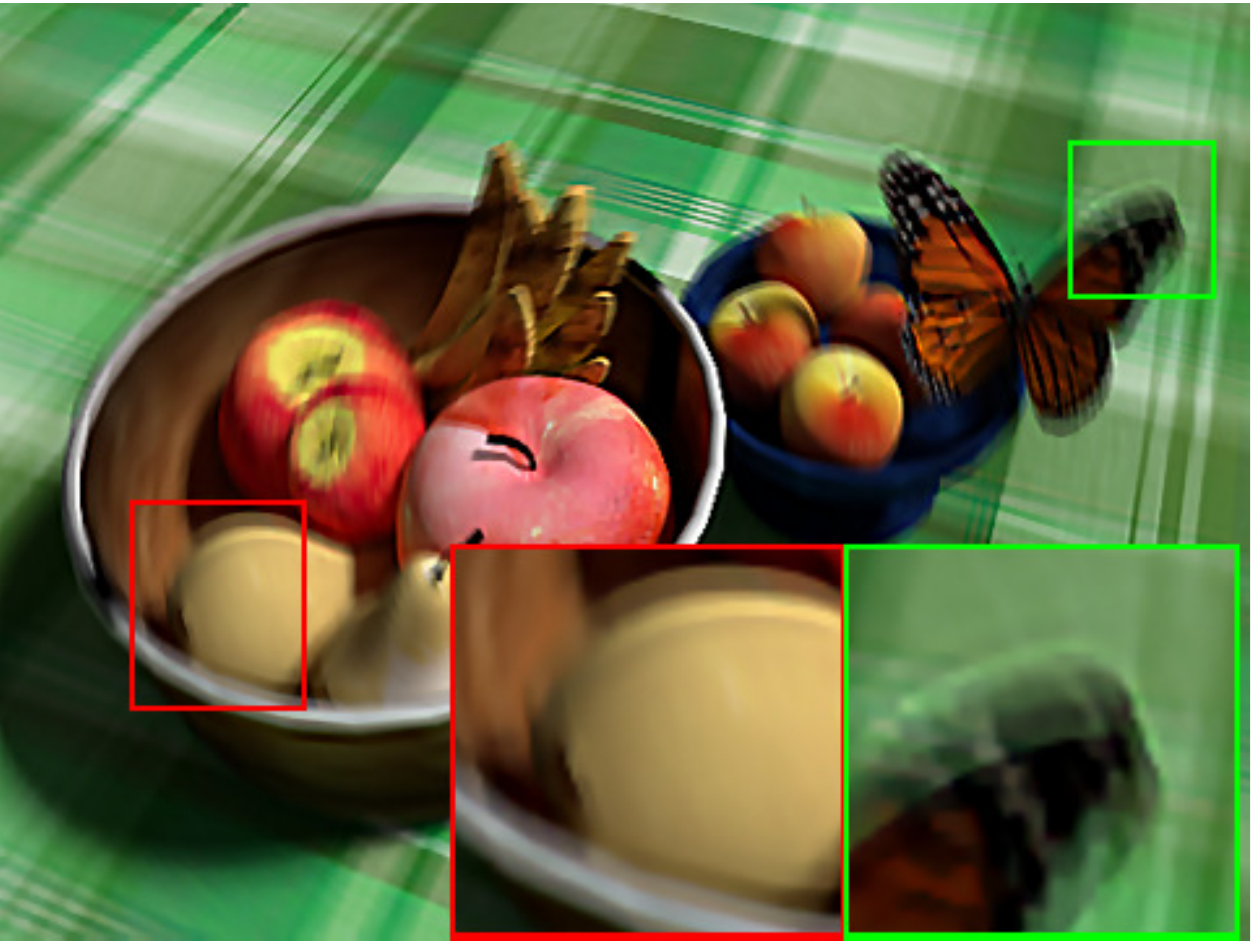}\hspace*{0.01in}
		\includegraphics[width=0.2\textwidth]{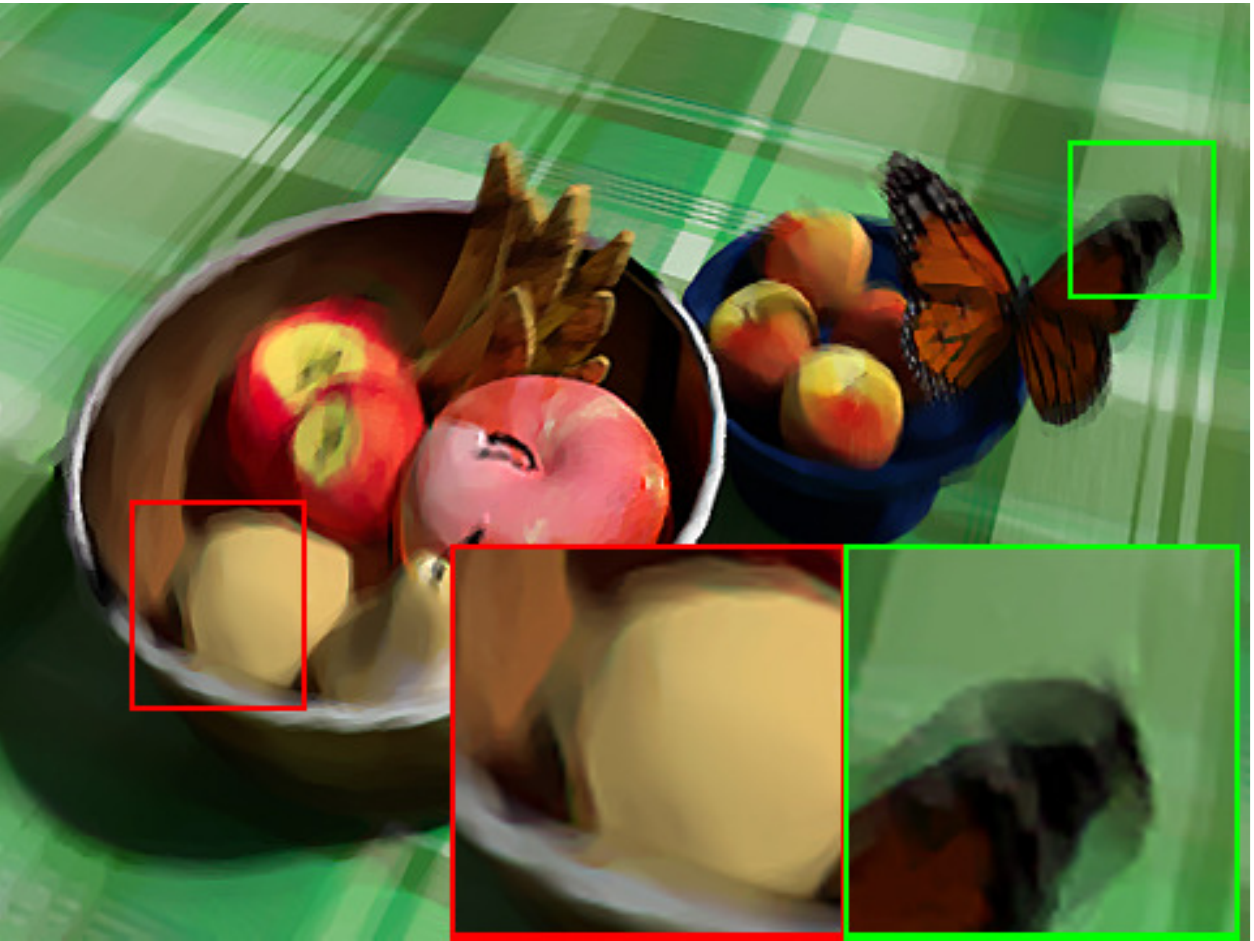}\hspace*{0.01in}
		\includegraphics[width=0.2\textwidth]{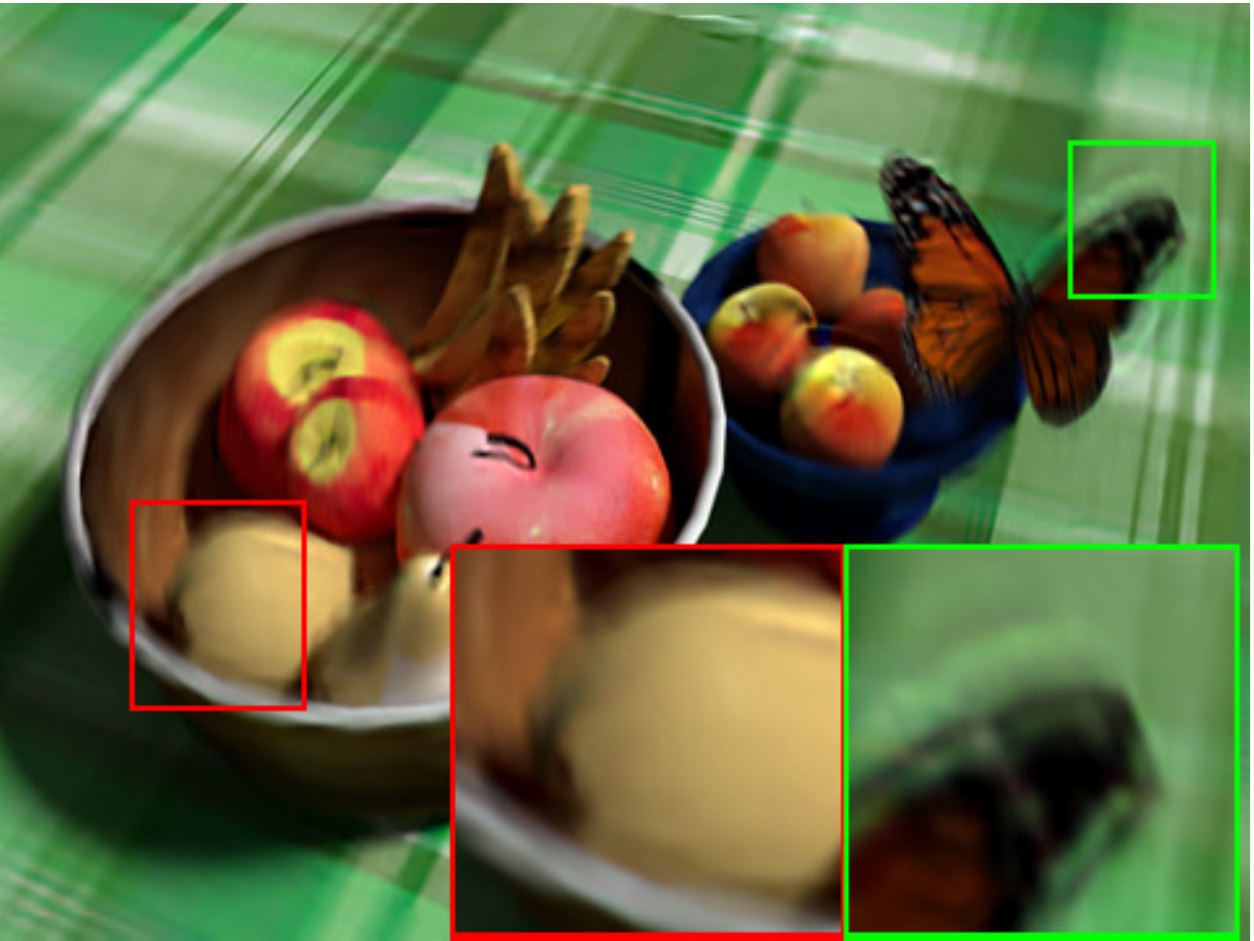}\hspace*{0.01in}
		\includegraphics[width=0.2\textwidth]{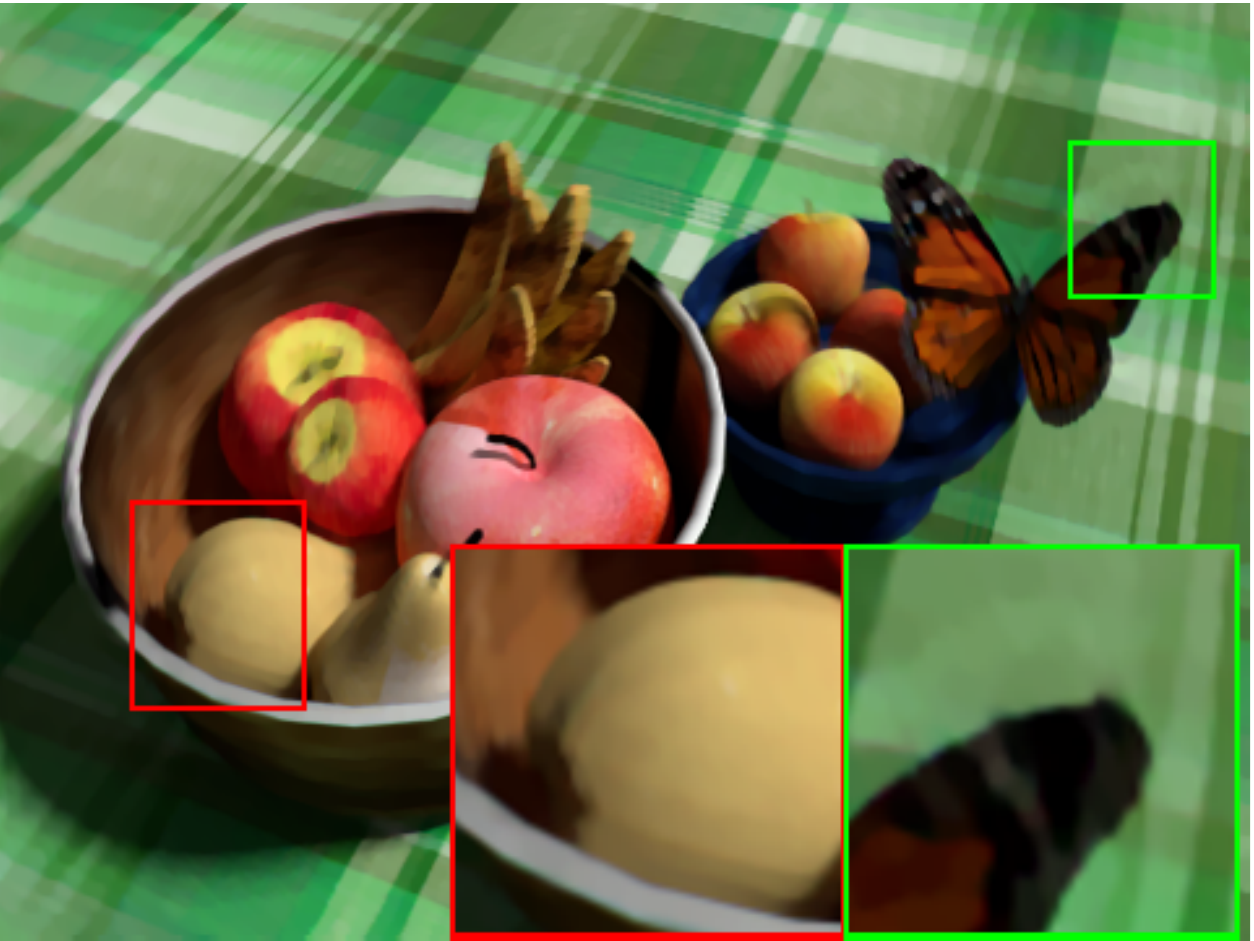}\hspace*{0.01in}
		\vspace*{-0.11in}		
\\		\subfloat[]{\includegraphics[width=0.2\textwidth]{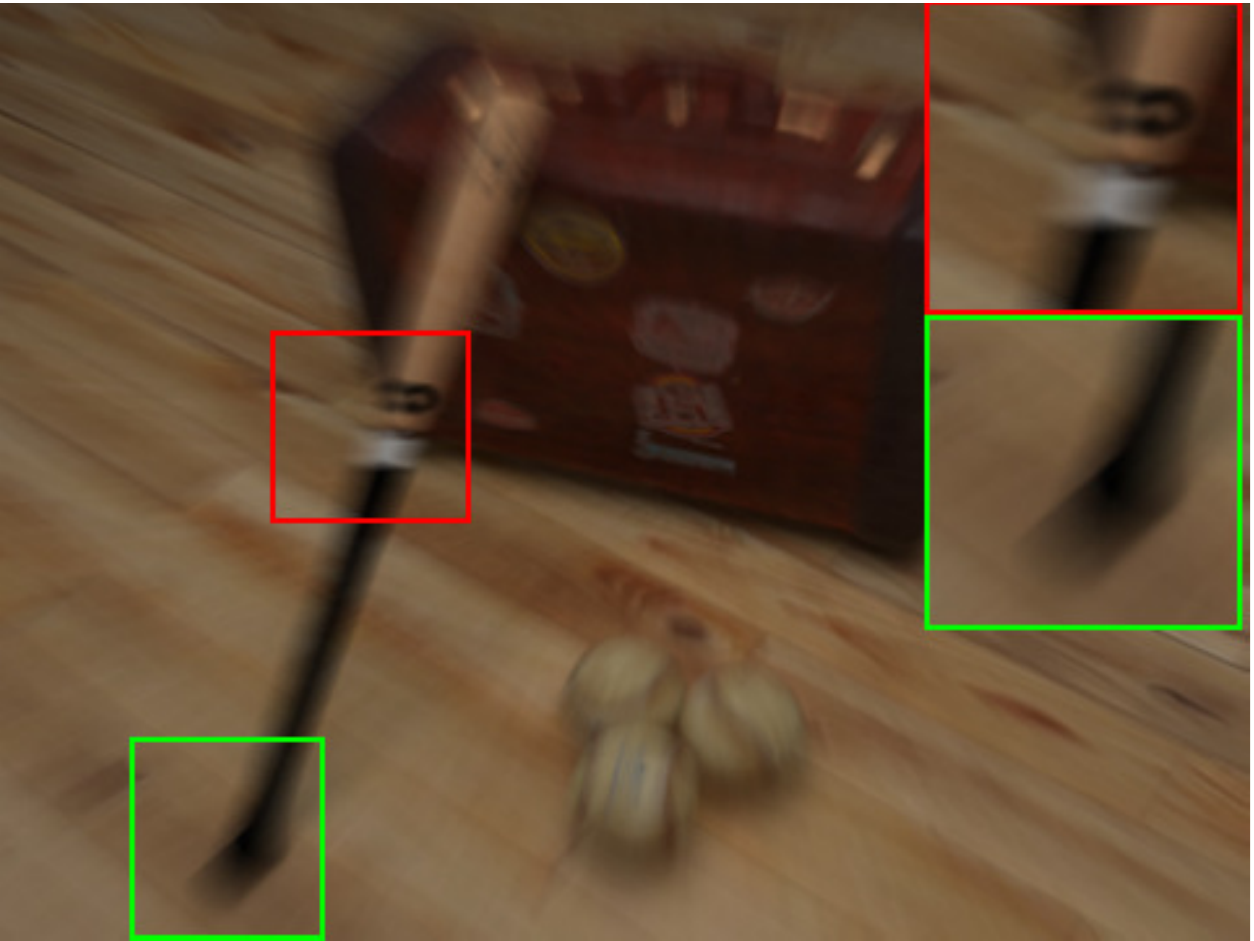}}\hspace*{0.01in}
		\subfloat[]{\includegraphics[width=0.2\textwidth]{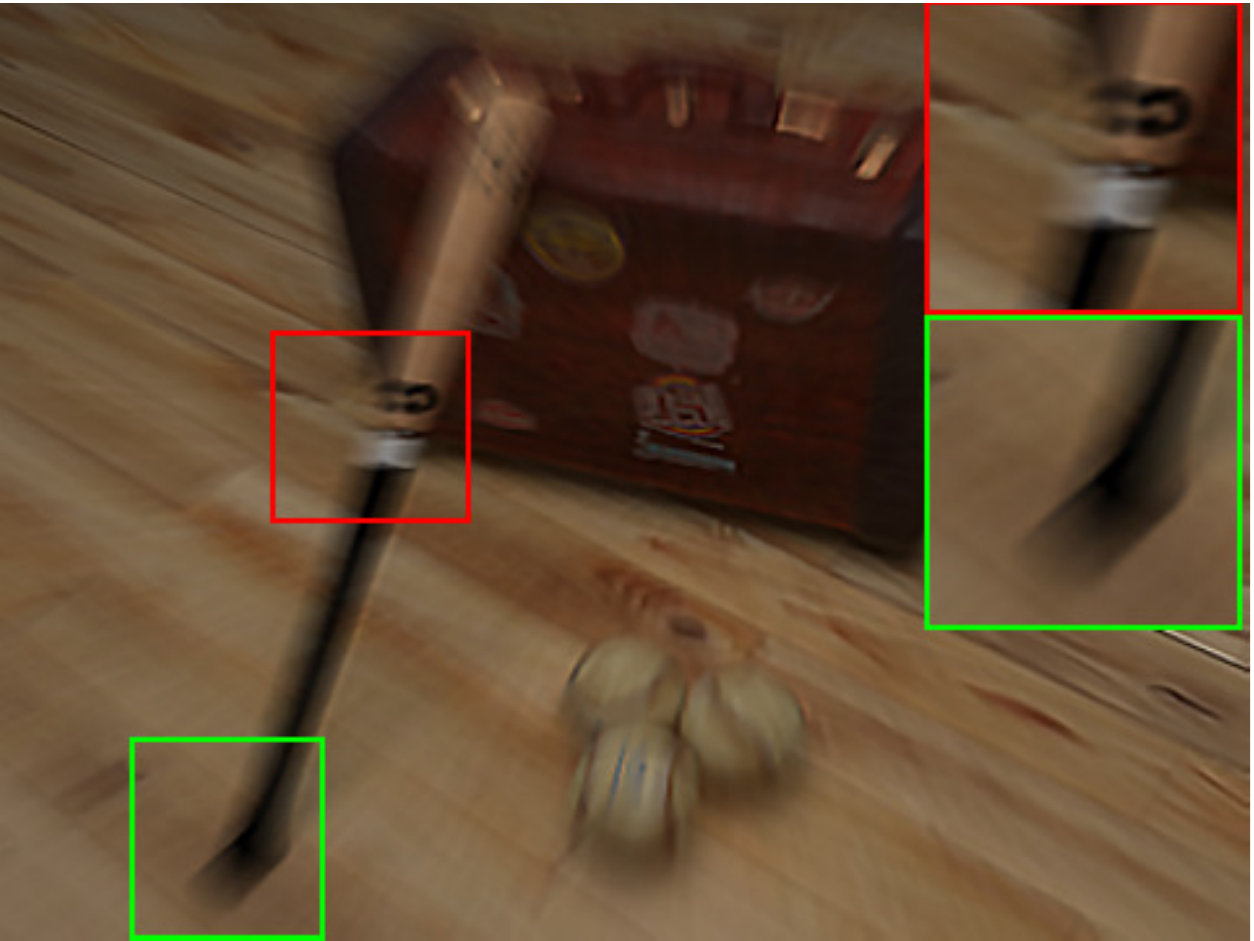}}\hspace*{0.01in}
		\subfloat[]{\includegraphics[width=0.2\textwidth]{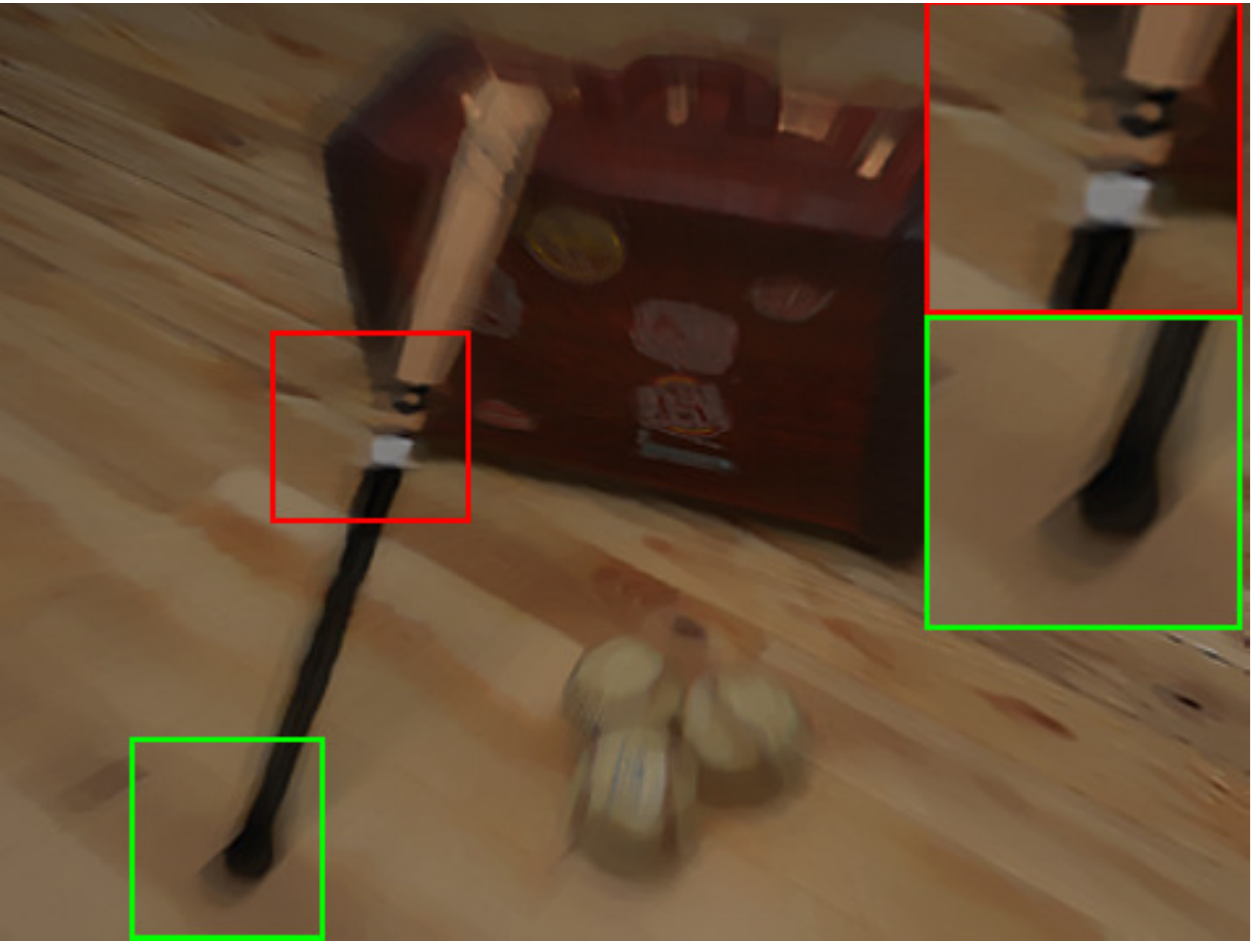}}\hspace*{0.01in}
		\subfloat[]{\includegraphics[width=0.2\textwidth]{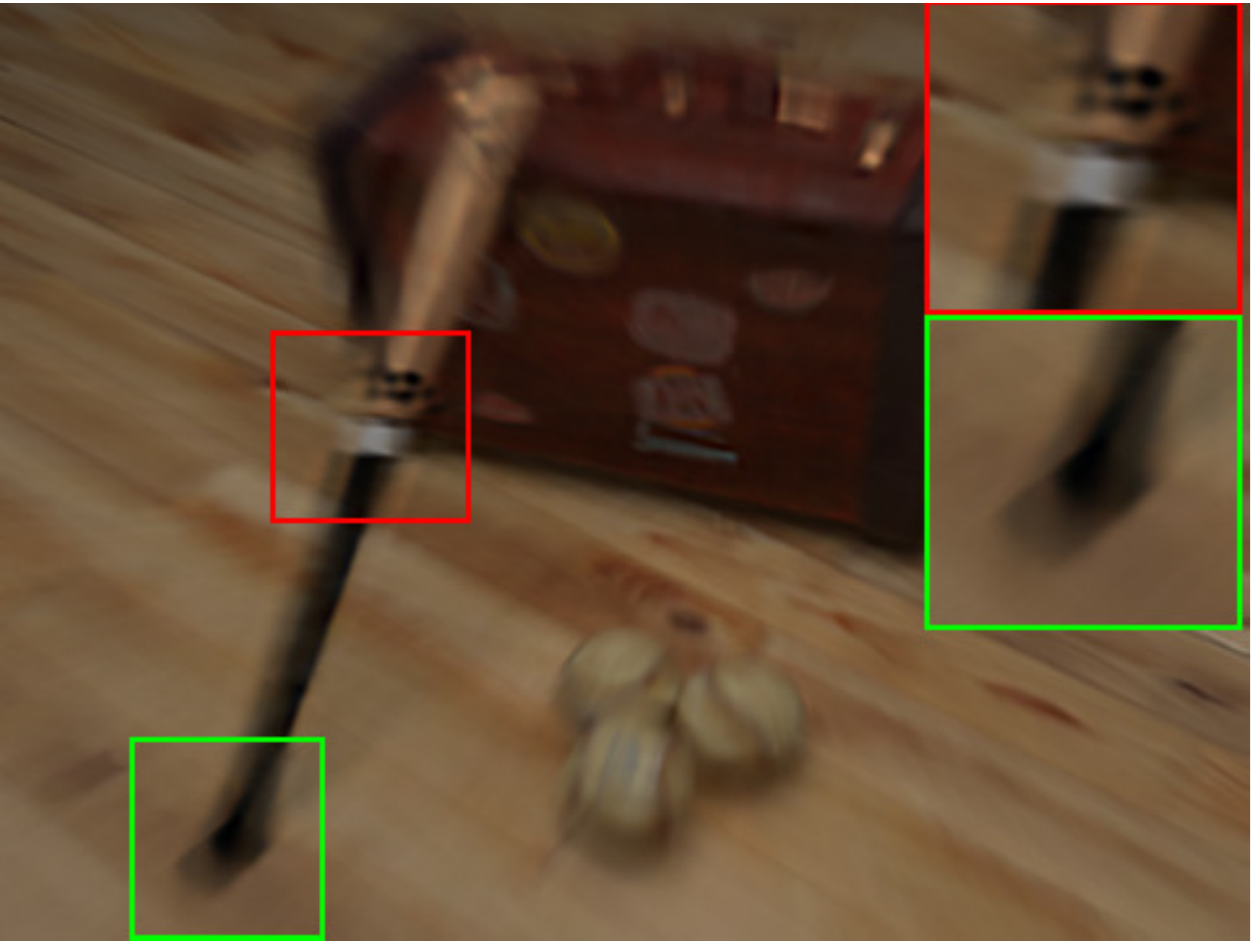}}\hspace*{0.01in}
		\subfloat[]{\includegraphics[width=0.2\textwidth]{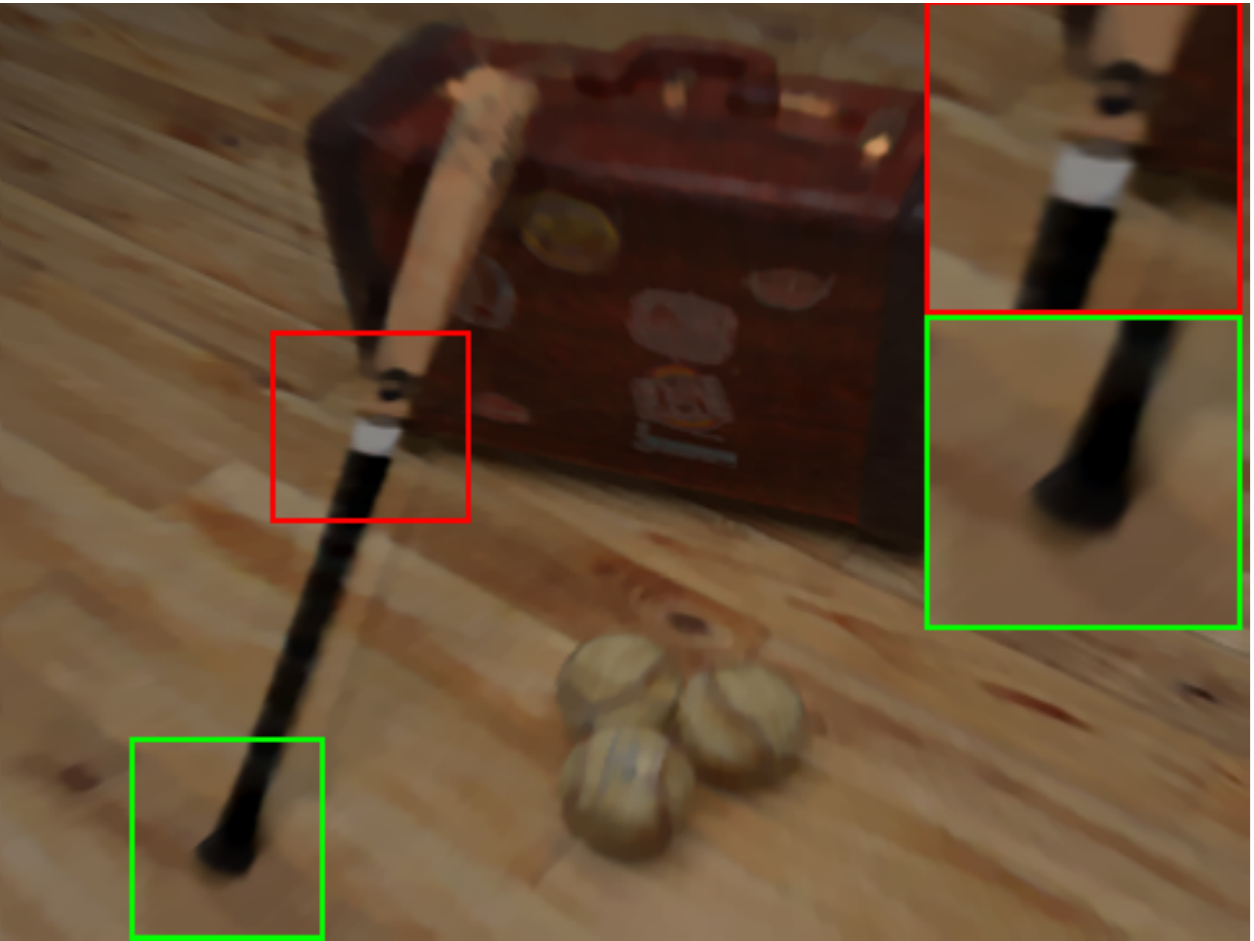}}\hspace*{0.01in}
		\vspace*{-0.13in}
\\
	\end{center}
	\vspace*{-0.05in}
	\caption{Deblurring result for synthetic light field. (a) Blurred input light field. (b) Result of Hu~et al.~\cite{hu2014joint}. (c)  Kim and Lee~\cite{kim2014segmentation}. (d)  Sun~et al.~\cite{sun2015learning}. (e) Proposed algorithm.}
	\label{qual:deblur_synthetic}
	\vspace*{-0.05in}
\end{figure*}

\begin{table*}[t]
	\caption{Quantitative evaluation of deblurring on synthetic light field dataset (in PSNR and SSIM).}
	\centering
\renewcommand{\arraystretch}{1.1}{
	\scalebox{0.8}{
		\begin{tabular}{ccccccccccccc}
			\toprule
			& \multicolumn{2}{c}{Forward} &  \multicolumn{2}{c}{Rotation} & \multicolumn{2}{c}{Translation} & \multicolumn{2}{c}{Forward+Rot.} & \multicolumn{2}{c}{Forward+Tran.} & \multicolumn{2}{c}{Rot.+Tran.}\\
			\multicolumn{1}{c}{Methods}& PSNR & SSIM & PSNR & SSIM & PSNR & SSIM & PSNR & SSIM & PSNR & SSIM & PSNR & SSIM \\
			\cmidrule(r){1-1}\cmidrule(lr){2-3}\cmidrule(lr){4-5}\cmidrule(lr){6-7}\cmidrule(lr){8-9}\cmidrule(lr){10-11}\cmidrule(l){12-13}
			\multicolumn{1}{c}{Input}& 20.72 & 0.740 & 20.37 & 0.731 & 21.82 & 0.758 & 19.84 & 0.723 & 20.30 & 0.731 & 19.79 & 0.728 \\
			\multicolumn{1}{c}{Hu~et al.~\cite{hu2014joint}}& 20.00 & 0.716 & 19.42 & 0.704 & 21.42 & 0.745 & 19.18 & 0.701 & 19.43 & 0.699 & 19.24 & 0.711 \\
			\multicolumn{1}{c}{Kim and Lee~\cite{kim2014segmentation}}& 20.06 & 0.721 & 19.78 & 0.714 & 21.42 & 0.749 & 19.32 & 0.706 & 19.65 & 0.708 & 19.34 & 0.714 \\
			\multicolumn{1}{c}{Sun~et al.~\cite{sun2015learning}}& 27.69 & 0.896 & 27.68 & 0.881 & 25.41 & 0.856 & 27.40 & 0.874 & 27.23 & 0.899 & 26.42 & 0.868 \\
			\multicolumn{1}{c}{Proposed Method}& \bf{29.24} & \bf{0.915} & \bf{29.14} & \bf{0.913} & \bf{26.99} & \bf{0.876} & \bf{28.92} & \bf{0.905} & \bf{28.91} & \bf{0.922} & \bf{27.85} & \bf{0.893} \\
			\midrule
			\multicolumn{1}{c}{Input~(cropped)}& 21.01 & 0.758 & 21.19 & 0.746 & 19.39 & 0.698 & 21.73 & 0.758 & 21.67 & 0.782 & 20.50 & 0.745 \\
			\multicolumn{1}{c}{Srinivasan~et al.~\cite{srinivasan2017light}}& 17.15 & 0.730 & 19.02 & 0.652 & 16.28 & 0.620 & 19.17 & 0.660 & 16.20 & 0.726 & 16.38 & 0.626 \\
			\multicolumn{1}{c}{Proposed Method}& \bf{27.15} & \bf{0.871} & \bf{27.32} & \bf{0.870} & \bf{25.30} & \bf{0.836} & \bf{28.83} & \bf{0.904} & \bf{28.01} & \bf{0.901} & \bf{25.88} & \bf{0.867} \\
			\bottomrule
		\end{tabular}
	}
}
	\label{quan:deblur}
	\end{table*}
\noindent

\textbf{Synthetic Data.}~The performance of the proposed algorithm is evaluated using synthetic light field dataset, as shown in Figure~\ref{qual:deblur_synthetic} and Table~\ref{quan:deblur}.
The synthetic data consists of forward, rotation, in-plane translation motion and their combinations.
In Figure~\ref{qual:deblur_synthetic}, we visualize and compare the deblurring performance with existing motion flow methods~\cite{kim2014segmentation,sun2015learning} and a camera motion method~\cite{hu2014joint}.
In all examples, the proposed algorithm produces sharper deblurred images than others as shown clearly in the cropped boxes.

Table~\ref{quan:deblur} shows the quantitative comparison of deblurring performance by measuring PSNR and SSIM to the ground truth.
It shows that the proposed joint estimation algorithm significantly outperforms the others.
Sun~et al.~\cite{sun2015learning} achieves comparable performance to the proposed algorithm in which CNN is trained with MSE loss.
Other algorithms achieve minor improvement from the input image because the assumed blur models are simple and inconsistent with the ground truth blur.

For the comparison with \cite{srinivasan2017light}, we crop the each light field to $200\times200$ because of the GPU memory overflow. Note that we use the original setting of \cite{srinivasan2017light}.
\cite{srinivasan2017light} shows lower performance than the input blurred light field due to the spatial viewpoint shift as in the output of \cite{srinivasan2017light}.
Since the original point exists at the end point of the camera motion path in \cite{srinivasan2017light}, the viewpoint shift occurs when the estimated 3D motion is large.
It is observed that this is an additional cause to decrease PSNR and SSIM when the estimated 3D motion is different from the ground truth.
The proposed algorithm estimates the latent image with ignorable viewpoint shift because the origin is located in the middle of the camera motion path. 

\subsection{Light Field Depth Estimation}

To show the performance of light field depth estimation, we compare the proposed method with several state-of-the-art  methods~\cite{chen2014light,jeon2015accurate,tao2015depth,wang2015occlusion,williem2016robust}.
For comparison, all blurred sub-aperture images are independently deblurred using~\cite{sun2015learning} before running their own depth estimation algorithms. 

\begin{figure*}[t]
	\begin{center}
		\vspace*{0.15in}
		\subfloat[]{\includegraphics[width=0.245\textwidth]{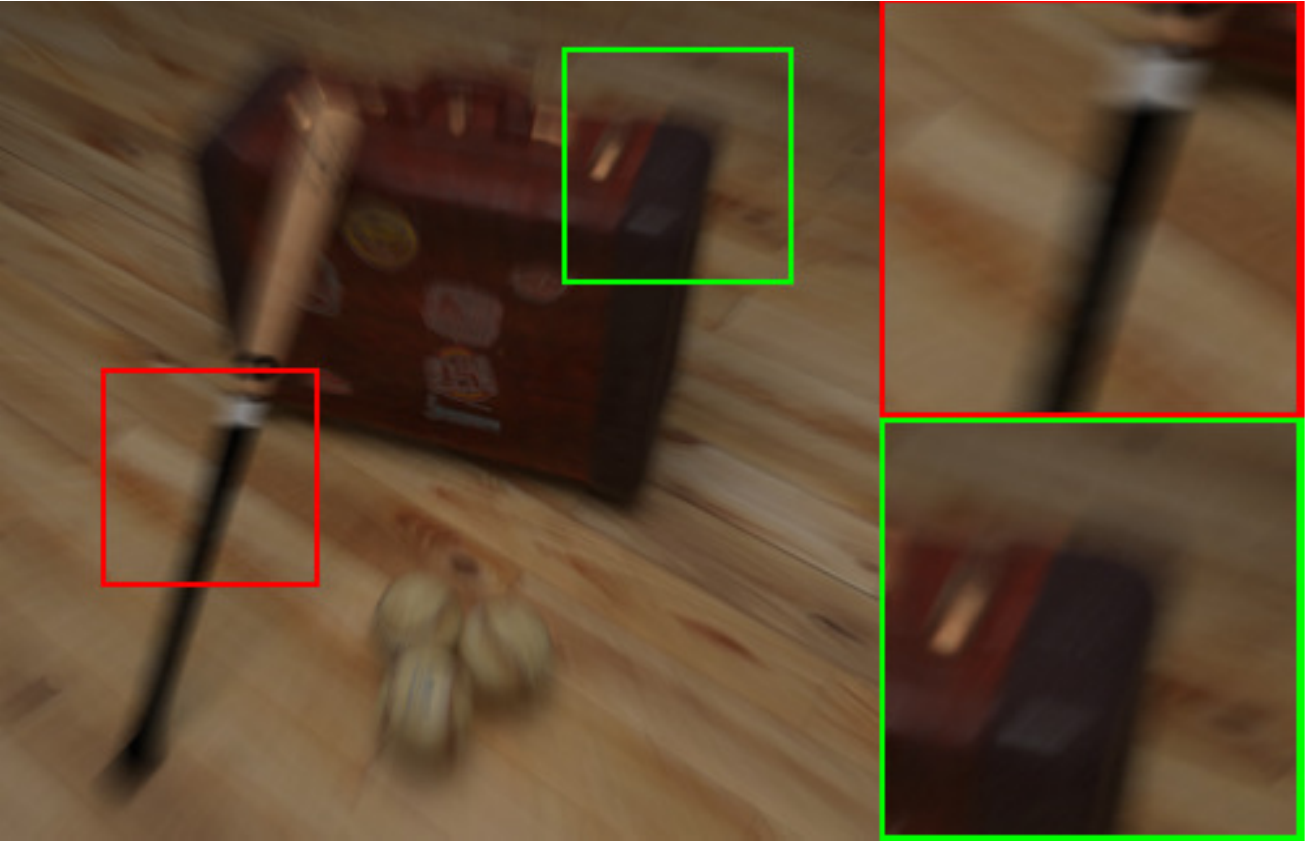}}\hspace*{0.01in}
		\subfloat[]{\includegraphics[width=0.245\textwidth]{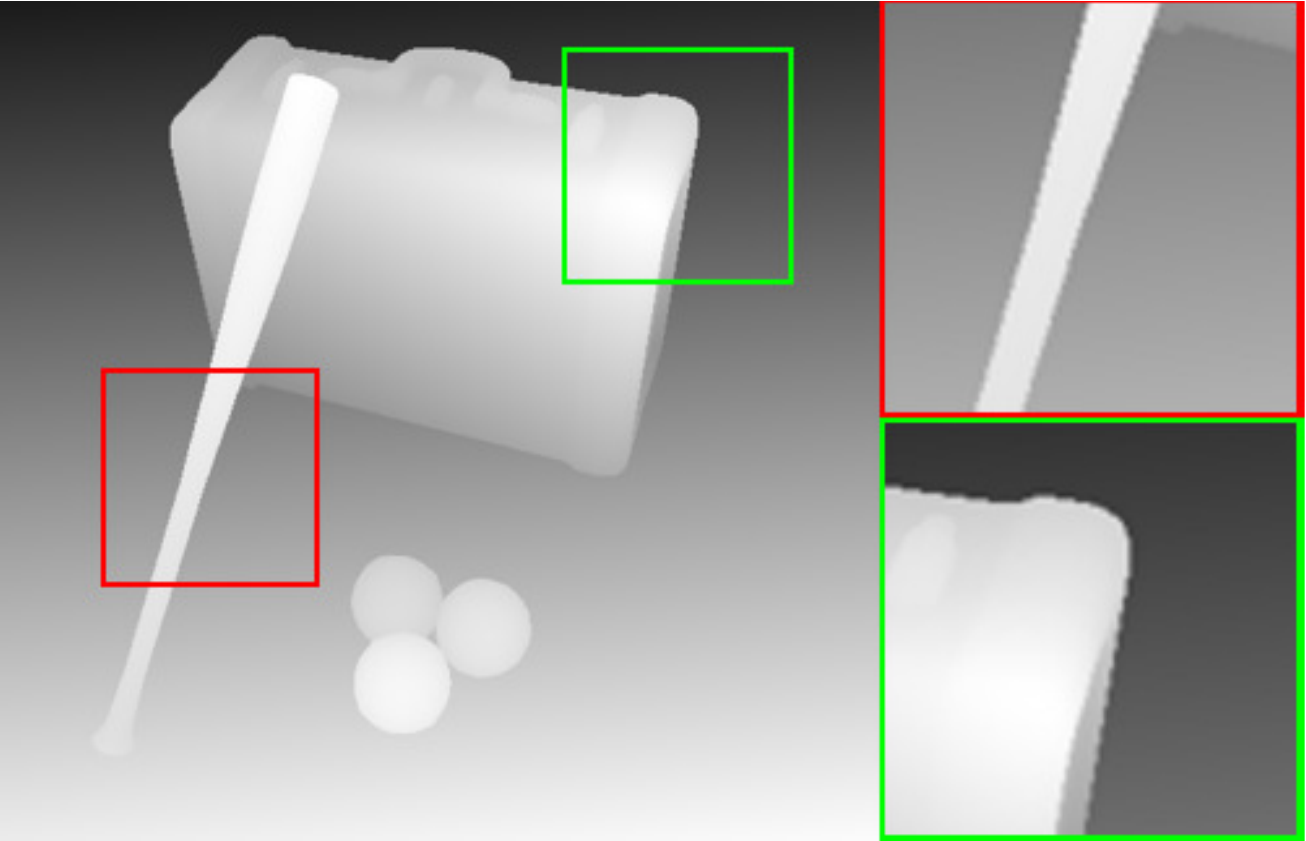}}\hspace*{0.01in}
		\subfloat[]{\includegraphics[width=0.245\textwidth]{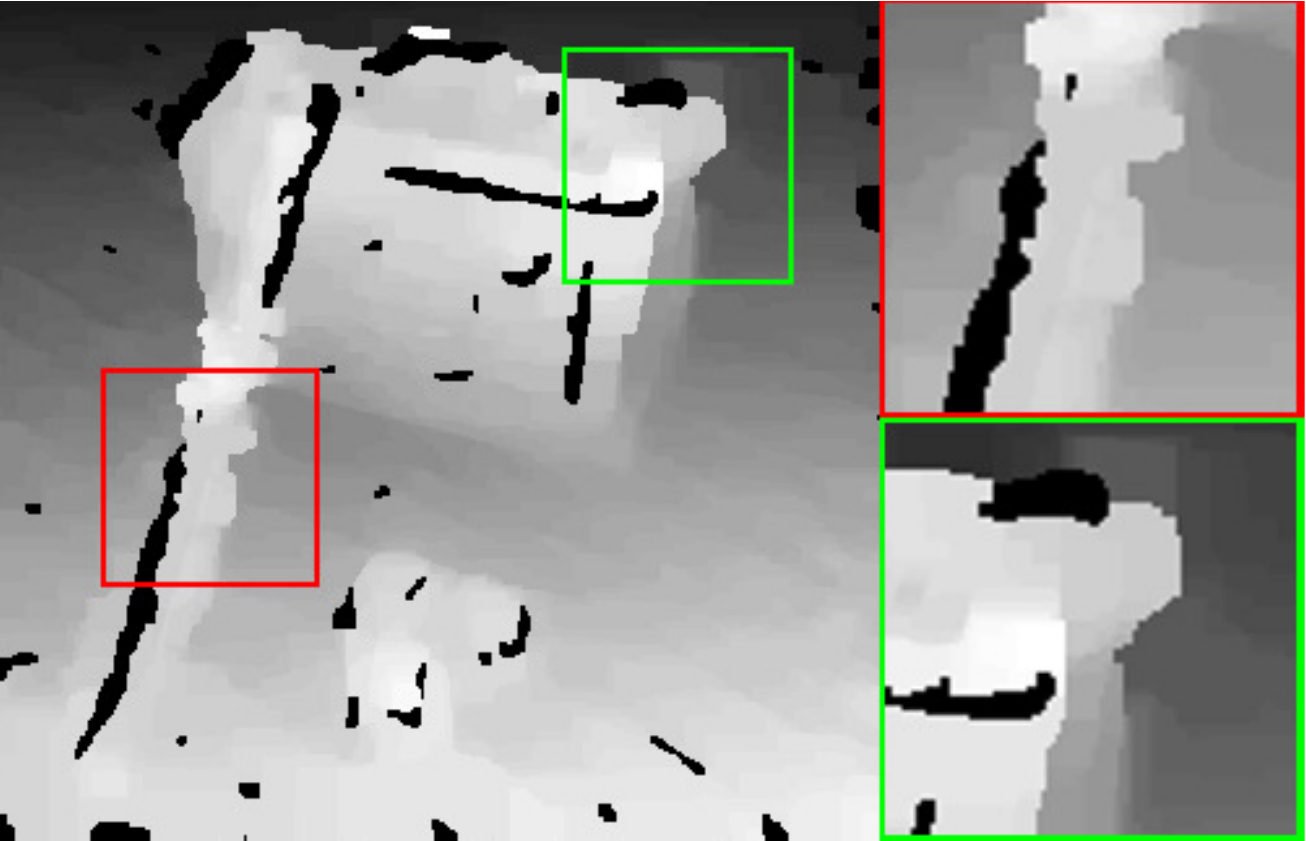}}\hspace*{0.01in}
		\subfloat[]{\includegraphics[width=0.245\textwidth]{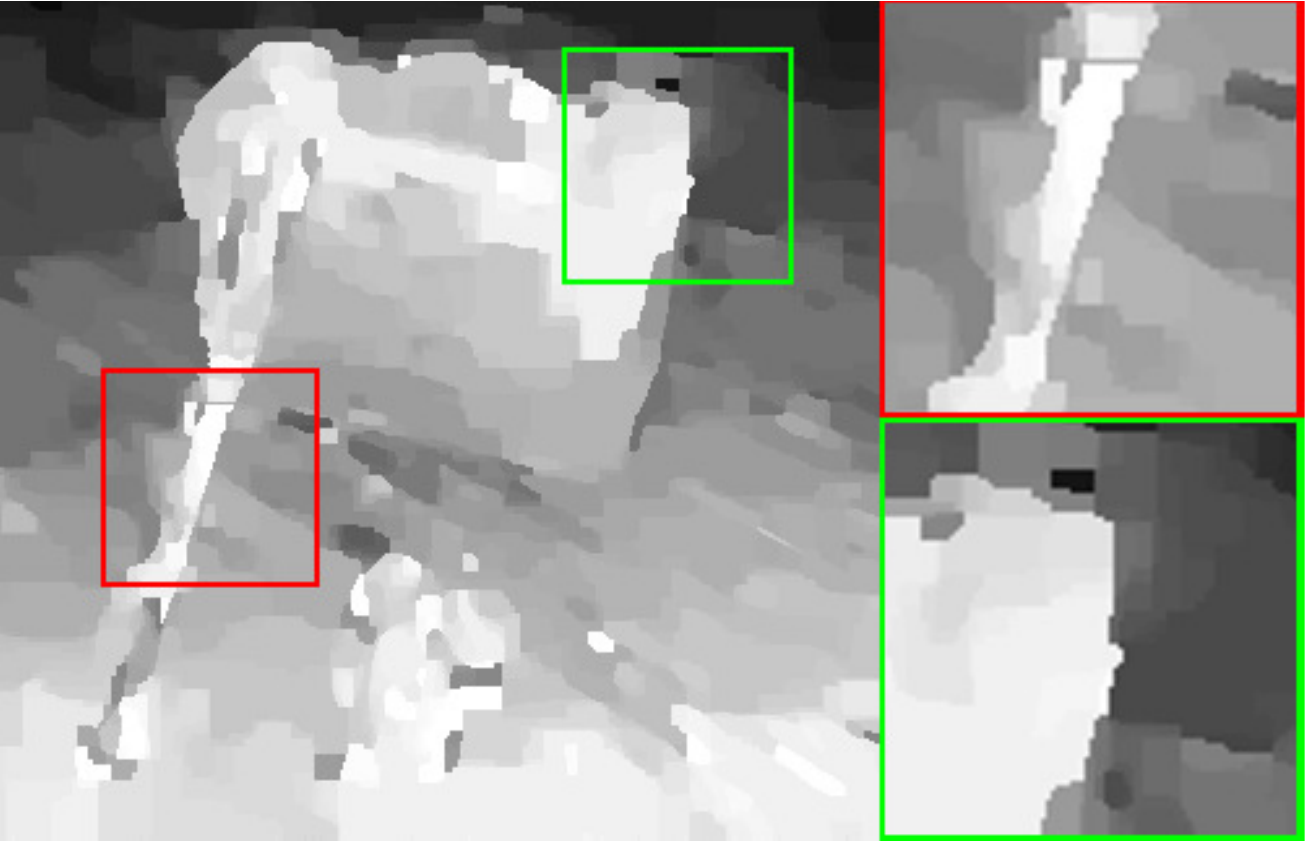}}\hspace*{0.01in}\\
		\vspace*{-0.125in}			
		\subfloat[]{\includegraphics[width=0.245\textwidth]{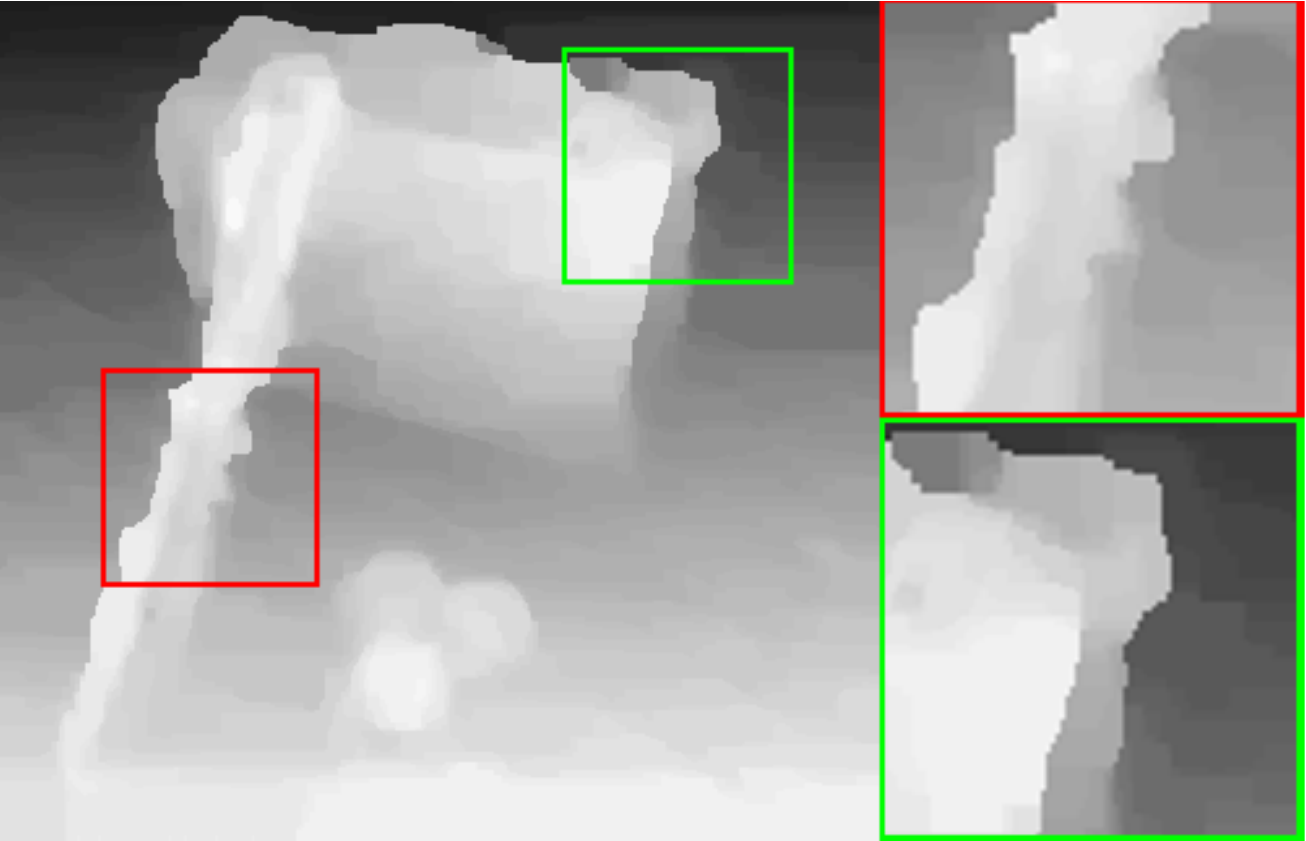}}\hspace*{0.01in}
		\subfloat[]{\includegraphics[width=0.245\textwidth]{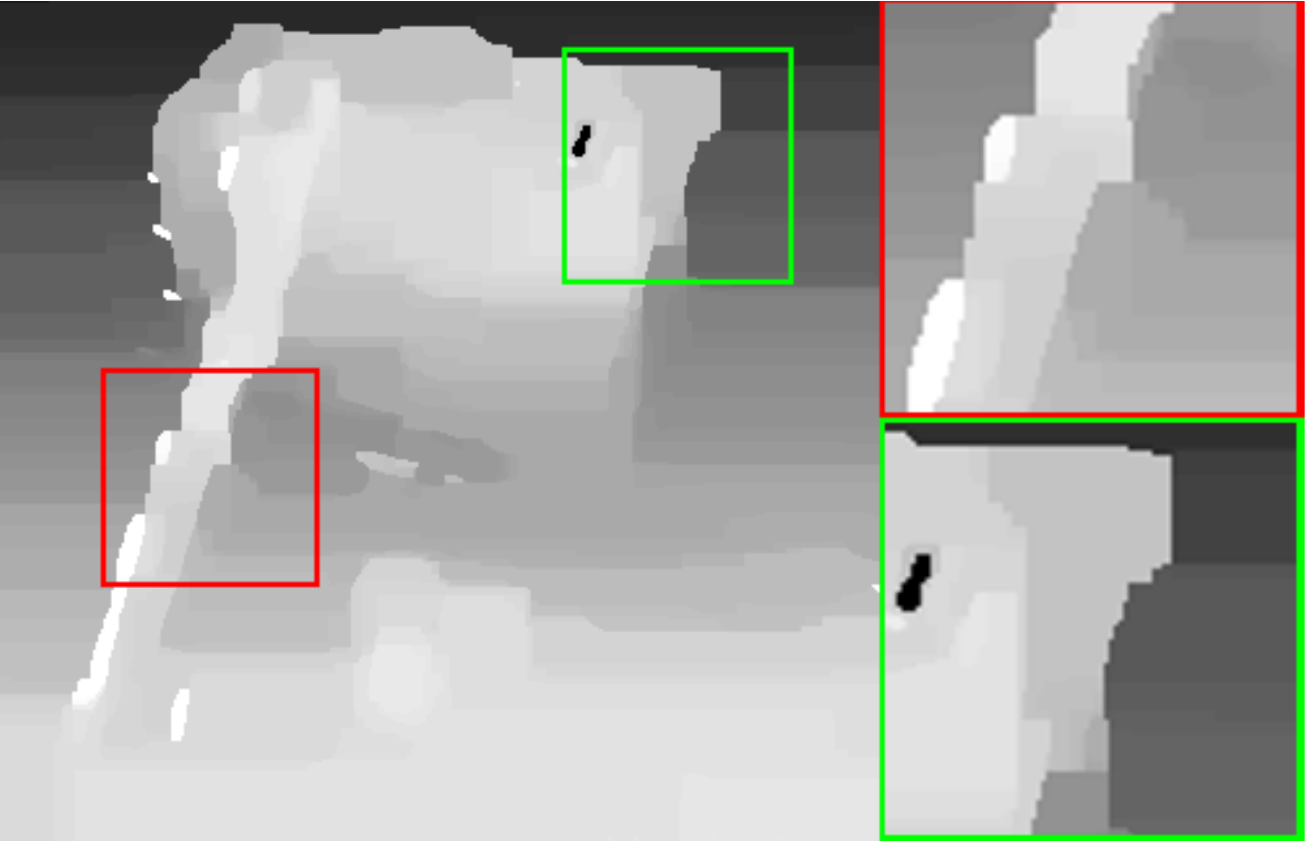}}\hspace*{0.01in}
		\subfloat[]{\includegraphics[width=0.245\textwidth]{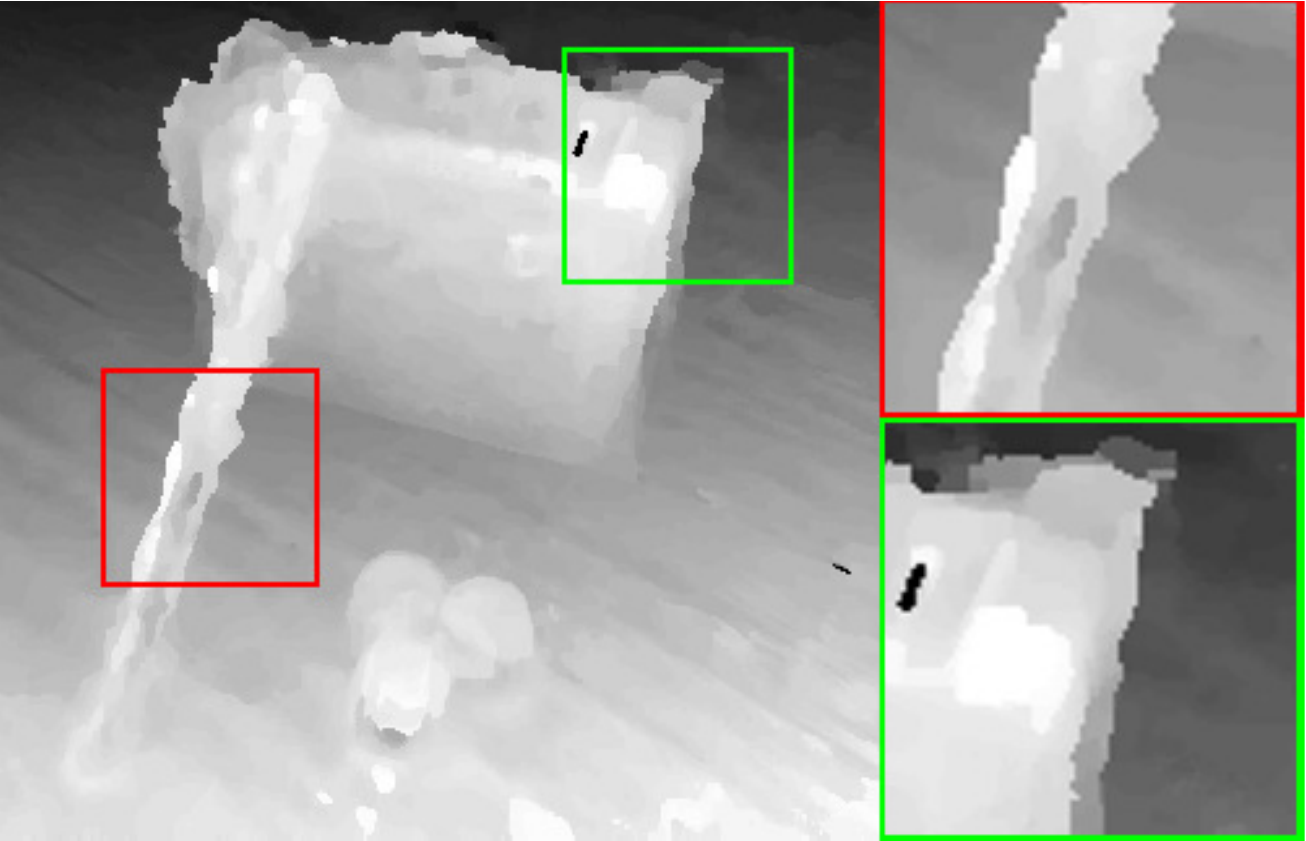}}\hspace*{0.01in}
		\subfloat[]{\includegraphics[width=0.245\textwidth]{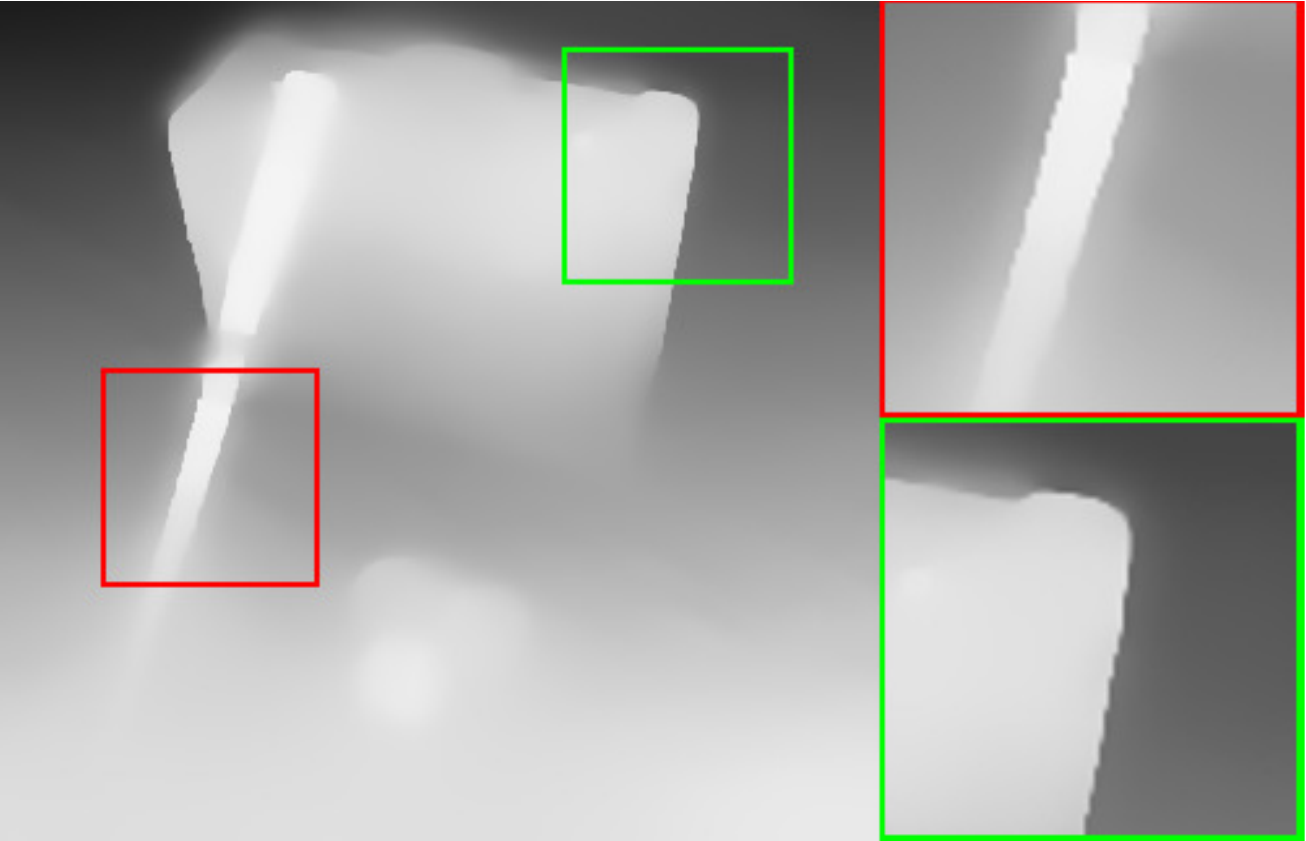}}\hspace*{0.01in}
	\end{center}	
	\vspace*{-0.2in}
	\caption{Depth estimation results on blurred light field. (a) Blurred center sub-aperture image. (b) Ground truth depth. (c) Result of Jeon et al.~\cite{jeon2015accurate}. (d) Williem and Park~\cite{williem2016robust}. (e) Tao et al.~\cite{tao2015depth}. (f) Wang et al.~\cite{wang2015occlusion}. (g) Chen et al.~\cite{chen2014light}. (h) Proposed algorithm.}
	\label{qual:depth}
\end{figure*}

Figure~\ref{qual:depth} shows the visual comparison of estimated depth map generated by different methods, which confirms that the proposed algorithm produces significantly better depth map in terms of accuracy and completeness.
Since independent deblurring of all sub-aperture images does not consider correlation between them, conventional correspondence and defocus cue do not produce reliable matching, yielding noisy depth map.
Only the proposed joint estimation algorithm results in sharp and unfattened object boundary, and produces the closest result to the ground truth.

Quantitative performance comparison of depth map estimation is shown in Table~\ref{quan:depth}.
For three synthetic scenes with three different motion for each scene, the average L1-rel error of the estimated depth map is computed and compared.
The comparison clearly shows that the proposed method produces the lowest error in all types of camera motion.
Note that the second best result is achieved by Chen et al.~\cite{chen2014light}, which is relatively robust in the presence of motion blur because bilateral edge preserving filtering is employed for cost computation.

\begin{table}[!t]
	\caption{Comparison of depth estimation (in average L1-rel error).}
	\centering
\renewcommand{\arraystretch}{1.1}{
	\scalebox{1.1}{
		\begin{tabular}{ccccc}
			\toprule
			Methods & Forward & Rotation & Trans. & Overall\\
			\cmidrule(r){1-1}\cmidrule(lr){2-2}\cmidrule(lr){3-3}\cmidrule(lr){4-4}\cmidrule(l){5-5}
			Chen et al.~\cite{chen2014light} & 0.0251 & 0.0326 & 0.0331 & 0.0303 \\
			Tao et al.~\cite{tao2015depth} & 0.0251 & 0.0359 & 0.0371 & 0.0327 \\
			Wang et al.~\cite{wang2015occlusion} & 0.0312 & 0.0377 & 0.0400 & 0.0363 \\
			Jeon et al.~\cite{jeon2015accurate} & 0.0835 & 0.0916 & 0.0921 & 0.0891 \\
			Williem and Park~\cite{williem2016robust} & 0.0615 & 0.0895 & 0.0966 & 0.0825 \\
			Proposed Method & \bf{0.0198} & \bf{0.0150} & \bf{0.0243} & \bf{0.0197} \\
			\bottomrule
		\end{tabular}
	}
}
	\vspace*{-0.2in}
	\label{quan:depth}
\end{table}

The depth estimation experiment demonstrates that consistent deblurring between sub-aperture images is essential.
The cues used in conventional depth estimation are seriously hampered in the independently deblurred image.
However, the proposed joint estimation algorithm results in robust and accurate depth estimation, providing the justification of solving deblurring and depth estimation in a joint manner.

\subsection{Camera Motion Estimation}
Table~\ref{quan:motion} shows the EPE of the estimated motion on synthetic light field dataset.
Compared with other methods~\cite{kim2014segmentation,sun2015learning}, the proposed method improves the accuracy of the estimated motion significantly.
In particular, a large gain is obtained in the rotational motion, which indicates that the rotational motion cannot be modeled accurately as a linear blur kernel used in~\cite{kim2014segmentation,sun2015learning}.

\begin{figure}[t]
\begin{center}
	\subfloat[]{\includegraphics[width=0.24\columnwidth]{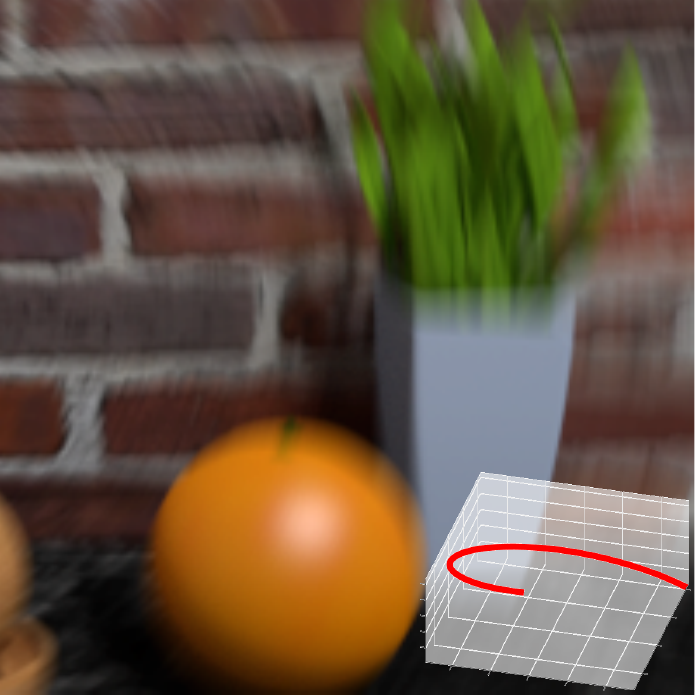}}\hspace*{0.05in}
	\subfloat[]{\includegraphics[width=0.24\columnwidth]{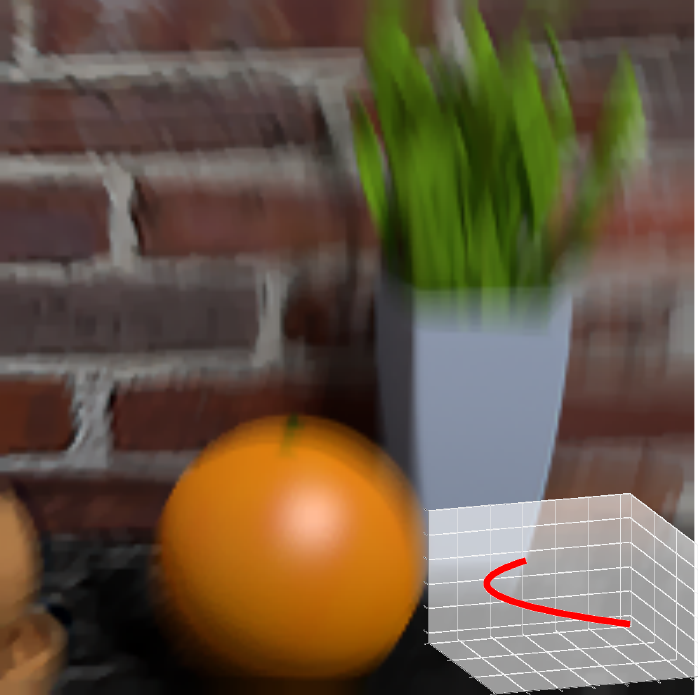}}\hspace*{0.03in}			
	\subfloat[]{\includegraphics[width=0.24\columnwidth]{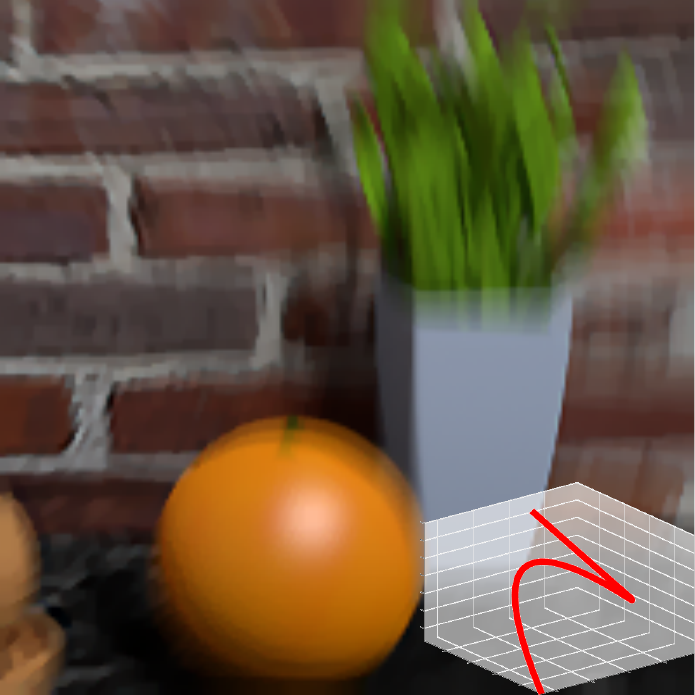}}\hspace*{0.03in}
	\subfloat[]{\includegraphics[width=0.24\columnwidth]{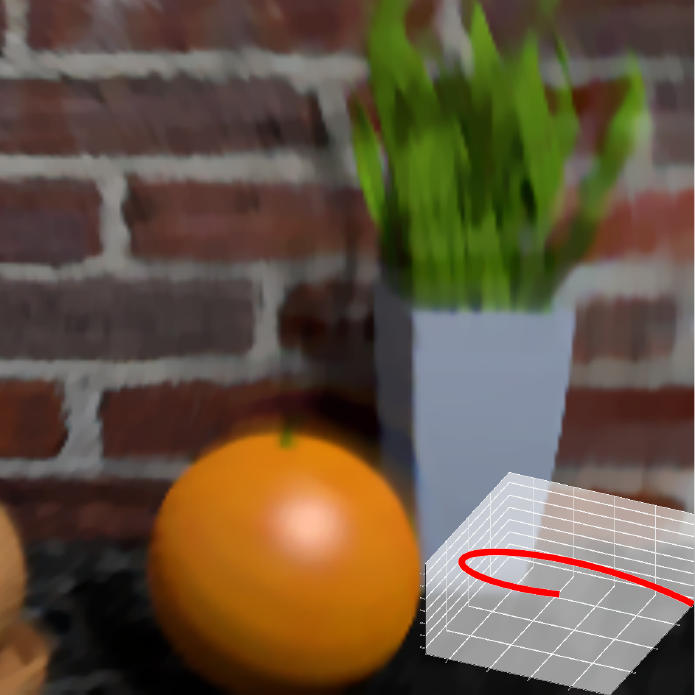}}\hspace*{0.03in}
	
	\vspace*{-0.05in}
\end{center}
\caption{Deblurring and camera motion estimation result for synthetic light field with comparison to \cite{srinivasan2017light}.
	(a) Input light field and ground truth camera motion.
	(b) Result of Srinivasan et al.~\cite{srinivasan2017light} (quadratic).
	(b) Srinivasan et al.~\cite{srinivasan2017light} (cubic).
	(d) Proposed algorithm.
}
\vspace*{-0.1in}
\label{fig:motion}
\end{figure}

Figure~\ref{fig:motion} shows the motion estimation results compared to the ground truth motion.
Since the camera orientation changes while the camera is moving, the 6-DOF camera motion can not be recovered properly by~\cite{srinivasan2017light}.
As shown in Figure~\ref{fig:motion}(b) and Figure~\ref{fig:motion}(c), the deblurring results are similar to the input, because the motion can not converge to the ground truth.
In contrast, the proposed algorithm converges to the ground truth 6-DOF motion and also produces the sharp deblurring result.
\begin{table}[t]
\caption{Comparison of motion estimation (in EPE).}
\centering
\renewcommand{\arraystretch}{1.1}{
	\scalebox{1.1}{
		\begin{tabular}{cccc}
			\toprule
			Methods & Forward & Rotation & Translation \\
			\cmidrule(r){1-1}\cmidrule(lr){2-2}\cmidrule(lr){3-3}\cmidrule(l){4-4}
			Kim and Lee~\cite{kim2014segmentation} & 2.153 & 3.317 & 1.989 \\
			Sun et al.~\cite{sun2015learning} & 1.492 & 2.557 & 1.810 \\
			Proposed Method & \bf{0.325} & \bf{0.171} & \bf{0.590}\\
			\bottomrule
	\end{tabular}}}
	\label{quan:motion}
	\vspace*{-0.05in}
\end{table}

\section{Conclusion}
In this paper, we presented the novel light field deblurring algorithm that estimated latent image, sharp depth map, and camera motion jointly.
Firstly, we modeled all the blurred sub-aperture images by center-view latent image using 3D warping function.
Then, we developed the algorithm to initialize the 6-DOF camera motion from the local linear blur kernel and scene depth.
In the iterative joint optimization, the nonlinear energy minimization was solved efficiently using IRLS.
The evaluation on both synthetic and real light field data showed that the proposed model and algorithm worked well with general camera motion and scene depth variation.

\bibliographystyle{splncs}
\bibliography{ECCV_JBMDDEL_arxiv}
\end{document}